\documentclass[10pt,twocolumn,letterpaper]{article}

\usepackage[pagenumbers]{cvpr} 

\usepackage[utf8]{inputenc} 
\usepackage[T1]{fontenc}    
\usepackage{amsfonts}       
\usepackage{nicefrac}       
\usepackage{microtype}      
\usepackage{epsfig}
\usepackage{algorithm}
\usepackage{algpseudocode}
\usepackage{mathtools}
\usepackage{listings} 
\usepackage{array,multirow}
\usepackage{ifthen}
\usepackage{tcolorbox}
\usepackage{adjustbox}
\usepackage{dblfloatfix}
\usepackage{balance}
\newcommand{\sqkm}{km$^2$ }
\newcommand{\refFig}[1]{Fig. \ref{#1}}
\newcommand{\refTab}[1]{Tab. \ref{#1}}

\DeclareMathOperator*{\argmin}{arg\,min}

\newcommand{\norm}[1]{\lVert #1 \rVert}
\newcommand{\defeq}{\overset{\mathrm{def}}{=\joinrel=}}
\newcommand{\degree}{$^{\circ}$}
\newcommand{\cross}{$\times $}
\newcommand{\red}[1]{{\color{red}#1}}

\newcommand{\green}[1]{{\color{green}#1}}
\definecolor{flgreen}{HTML}{08FF08}
\newcommand{\flgreen}[1]{{\color{flgreen}#1}} 
\newcommand{\TBD}{To be disclosed later.}
\newcommand{\NA}{Not applicable.}

\definecolor{cvprblue}{rgb}{0.21,0.49,0.74}
\usepackage[pagebackref=true,breaklinks=true,colorlinks=true,bookmarks=false,urlcolor=magenta,citecolor=ForestGreen]{hyperref}

\newcommand{\useAck}{True}
\newcommand{\anonSub}{False}

\newcommand{\github}{\href{https://github.com/rahuldeshmukh43/satdepth}{satdepth.git}}

\title{SatDepth: A Novel Dataset for Satellite Image Matching}

\author{
    Rahul Deshmukh
    \quad Avinash Kak\\[1mm]
    {\tt\small deshmuk5@purdue.edu} 
    \quad {\tt\small kak@purdue.edu}\\[1mm]
    Purdue University, West Lafayette
}

\begin{document}
\maketitle

\begin{abstract}
    Recent advances in deep-learning based methods for image matching have demonstrated their superiority over traditional algorithms, enabling correspondence estimation in challenging scenes with significant differences in viewing angles, illumination and weather conditions. However, the existing datasets, learning frameworks, and evaluation metrics for the deep-learning based methods are limited to ground-based images recorded with pinhole cameras and have not been explored for satellite images. In this paper, we present ``SatDepth'', a novel dataset that provides dense ground-truth correspondences for training image matching frameworks meant specifically for satellite images. Satellites capture images from various viewing angles and tracks through multiple revisits over a region. To manage this variability, we propose a dataset balancing strategy  through a novel image rotation augmentation procedure. This procedure allows for the discovery of corresponding pixels even in the presence of large rotational differences between the images. We benchmark four existing image matching frameworks using our dataset and carry out an ablation study that confirms that the models trained with our dataset with rotation augmentation outperform (up to 40\% increase in precision) the models trained with other datasets, especially 
    when there exist large rotational differences between the images.
    The dataset and code will be made available through \github.
  \end{abstract}
  
  \vspace*{-1ex}
  \section{Introduction}
  \label{sec:intro}
  Image matching is a fundamental problem in computer vision with applications in 3D reconstruction, image retrieval, visual localization, pose estimation, scene understanding etc. It involves finding correspondences between two images which can be used to estimate a geometric transformation between the images. Traditionally, the process of image matching can be broken into three different phases: (1) Feature extraction; (2) Feature description; and (3) Feature matching. In the detection phase, interest points such as corners \cite{harris_corner} or blob centers \cite{sift} are detected in the images. In the description phase, a descriptor vector is computed for each detected interest point using its local neighborhood. In the matching phase, the descriptors are compared using Nearest Neighbor (NN) search or other methods to find the best matches between the two images.

  \begin{figure}[!t]
    \centering
    \input{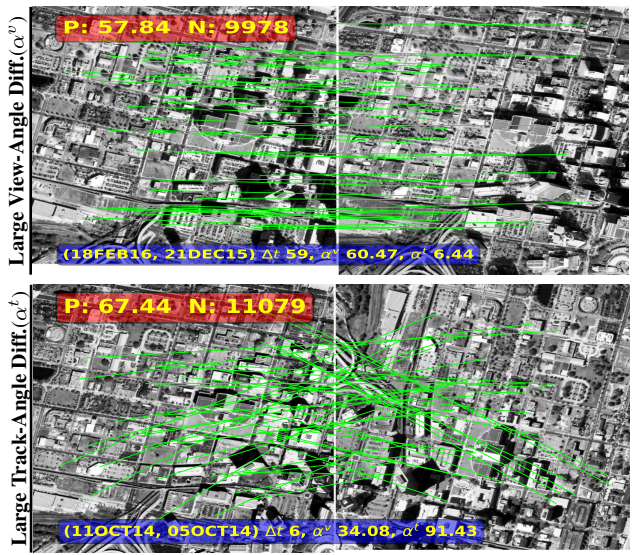}
    \caption{Image matches learned by satLoFTR trained using SatDepth dataset with rotation augmentation for image pairs with significant differences. The \flgreen{green} lines depict only 40 randomly chosen correctly detected matches.}
    \label{fig:intro_razzle_dazzle}    
  \end{figure}
  
  Recently, several learning-based approaches \cite{Sun_CVPR21_LOFTR, Li_NIPS20_DUALRC,Rocco_NIPS18_NCNET, Rocco_ECCV20_sparseNCNet, Qianqian_ECCV20_CAPS, MagicLeap_CVPR18_SuperPoint, MagicLeap_CVPR20_SuperGlue} have been developed for image matching of ground-based images that were shown to outperform the traditional feature detectors \cite{harris_corner, sift, surf, orb, FAST, brief} on benchmark datasets. These learning-based approaches either modify or combine one or more phases of the  traditional approaches to image matching. Due to their superior performance, 
  we now have the Image Matching Challenge \cite{Jin2020} and the annual ``Image Matching'' workshop that goes with it.
  
  One major contributing factor for the success of learning-based approaches is the availability of large-scale datasets \cite{aachen_day_night,taira2018inloc,dai2017scannet, MegaDepthLi18} for training and evaluation. These datasets provide a way to extract ground-truth correspondences between pairs of images using camera pose and/or depth information. The dataset construction methods can range from manual collection of dense ground-truth correspondences using a depth sensor \cite{chang2017matterport3d, dai2017scannet, taira2018inloc} to automatic generation of sparse or dense pseudo ground-truth correspondences \cite{MegaDepthLi18} using Structure from Motion (SfM) \cite{schoenberger2016sfm} and multi-view stereo reconstruction \cite{schoenberger2016mvs} respectively. 
  
  Despite the success of learning-based approaches to image matching for ground-based imagery recorded with pinhole cameras, similar approaches have not yet been explored for satellite images. This can be attributed to the following reasons: (1) Limited number of publicly available high resolution satellite images;  (2) Complex processing pipelines for 3D reconstruction of satellite images; and (3) The pushbroom camera model for satellite images does not support the construction of conventional depth maps for extracting matching points. 
  Another very important reason for why the learning-based approaches have not been explored for satellite images
   is due to the \textbf{imbalance} in the distribution of available images with respect to the track-angle differences (\cref{fig:pair_distribution_main}). While satellites can capture images at various view angles, track-angle diversity is limited, leading to sparse and uneven distribution. This imbalance, if unaddressed, can degrade image-matching network generalization. 
  Furthermore, on account of the fact that satellite images are recorded with pushbroom cameras, the pinhole camera based metrics for ground-based imagery cannot be applied directly to evaluate the performance of image matching networks for satellite images. 
  Motivated by these gaps in the literature, we propose a novel benchmarking dataset for satellite image matching called SatDepth.

  Our method for extracting ground-truth correspondences between satellite images relies on height estimates above the latitude-longitude ($lat$-$lon$) plane. The height values are derived from a Digital Surface Model (DSM) constructed using a stereo matching algorithm applied to the satellite images.
  We opted against using LiDAR DSM due to its more limited geographic coverage compared to satellite data. Using the stereo-DSM, we construct ``SatDepth Maps'', analogous to depth maps in ground-based imagery, to associate a ``$(lat,\, lon,\, h)$-triples'' with each pixel in the satellite images. These enable us to train and evaluate image matching networks for satellite images, facilitating the generation of dense matches as shown in \cref{fig:intro_razzle_dazzle}.

  To summarize, our main contributions are: (\textbf{\textit{i}}) We construct a novel dataset for satellite image matching and verify its accuracy; 
  (\textbf{\textit{ii}}) We extend the existing image matching metrics for ground-based images to satellite images; and (\textbf{\textit{iii}}) We train four state-of-the-art \cite{wang2022matchformer,Sun_CVPR21_LOFTR,Li_NIPS20_DUALRC,Qianqian_ECCV20_CAPS} image matching networks on our dataset with a novel rotation augmentation procedure and, based on our experiments, provide recommendations for training image matching networks with satellite images. 

  \section{Related Work}
  \textbf{Image Matching Datasets and Models}:
  Image matching for ground-based images has been an active area of research with a rich history. This domain has relied on several seminal algorithms for sparse correspondence detection, such as SIFT \cite{sift} and SURF \cite{surf}. These algorithms have been integral components of numerous 3D reconstruction pipelines like COLMAP \cite{colmap} and others \cite{wu2011visualsfm, moulon2016openmvg}. However, they perform poorly when presented with images that have significant variations in viewing angles, illumination, seasonal changes, presence of repeating patterns, scene changes due to human activity etc.
  
  Several datasets have been curated to capture the above variations and facilitate the training of learning-based methods for image matching. MegaDepth \cite{MegaDepthLi18} offers outdoor images with varying viewing angles, accompanied with ground-truth camera poses and depth maps generated using COLMAP. ScanNet \cite{dai2017scannet} provides RGB-D data for indoor scenes, for which depth maps were acquired using a depth camera and poses were estimated using \cite{dai2017bundlefusion}. The Aachen Day-Night \cite{aachen_day_night} is comprised of images of the same scene captured at different times of day and night, camera poses computed using manual annotation of image matches and an underlying 3D model.
  
  These datasets have been used to train several state-of-the-art image matching networks. The first generation of matching networks replaced Difference of Gaussian (DoG) keypoint extractors with Convolutional Neural Network (CNN) feature extractors over image pyramids \cite{LIFT,KEY_NET,D2NET, LF_NET}. These models involved separate modules for keypoint detection, description, and orientation computation, and were trained using a variety of losses including Margin loss \cite{D2NET}, Triplet loss \cite{LF_NET}, Regression \cite{LIFT, LF_NET} and Intersection over Union (IoU) \cite{LIFT}. 
  The second generation of networks \cite{GEODESC,MagicLeap_CVPR18_SuperPoint, Rocco_NIPS18_NCNET} combined the detection and description phases into a single architecture and were trained end-to-end, mimicking key components of traditional image matching pipelines based on the Neighborhood Consensus \cite{Rocco_NIPS18_NCNET} and the SIFT ratio test \cite{GEODESC}. 
  The latest generation of networks \cite{Qianqian_ECCV20_CAPS, Li_NIPS20_DUALRC, Sun_CVPR21_LOFTR, MagicLeap_CVPR20_SuperGlue,lindenberger2023lightglue,Patch2PixEP,Rocco_ECCV20_sparseNCNet} have also formulated differentiable losses for geometric matching.

  Architecturally, networks have progressed from CNNs \cite{LIFT,KEY_NET,D2NET, LF_NET,GEODESC} to the adoption of Transformers \cite{Sun_CVPR21_LOFTR, wang2022matchformer} and Graph Neural Network \cite{MagicLeap_CVPR20_SuperGlue,lindenberger2023lightglue}. To manage the search space for matching, recent networks \cite{Li_NIPS20_DUALRC,Sun_CVPR21_LOFTR,wang2022matchformer,Patch2PixEP,MagicLeap_CVPR20_SuperGlue} have converged to a \textit{``coarse-to-fine''} detection strategy. Training these networks involves various methods, including weak supervision with class labels \cite{Rocco_NIPS18_NCNET} or camera pose \cite{Qianqian_ECCV20_CAPS}, dense supervision using depth and camera poses \cite{Sun_CVPR21_LOFTR,Li_NIPS20_DUALRC}, and reinforcement learning \cite{DISK}.
  
  \vspace*{1ex}
  \noindent\textbf{Satellite Image Datasets and Processing Pipelines}: 
  The past decade has witnessed increased availability of high-resolution satellite images, 
  \ie 
  images at 0.25 - 0.5 m Ground Sampling Distance (GSD) from commercial vendors like Maxar and PlanetLabs. This has spurred research in various areas, including 3D reconstruction \cite{Franchis_ISPRS14_S2P,nasa_ames,qin2016rpc}, road and building detection \cite{comandur_JSTARS21_noisyosm, Somrita_JSTARS22_segmentation}, and change detection \cite{SN7_muds}. To foster research in these areas, the community has created several well-curated datasets such as DFC-2019 \cite{ieee_dfc2019_mvs}, SpaceNet Challenges \cite{SN7_muds, SN8}. Additionally, several open-source 3D reconstruction pipelines for satellite images are currently under development, including the NASA's ASP \cite{nasa_ames}, S2P \cite{Franchis_ISPRS14_S2P}, and others \cite{qin2016rpc, dlr_catena, rupnik20183d_micmac}. These pipelines vary in their capabilities and approaches to 3D reconstruction.

  However, these pipelines still rely on traditional matching algorithms based on SIFT. While learning-based approaches for matching ground-based images recorded with pinhole cameras have gained traction \cite{Jin2020}, the same cannot be said for satellite images recorded with pushbroom cameras. Prior works \cite{marcos2016geospatial, vakalopoulou2019image} have employed deep-learning methods for registering satellite images by estimating a warping function. However, these methods are limited to orthorectified (top-down view) images and do not extract matching points. This limitation makes them unsuitable for our problem, as we need 
  to match multi-view images recorded from diverse viewpoints.  
  
  \section{Dataset Generation}
  \label{sec:dataset_generation}
\begin{figure}[!h]
    \centering
    \input{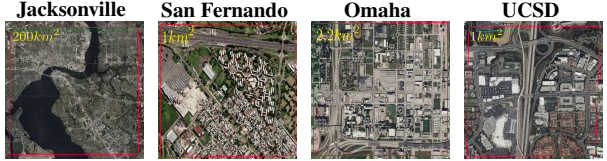}
    \caption{Spatial extents of each AOI (red box) in the SatDepth dataset overlaid on Bing Maps.}
    \label{fig:sat_coverage}
    \vspace{-1em}
  \end{figure}

  %
  We used the following two sources of high-resolution satellite images for creating the SatDepth dataset: (1) the COREED dataset \cite{Brown_SPIE2018_core3d, core3d_aws_public, core3d_iarpa}; and (2) the MVS3DM dataset \cite{bosch2016multiple_mvs3d,mvs3d_aws_public}.  Both these satellite image datasets are drawn from the WorldView (WV) images, with their spatial resolution of 0.25 - 0.5 m GSD.
  %
  These two satellite image datasets contain a total of 198 panchromatic (PAN) images over four Areas of Interest (AOI).
  These AOIs encompass diverse terrains, varying satellite viewing angles and satellite tracks.
  To create the SatDepth dataset, we selected a spatial subset of the four AOIs, as shown in \cref{fig:sat_coverage}. We chose Jacksonville as our large-area AOI as we had Ground Control Points (GCPs) available for this region, allowing us to assess the accuracy of the SatDepth dataset (\cref{sec:dataset_accuracy}).
  We present details pertaining to view distribution and image coverage of the SatDepth dataset in the supplementary material.

  SatDepth serves as a valuable dataset for training image matching models on a single large-area AOI and evaluating them across other geographical regions. The subsequent sub-sections provide an overview of the satellite camera model, details of our processing pipeline, and an assessment of the dataset's accuracy.

  \subsection{Satellite Camera Model}\label{sec:camera_model}
  A satellite image is captured with a pushbroom camera, which consists of a linear array of sensors that records one row of the image at a time as the satellite moves along its track.  The satellite's motion makes its camera model more intricate than that of ground-based pinhole cameras. Instead of a physics-based model, vendors provide users with a third-order approximate model called the Rational Polynomial Coefficients (RPC) camera model. This model facilitates the mapping of a 3D world point $(\boldsymbol{X}\in\mathbb{R}^3)$ to a pixel location $(\boldsymbol{x}\in\mathbb{R}^2)$ in the image. This mapping is also known as the forward projection and is denoted by $\boldsymbol{x}=\mathcal{P}(\boldsymbol{X})$. The inverse mapping, known as the back projection, results in a 3D ray joining the camera center and the pixel location ($\boldsymbol{x}$). Due to the non-linear nature of $\mathcal{P}$, there is no closed-form solution for the inverse mapping. However, given the height ($h$) of a 3D point on this ray, we can compute the corresponding $(lat,\, lon)$ by minimization of the reprojection error. We denote this operation as: $\,lat,\, lon = \mathcal{P}^{-1}(\boldsymbol{x},h)$. We present more details in the supplementary material.

  While $\mathcal{P}$ is a non-linear function, it has been shown in the literature \cite{Franchis_ISPRS14_S2P,okamoto1993orientation} that for a small image patch, it can be approximated by an affine camera, $\widehat{\mathcal{P}}$, using a first-order Taylor series.
  We will use both $\mathcal{P}$ and $\widehat{\mathcal{P}}$ in our processing pipeline as well as for training and evaluation of image matching networks. 
  
  \begin{figure*}[t]
    \centering
    \includegraphics[width=\linewidth]{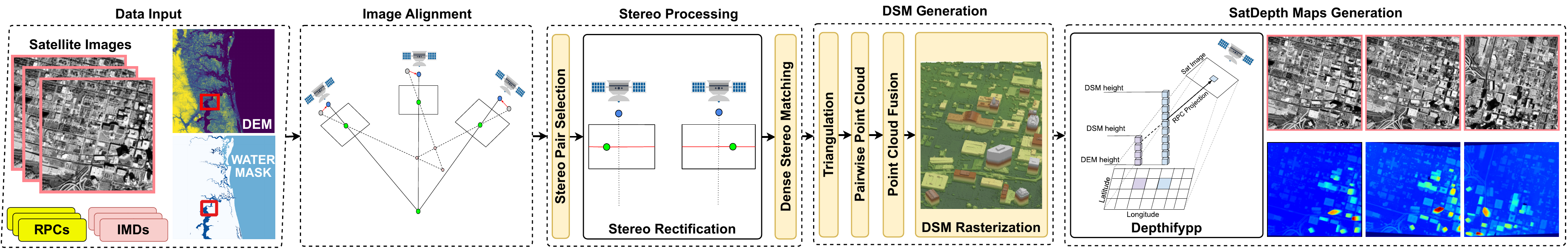}
    \caption{SatDepth processing pipeline: given a collection of satellite images (with cameras (RPCs) and metadata (IMDs)) and auxiliary data (DEM and Water Mask), we carry out a series of processing steps to obtain SatDepth Maps which are used to extract ground-truth correspondences.} 
    \label{fig:sat_pipeline}
  \end{figure*}

  \subsection{Processing Pipeline}\label{sec:processing_pipeline}
  As mentioned in \cref{sec:intro}, SatDepth Map associates a $(lat,lon,h)$ coordinate with each pixel in the satellite images. To generate the SatDepth Maps, we need aligned cameras and a 3D reconstruction of the scene.
  \Cref{fig:sat_pipeline} shows the processing steps that are carried out for generating the SatDepth Maps.
  The alignment of satellite images and construction of 3D reconstruction for satellite images is more complex than for ground-based images due to the nature of the nonlinearity involved in how an image is recorded by a satellite that is in constant motion. 
  Additionally, individual satellite images can be as large as 30,000 \cross 16,000 pixels, necessitating highly parallel and distributed modules.
  What follows is a brief explanation of individual steps of our SatDepth processing pipeline.

  \subsubsection{Pre-processing}\label{sec:preprocessing}
  After the satellite images have been subject to what is known as radiometric correction \cite{wv2_tech_note}, 
  we then divide the set of images over an AOI into groups 
  on the basis of the coverage over 
  to 1.6 km \cross 1.6 km ground tiles. Each tile includes a central 1 km \cross 1 km area with a 300 m overlapping region with adjacent tiles. This size allows for affine approximations of the camera models for the images covering the tile.
  The reason for padding the tiles is explained in \cref{sec:dsm_generation}. 
  Tiling also facilitates distributed processing. For SatDepth, we carried out the tiling process for Jacksonville (196 tiles) only. The other AOIs are small enough to require only a single tile.
  
  \subsubsection{Image Alignment}\label{sec:image_alignment}
  As mentioned earlier, we need aligned cameras to generate SatDepth Maps. 
  The initial camera parameters provided by a satellite-image vendor have residual alignment errors that are corrected using Bundle Adjustment (BA). BA is an important step as it allows for accurate 3D triangulation and stereo fusion. Grodecki and Dial \cite{Grodecki_ASPRS03_blockAdjustment} demonstrated that the residual misalignment error can be modeled using a bias ($\boldsymbol{b}\in\mathbb{R}^2$) correction in the image plane which can be visualized as a shift correction of camera centers as shown in \cref{fig:sat_pipeline}. BA is formulated as a Maximum a Posteriori (MAP) problem to estimate optimal bias ($\boldsymbol{b}^{*}$) for all images. We denote un-aligned parameterized camera as $\mathcal{P}(\boldsymbol{b}, \boldsymbol{X})$. Given  a set of $N$ images and cameras ($\{I_{i}, \mathcal{P}_{i}\}_{i=1}^{N}$), BA starts with feature extraction using SIFT for all images and then performs pairwise feature matching using nearest-neighbor search for all pairs $\mathcal{S}=\{(i,j)|1 \leq i < j \leq N\}$. We then carry out pairwise outlier rejection using RANSAC and obtain the set of inlier correspondences ($\{\mathcal{M}_{ij}\} \, \forall \, (i,j)\in \mathcal{S}:  \mathcal{M}_{ij} = \{ \boldsymbol{x}^{k}_{i} \leftrightarrow \boldsymbol{x}^{k}_{j}|\boldsymbol{x}^{k}_{i} \in I_{i}, \boldsymbol{x}^{k}_{j} \in I_{j} \}$). We associate a putative world point with every match ($\boldsymbol{x}^{k}_{i} \leftrightarrow \boldsymbol{x}^{k}_{j} \leftrightarrow \boldsymbol{X}^{k}_{ij}$) and the MAP problem is formulated as a minimization of L-2 regularized reprojection error as shown in \cref{eq:bundle_adjustment}.
  
  \vspace*{-1em}
  \begin{align}
    \begin{split}
    \!
    \boldsymbol{b}^{*}_{i}, \boldsymbol{X}^{*k}_{ij} \!
    &= \! \argmin\!\!\!
    \sum_{ (i,j)\in \mathcal{S} } \!\! \sum_{k=1}^{|\mathcal{M}_{ij}|} \!\!
    \Bigl( 
    \norm{\boldsymbol{x}^{k}_{i} - \mathcal{P}_i(\boldsymbol{b}_i, \boldsymbol{X}^{k}_{ij})}^2_{2}\\
    &+
    \norm{\boldsymbol{x}^{k}_{j} - \mathcal{P}_j(\boldsymbol{b}_j, \boldsymbol{X}^{k}_{ij})}^2_{2}
    \Bigr)
    +  \lambda\!\sum_{i=1}^{N}\norm{\boldsymbol{b}_i}^2_{2} \!\! 
    \end{split}
    \label{eq:bundle_adjustment}
  \end{align}
  
  We choose $\lambda = 0.5$ and solve \cref{eq:bundle_adjustment} using Sparse BA \cite{Lourakis_TOMS09_SBA}. We can also use \cref{eq:bundle_adjustment} for triangulation, which is the process of computing world points when given a set of correspondences. We carry out triangulation by removing the regularization term ($\lambda=0$) from \cref{eq:bundle_adjustment} and solving for the unknown world points.
   We align all the tiles independently and provide the corrected cameras 
   in SatDepth.

  Image alignment is followed by 3D scene reconstruction that consists of the following two stages: (1) Stereo Processing, and (2) DSM Generation. What follows is a brief overview of the stages.
  
  \vspace*{-1em}
  \subsubsection{Stereo Processing}
  The goal of stereo processing is to compute a dense set of pixel-to-pixel matches for the image pairs that are selected for the purpose. Listed below are the three main steps of stereo processing.
  
  \vspace*{1ex}
  \noindent \textbf{Stereo Pair Selection}: Satellite images come with metadata that provide information about the satellite and sun angles, and also the image acquisition time. For stereo matching, we aim to construct image pairs based on two competing considerations: (1) Increasing the baseline separation enhances triangulation accuracy, while (2) Reducing the baseline separation improves the reliability of searching for corresponding pixels. These considerations must be balanced alongside the differences in image acquisition time, sun angles, and other factors, necessitating the use of effective heuristics. We employ heuristics similar to \cite{Facciolo2017Automatic3R,sonali_arxiv_SatStereo, comandur_JSTARS21_noisyosm} to select ``good'' stereo pairs.
  %

  %
  \vspace*{1ex}
  \noindent \textbf{Stereo Rectification}: We follow an approach similar to \cite{Franchis_ISPRS14_S2P} for stereo rectification, employing affine cameras $\widehat{\mathcal{P}}$. Initially, for a given image pair, we determine the world point corresponding to the image center using the inverse mapping $\mathcal{P}^{-1}$.
   Subsequently, we derive $\widehat{\mathcal{P}}$ for both images, enabling the computation of the affine fundamental matrix $\widehat{F}$ \cite{hartley2003multiple, Franchis_ISPRS14_S2P}. Leveraging $\widehat{F}$, we calculate rectification homographies and resample the images. These rectification homographies are retained for subsequent computations.

  \vspace*{1ex}
  \noindent \textbf{Dense Stereo Matching}: After rectification, we generate disparity maps using t-SGM \cite{rothermel2017development}. The disparity maps themselves are not stored; instead, they are used to construct a dense set of image matches ($\boldsymbol{x}^{k}_{i} \leftrightarrow \boldsymbol{x}^{k}_{j}$) using the rectification homographies to convert rectified coordinates to image coordinates. Since dense stereo matching is computationally intensive, we process a maximum of 80 ``good'' stereo pairs and then fuse the information in the next step.
  
  \subsubsection{Digital Surface Map Generation}
  \label{sec:dsm_generation}
  In this step, we fuse the information from all stereo pairs to produce a stereo-fused DSM. The DSM is a 2.5 D height source which records the height of the highest visible point with a nadir view (top-down view). The DSM construction can be broken into three steps: (1) \textbf{Triangulation}: Here we triangulate the dense stereo matches ($\boldsymbol{x}^{k}_{i} \leftrightarrow \boldsymbol{x}^{k}_{j}$) to obtain the set of world points $\boldsymbol{X}^{k}_{ij}$. For triangulation, we use pairwise BA as explained in \cref{sec:image_alignment}.  We store the list of triangulated points as the stereo point cloud. (2) \textbf{Pairwise Point Cloud Fusion}: Here, we concatenate the lists of world points for all stereo point clouds, resulting in what we call the fused-point cloud. (3) \textbf{DSM Rasterization}: 
  The last step consists of creating a grid of cells in the $lat,\ lon$ plane with a resolution of 0.25 m. Then the height values from the fused-point cloud are accumulated over the cells. Finally, we compute the median of the top-N points at each cell to produce the DSM.

  Since t-SGM performs poorly around borders of the images, we initially generate DSMs for tiles with padding (\cref{sec:preprocessing}), and then crop them to remove the padding.

  Now that we have aligned the cameras and generated a DSM, we proceed to generate the SatDepth Maps as explained in the next section.
  
  \subsubsection{SatDepth Map Generation}\label{sec:depth_generation}
  When creating datasets for training image matching networks meant for ground-based images with pinhone cameras, 
  depth maps are used to associate a world point in the camera's relative coordinate frame with each pixel. However, in satellite imagery, we associate a world point $(lat,\, lon,\, h)$ in an absolute coordinate frame with each pixel. We store this information in what we call ``SatDepth Maps''. To generate these maps, we have to first construct a 3D model of the scene and project it onto the image. Given that the DSM only records the height of the highest visible point from a nadir view, we construct all building façades by extending the roof boundaries to ground level. This assumption simplifies the 3D model but may omit details such as overhangs, underbelly of spherical water towers, etc.

  To generate the SatDepth Maps, we create a 3D grid of points with fixed grid spacing ($\Delta_z$) spanning from ground height to roof height for each $(lat,\, lon)$ position, excluding water areas. We obtain the ground height from \cite{srtm_dem} and the roof height from our DSM. Next, we project this 3D grid onto the image using $\mathcal{P}$ as defined in \cref{sec:camera_model}. For each pixel location, we retain the coordinates of the 3D point with the largest height value as shown in \cref{fig:sat_pipeline}, ensuring we capture the location of the visible point. Due to the high computational cost of this procedure, we developed a C++ and OpenMP-based module called \textbf{\textit{depthifypp}} for efficient depth computation. In addition to multi-threading, \textbf{\textit{depthifypp}} limits memory usage by distributing processing to smaller blocks as explained in supplementary material. This module is being made publicly available.

  Finally, to extract a dense set of ground-truth correspondences ($\boldsymbol{x}^{k}_{i} \leftrightarrow \boldsymbol{x}^{k}_{j}$) for a given image pair ($I_i, I_j$), we first read the world point, $\boldsymbol{X}_i^{k}$, at $\boldsymbol{x}_i^{k}$ from SatDepth Maps of $I_i$. We then compute its projection, $\boldsymbol{x}_j^{k} = \mathcal{P}_j(\boldsymbol{X}_i^{k})$, in $I_j$. Then we read the world point, $\boldsymbol{X}_j^{k}$, at $\boldsymbol{x}_j^{k}$ from SatDepth Maps of $I_j$ and compute the distance $\norm{\boldsymbol{X}_i^{k} - \boldsymbol{X}_j^{k}}_2$. If this distance is below a threshold $\delta_{3D}$, then we declare ($\boldsymbol{x}^{k}_{i},\,\boldsymbol{x}^{k}_{j}$) as a true match. The accuracy of correspondences extracted through this process is limited by the accuracy of the DSM. 
  %
  
  To generate the SatDepth dataset, we used 6 virtual machines, each with 16 VCPUs and 120 GB RAM, hosted on a OpenStack-based cloud infrastructure. The entire dataset generation process took approximately 30 days, with stereo processing being the most computationally intensive step.
  
  \subsection{Dataset Accuracy Assessment}\label{sec:dataset_accuracy}
  We assess the accuracy of our dataset using both LiDAR and GCPs as discussed below.
  
\begin{figure}[!h]
    \centering
    \input{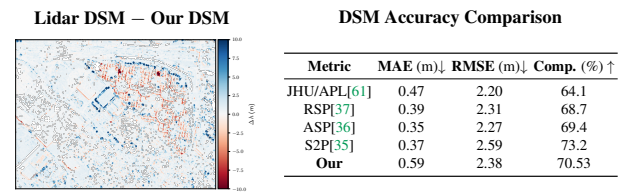}
    \caption{Comparison of the quality of the DSM generated from different stereo pipelines over San Fernando.}
    \label{fig:pipeline_comparison_main}
  \end{figure}

  \vspace*{1ex}
  \noindent \textbf{DSM Accuracy using LiDAR}:  We follow the approach of \cite{bosch2017metric} for comparing the quality of the DSM generated from our pipeline against other pipelines. We use the ground-truth LiDAR over San Fernando \cite{bosch2016multiple_mvs3d} to calculate the accuracy of our DSM as shown in \cref{fig:pipeline_comparison_main}. The accuracy metrics (defined by \cite{bosch2017metric}) indicate that our pipeline's performance is similar to other pipelines. 
  This means that other pipelines could be used to generate aligned cameras and DSMs, after which \textbf{\textit{depthifypp}} would create SatDepth Maps, allowing researchers to extend the dataset as more satellite images become publicly available.

\begin{figure}[!h]
    \begin{minipage}{0.22\linewidth}      
    \centering
    \scriptsize{
        \textbf{GCPs Collected}
        }
  \end{minipage}
  %
  \begin{minipage}{0.22\linewidth}
    \centering
    \scriptsize{
        \textbf{Annotation}
        }
  \end{minipage}
  %
  \begin{minipage}{0.45\linewidth}
    \centering
    \scriptsize{
        \textbf{Accuracy Summary using GCPs}
        }
  \end{minipage}
  \hfill
  %
  \begin{minipage}{0.22\linewidth}
  \centering
    \includegraphics[width=\linewidth, height=\linewidth]{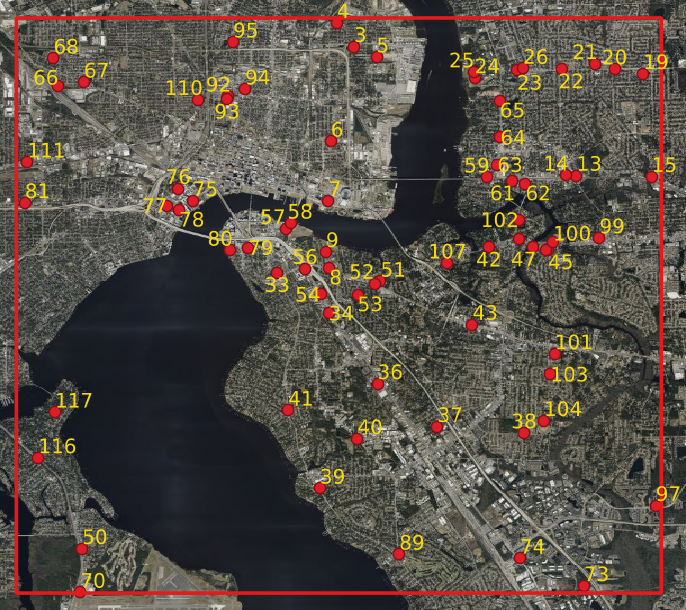}
  \end{minipage}
  \hspace{1mm}
  \begin{minipage}{0.22\linewidth}
   \centering
      \includegraphics[width=\linewidth, height=\linewidth]{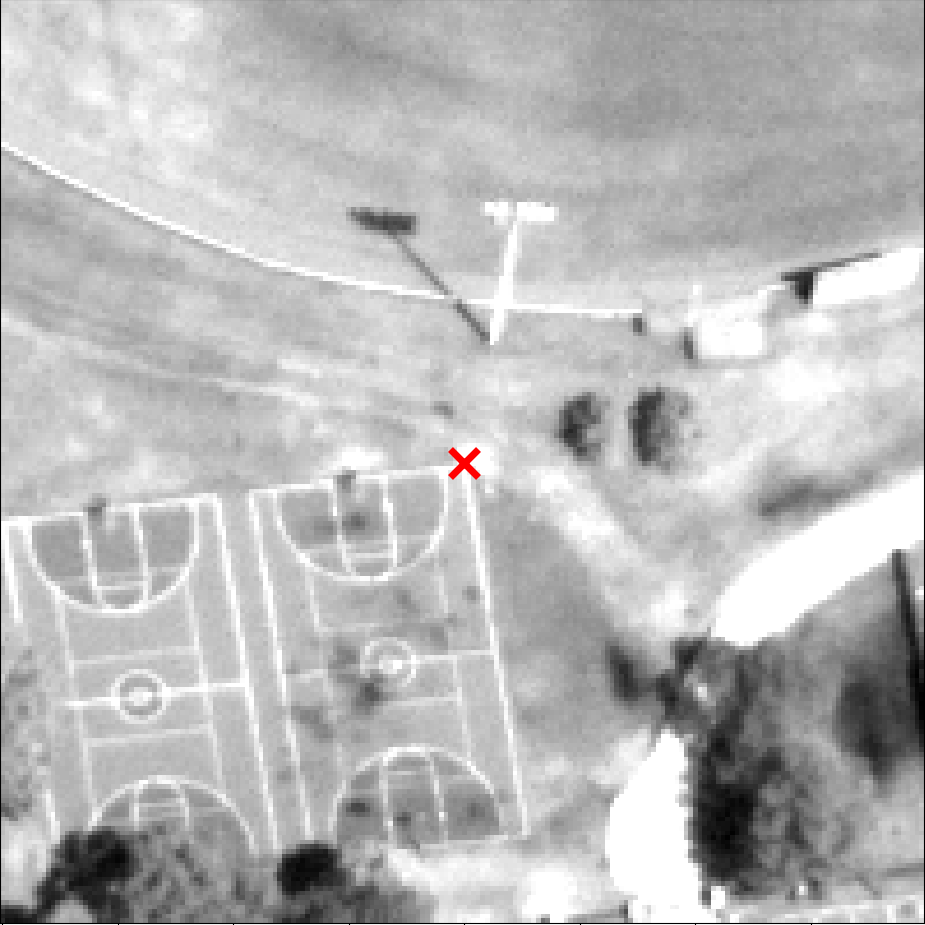}
  %
  %
  \end{minipage}
\hspace{2mm}
\begin{minipage}{0.45\linewidth}
  \centering
  \centering
  \resizebox{1\linewidth}{!}{%
  \setlength\tabcolsep{2.0pt}
    \begin{tabular}{lccc}
    \toprule
    \textbf{Error}     & \textbf{Mean} & \textbf{Std-Dev} & \textbf{Median} \\
    \midrule
    $\epsilon^{a}_{3D}$ (m)
    & 1.44          & 1.43             & 1.31   \\
    $\epsilon^{r}_{3D}$ (m)
    & 0.95          & 1.50             & 0.72   \\
    $\epsilon^{r}_{2D}$ (pxl)
    & 2.22          & 1.90             & 2.06  \\
    \bottomrule
    \end{tabular}
  }
\end{minipage}
\hfill
  \caption{GCP error summary for Jacksonville AOI. A total of 76 GCPs were annotated on satellite images to measure accuracy.}
  \label{fig:gcp_error_summary_main}
  \end{figure}

  \noindent \textbf{SatDepth Map Accuracy using GCPs}: To assess the accuracy of our SatDepth maps, we use Ground Control Points (GCPs) made available by \cite{accenture_gcps}. These GCPs were collected manually by surveyors who visited the sites in person and measured their 3D world coordinates ($\boldsymbol{X}^{GCP}$) using their surveying instruments. 
  We used a total of 76 GCPs spanning 50 tiles of our Jacksonville AOI, resulting in roughly 2000 annotations in the satellite images as shown in \cref{fig:gcp_error_summary_main}.
  We then compute three types of errors: (1) \textbf{Absolute 3D Error} $\epsilon^{a}_{3D}\defeq \norm{\boldsymbol{X}_i - \boldsymbol{X}^{GCP}}_2$ where $\boldsymbol{X}_i$ is read from SatDepth Maps of $I_i$ at annotated pixel $\boldsymbol{x}_i$. We compute this error for each GCP and all images; (2) \textbf{Relative 3D Error} $\epsilon^{r}_{3D}\defeq \norm{\boldsymbol{X}_i - \boldsymbol{X}_j}_2$ where ($\boldsymbol{X}_i,\, \boldsymbol{X}_j$) are read from SatDepth Maps of ($I_i,\, I_j$) at the annotated pixels ($\boldsymbol{x}_i,\,\boldsymbol{x}_j $) for the same GCP. We compute this error for all GCPs and all image pairs; and (3) \textbf{Relative 2D Error} $\epsilon^{r}_{2D}\defeq \norm{\boldsymbol{x}_j - \mathcal{P}_j(\boldsymbol{X}_i)}_2$ where $\boldsymbol{x}_j$ is the annotated pixel in $I_j$ and $\boldsymbol{X}_i$ is read from SatDepth Maps of $I_i$ at the annotated pixel $\boldsymbol{x}_i$. We compute this error for all GCPs and all image pairs. 
  %
  We show a summary of the residual errors in \cref{fig:gcp_error_summary_main}.

  Our absolute 3D error is under 1.5 m which agrees with such errors reported in \cite{nasa_ames}. The relative 3D error is under 1 m which indicates that world coordinates associated with individual pixels in SatDepth Maps at corresponding pixels are very close. The relative 2D error gives us a measure of the quality of ground-truth correspondences that can be extracted from SatDepth dataset. This error is around 2 pixels, which is acceptable considering that it is highly sensitive to annotation and quantization errors, potentially adding up to 1-2 pixels for a pair of annotations ($\boldsymbol{x}_i,\, \boldsymbol{x}_j$). We present further discussion in supplementary material.

  \section{Model Benchmarking}\label{sec:dataset_benchmarking}
  We benchmark our dataset using four image matching models: \textit{satMatchFormer}\cite{wang2022matchformer}, \textit{satLoFTR}\cite{Sun_CVPR21_LOFTR}, \textit{satDualRC-Net}\cite{Li_NIPS20_DUALRC} and SIFT + \textit{satCAPS}\cite{Qianqian_ECCV20_CAPS}. These models were originally trained on ground-based images from MegaDepth. However, we train them from scratch using SatDepth and rename them by prepending ``sat'' to the model name. The benchmarking details are as follows:
  
  \vspace*{1ex}
  
  \noindent \textbf{Training and Testing workflow}: 
  Each 1 km \( \times \) 1 km tile  satellite image 
  can be of size \( 4000 \times 4000 \) (\cref{sec:preprocessing}), which is impractical for training deep-learning networks. Thus, we work with smaller \( p \times p \) image patches. 
  Our training workflow begins by extracting pairs of image patches $(I_i, I_j)$ by projecting random 3D points, as defined by their (\textit{lat, lon, h}) values in the DSM, into the individual images where they serve as patch centers. We then compute the affine cameras for these patches and use them, along with SatDepth Maps, to extract a set of ground-truth correspondences \( \{\boldsymbol{x}^{k}_{i} \leftrightarrow \boldsymbol{x}^{k}_{j}\} \). The image patches are fed to the network and the ground-truth correspondences are used for supervision. The deep-learning based matcher predicts a set of correspondences between the patch pairs, and a loss is computed by comparing the set of predicted correspondences with the set of ground-truth correspondences. This loss is backpropagated through the network to update the weights, and the process is repeated 
   till the model training and validation losses converge.
  Subsequently, during testing, we generate a uniform grid of 3D points, project them onto an image pair to obtain the image patches, extract matches using the trained network, and then join the sets of predicted matches to form the final set of matches for the image pair. This workflow is used for all four benchmarked models, with minor deviations due to the different forms of supervision employed by each network. 

\begin{figure}[!h]
    \centering
    \includegraphics[width=\linewidth]{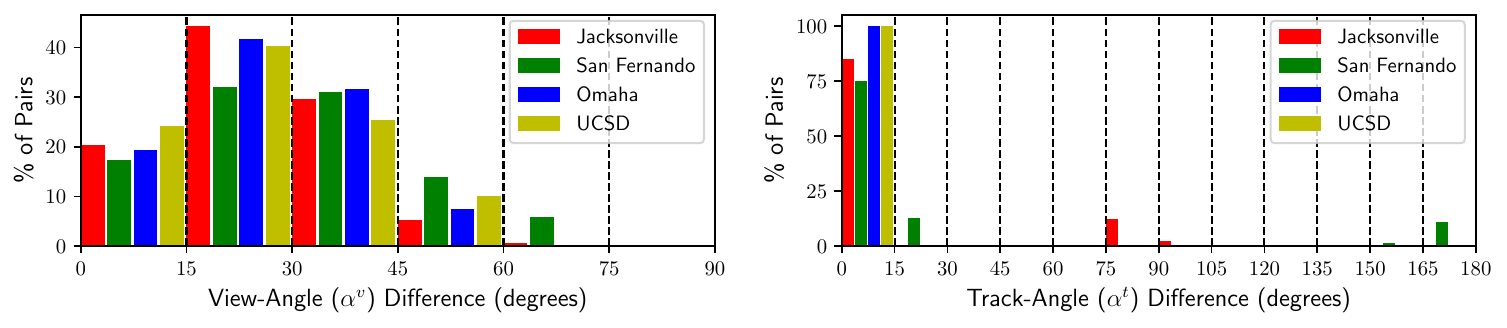}
  \caption{Image Pair distribution w.r.t. $\alpha^v$ and $\alpha^t$ for all SatDepth AOIs with 15\degree ~bins (dashed black lines). Image pair distribution w.r.t. $\alpha^t$ is highly sparse and imbalanced.}
  \label{fig:pair_distribution_main}
    \end{figure}

  \vspace*{1ex}
  \noindent \textbf{Dataset Splits and Balancing}: We use the Jacksonville AOI for training and reserve the other regions for testing. Within Jacksonville, we further split all available (124) non-water tiles into three groups: training (99), validation (11) and testing (14). The validation and testing tiles were randomly chosen to ensure uniform distribution across all available tiles. When using all possible image pairs for training, we face the challenge of highly imbalanced data due to the non-uniform distribution of view-angle difference ($\alpha^{v}$) and track-angle difference ($\alpha^t$) among image pairs as shown in \cref{fig:pair_distribution_main}. 
  The track angle difference refers to the angle between the two tracks on the ground plane corresponding to the satellites that recorded the two images.  On the other hand, the view-angle difference refers to the view angles of the two satellites with respect to the nadir (top-down) view.

  To address the imbalance stemming from $\alpha^{v}$, we implement a strategy of uniformly sampling pairs from a histogram as explained in supplementary material. Furthermore, to mitigate the imbalance induced by $\alpha^t$ angles, we utilize our novel rotation augmentation as explained below.

  \noindent \textbf{Rotation Augmentation}: To address the severe imbalance related to the track-angle difference, we employ rotation augmentation during training. Our objective with rotation augmentation is to simulate the rotation of the affine approximation  of the satellite camera about its camera viewing axis. This process involves transforming the image, its corresponding SatDepth maps, and the affine camera. We introduce a novel \textit{``crop-rotate-crop''} procedure for rotation augmentation, which entails cropping, rotating, and then cropping the images and the corresponding SatDepth maps, while also accounting for the camera transformation as shown in \cref{fig:rot_aug_main}. Further details are provided in the supplementary material.
  
  \vspace*{-1ex}
  \begin{figure}[h]
    \centering
    \begin{minipage}{0.45\linewidth}
      \centering
      \scriptsize{
          \textbf{``crop-rotate-crop'' procedure}
          }
      \end{minipage}
    \hspace*{1mm}
    \begin{minipage}{0.45\linewidth}
      \centering
      \scriptsize{
          \textbf{45\degree rotation}
          }
          ($\epsilon_{3D}^{r}=0.27$ m)
      \end{minipage}
      \hfill
      \vspace*{1mm}
    \begin{minipage}{0.45\linewidth}
      \centering
      \includegraphics[width=\linewidth]{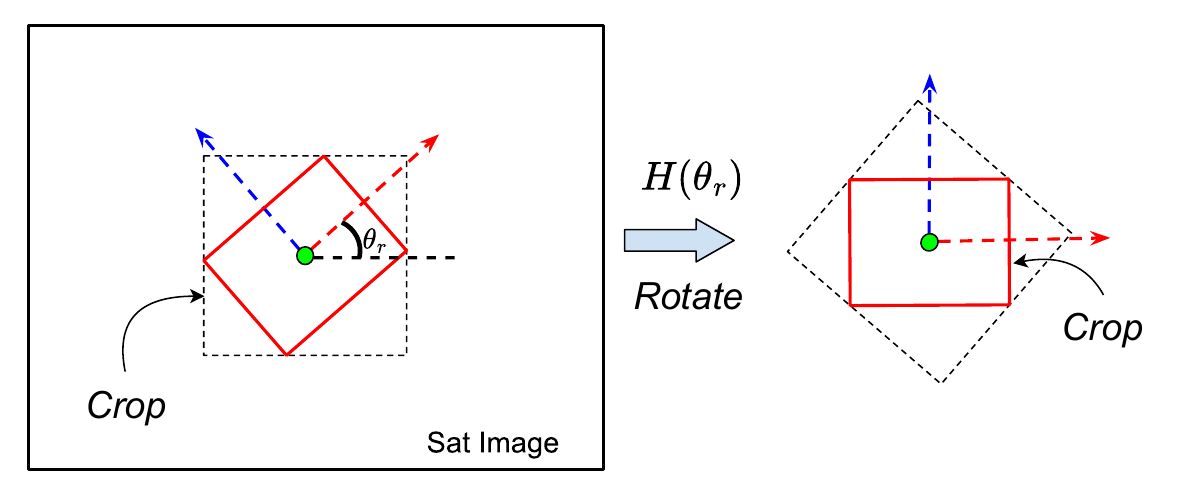}
    \end{minipage}
    \hspace*{1mm}
  \begin{minipage}{0.45\linewidth}
    \centering
    \includegraphics[width=\linewidth]{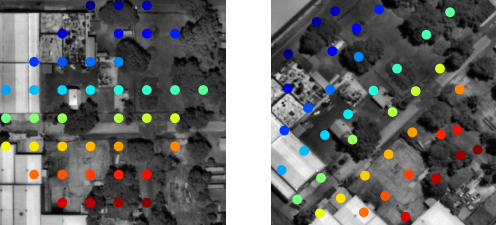}
  \end{minipage}
  \hfill
  \caption{\textbf{Left}: 
  Given a patch center ({\color{green}green} dot) and rotation angle ($\theta_r$), we create a rotated window ({\color{red}red} box) centered at the patch, crop the image, rotate it using $\theta_r$, and crop the new image. \textbf{Right}: Rot Aug example with colored point correspondences.}%
  \label{fig:rot_aug_main}
  \end{figure}

  \noindent \textbf{Evaluation Metrics}: We assess the accuracy of matches using the following metrics:
  \begin{enumerate}[itemsep=1ex]
    \item \textbf{Precision of Matches}: This is the proportion of correctly detected matches. We report precision using the symmetric epipolar distance $d_{epi}$ with threshold $\delta_{epi}$, similar to \cite{Sun_CVPR21_LOFTR}. Precision is computed for patches for the top ($K=200$) matches and for the whole image.
    \item \textbf{Pose Estimation Errors}: We follow the approach of \cite{Sun_CVPR21_LOFTR,MagicLeap_CVPR20_SuperGlue} to evaluate pose estimation errors. For pinhole cameras, the pose estimation error is defined as the maximum of angular error in rotation and translation. For SatDepth, we use the affine camera motion parameters: cyclotorsion ($\widehat{\theta}$), out-of-plane rotation ($\widehat{\phi}$), and scaling ($\widehat{s}$). We compute these motion parameters using the affine fundamental matrix $\widehat{F}$ \cite{hartley2003multiple}. We report the area under the cumulative curve (AUC) of the affine pose error as the maximum of angular error in ($\widehat{\theta},\,\widehat{\phi}$) for multiple thresholds. %
  \end{enumerate}
  
  \vspace*{1ex}
  \noindent \textbf{Implementation details}: For fair comparison, we trained the four models with a patch size of $p=448$ for 30 epochs without any hyperparameter tuning. 
  Since DualRC-Net and SIFT + CAPS were trained with RGB images, we adapted them to grayscale PAN band images without changing the architecture by repeating the grayscale input three times.
  The models were trained on 2 RTX A5000 GPUs, taking a total of 25 days. For more details, refer to the supplementary material.
  
  \section{Results}\label{sec:results}
\begin{figure*}
  \centering
  \input{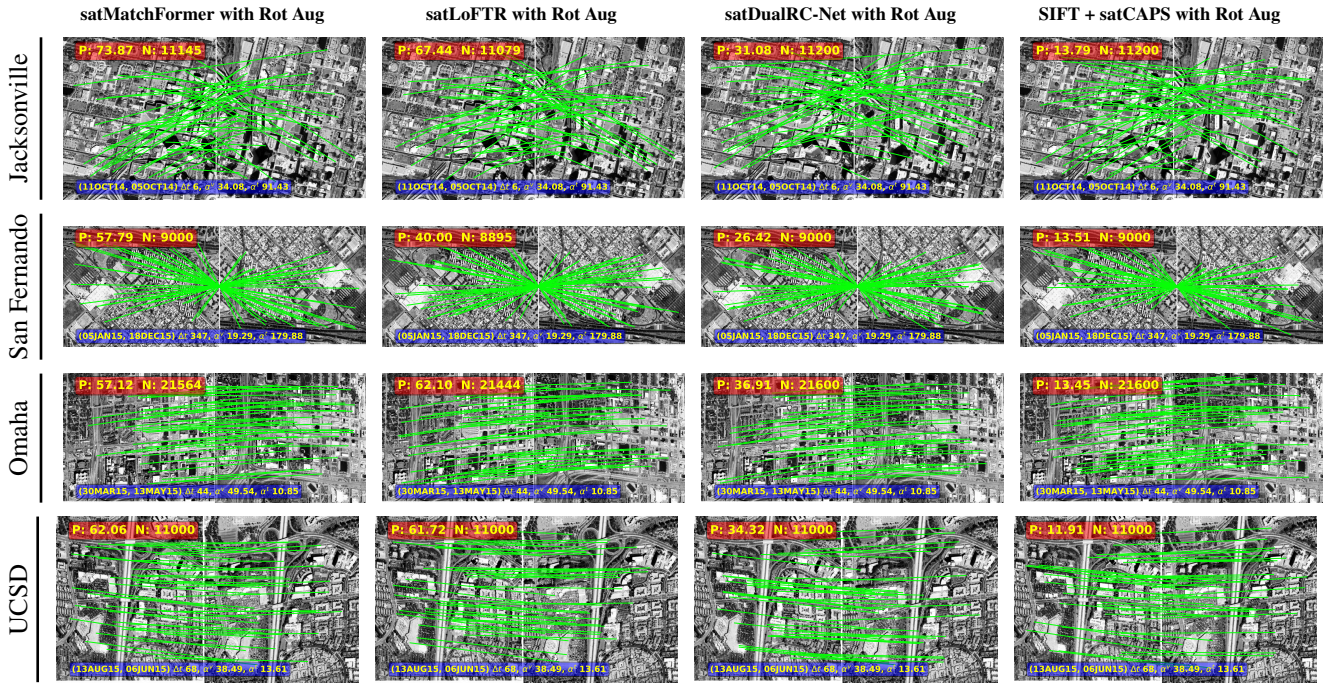}
\caption{Qualitative results for \textbf{large track-angle difference} ($\alpha^t$) for models trained on SatDepth with rotation augmentation. Precision (P) and number of matches (N) are displayed at the top of each plot. Image pair names, time difference ($\Delta t$), view-angle difference ($\alpha^v$), and track-angle difference ($\alpha^t$) are displayed at the bottom. The \flgreen{green} lines depict 40 randomly chosen \flgreen{true} matches.}
\label{fig:qualitative_matching_plots_rel_track}
\end{figure*}

  We want to emphasize that our goal in model benchmarking is not to achieve the best performance but to demonstrate that SatDepth can be used for training the four image matching networks \cite{wang2022matchformer, Sun_CVPR21_LOFTR,Li_NIPS20_DUALRC,Qianqian_ECCV20_CAPS} and to show the effectiveness of rotation augmentation, 
  which, 
  as our work shows, is absolutely required for satellite images. 
  We conducted an ablation study for three model configurations: with rotation augmentation, without rotation augmentation, and baseline. The baseline models
  were pre-trained on MegaDepth\cite{MegaDepthLi18} without rotation augmentation. When reporting averaged results, to account for the imbalance in the dataset 
  with respect to the track angles associated with the satellite images, 
  we use inverse of angular-bin counts as weights for computing the average. A summary of quantitative and qualitative comparison is presented in \cref{tab:matcher_performance_main} and \cref{fig:qualitative_matching_plots_rel_track} respectively. 

  \begin{table}[h]
    \centering
    \resizebox{\linewidth}{!}{%
    \setlength\tabcolsep{1.5pt}
  \begin{tabular}{cccccccc}
    \toprule
    \multicolumn{8}{c}{\centering \textbf{Jacksonville} | \textbf{San Fernando}}\\
    \midrule
    \multirow{2}{*}{\rotatebox[origin=c]{90}{Config}}
    &\multirow{2}{*}[-.4em]{Method} 
    & \multirow{2}{*}[-.4em]{\begin{tabular}[c]{@{}c@{}}Rot.\\ Aug.\end{tabular}}
    & \multicolumn{3}{c}{Pose estimation AUC $\uparrow$} 
    & \multirow{2}{*}[-.4em]{\begin{tabular}[c]{@{}c@{}}Precision $\uparrow$\\ \end{tabular}} 
    & \multirow{2}{*}[-.4em]{\begin{tabular}[c]{@{}c@{}} \# Matches $\uparrow$\\ ({\scriptsize{TP}})\end{tabular}}\\
    \cmidrule(lr){4-6}
    & & &@5\degree & @10\degree & @20\degree & &\\
    \midrule
    \multirow{4}{*}{\rotatebox[origin=c]{90}{Baseline}} 
    & \small{SIFT + CAPS}\cite{Qianqian_ECCV20_CAPS}
    & \cross & 37.04 | 36.08 & 40.06 | 38.65 & 44.36 | 42.90 & 2.02 | 1.81 & 4 | 4 \\
    &\small{DualRC-Net}\cite{Li_NIPS20_DUALRC}
    & \cross & 36.18 | 37.07 & 38.47 | 40.54 & 41.90 | 46.10 & 4.41 | 4.00 & 9 | 8 \\
    &\small{LoFTR}\cite{Sun_CVPR21_LOFTR}
    & \cross & 49.35 | 45.78 & 54.09 | 50.75 & 57.84 | 54.93 & 7.45 | 6.00 & 12 | 9 \\
    &\small{MatchFormer}\cite{wang2022matchformer}
    & \cross & 53.08 | 48.93 & 57.44 | 54.19 & 60.65 | 58.78 & 8.60 | 7.15 & 16 | 13\\
    \midrule
    \multirow{8}{*}{\rotatebox[origin=c]{90}{Trained on SatDepth}}  
    & \small{SIFT + satCAPS} & \cross & 39.61 | 36.90 & 44.37 | 40.19 & 50.27 | 45.27 & 3.21 | 2.14 & 6 | 4 \\
    &\small{satDualRC-Net} & \cross & 37.06 | 37.81 & 39.84 | 41.82 & 43.68 | 47.88 & 5.00 | 4.45 & 10 | 9 \\
    &\small{satLoFTR} & \cross & 84.29 | 37.89 & 90.63 | 40.89 & 94.46 | 44.51 & 64.87 | 4.23 & 129 | 6 \\
    &\small{satMatchFormer} & \cross & \textbf{86.15} | 46.50 & \textbf{92.05} | 52.35 & \textbf{95.43} | 58.08 & \textbf{69.44} | 4.39 & \textbf{139} | 7 \\
    \cmidrule{2-8}
    & \small{SIFT + satCAPS} & \checkmark & 38.49 | 36.75 & 43.26 | 40.09 & 50.94 | 45.69 & 10.67 | 7.89 & 21 | 16 \\
    &\small{satDualRC-Net} & \checkmark & 41.19 | 40.57 & 47.57 | 46.77 & 56.31 | 55.58 & 19.94 | 15.88 & 40 | 32 \\
    &\small{satLoFTR} & \checkmark & 78.48 | 53.60 & 87.02 | 62.96 & 92.30 | 71.34 & 54.87 | \textbf{42.58} & 108 | 71 \\
    &\small{satMatchFormer} & \checkmark & 81.37 | \textbf{54.57} & 89.01 | \textbf{64.15} & 93.56 | \textbf{72.68} & 61.96 | 39.83 & 124 | \textbf{73} \\
    \bottomrule
    \end{tabular}
    }
    \caption{Weighted average of Precision, Pose error, and number of True Positive (TP) matches over all testing image patches for Jacksonville and San Fernando AOIs.}
    \label{tab:matcher_performance_main}
  \end{table}

  \vspace*{-1em}
  \begin{figure}[h]
    \begin{subfigure}{0.48\linewidth}
      \centering
      \includegraphics[width=\linewidth]{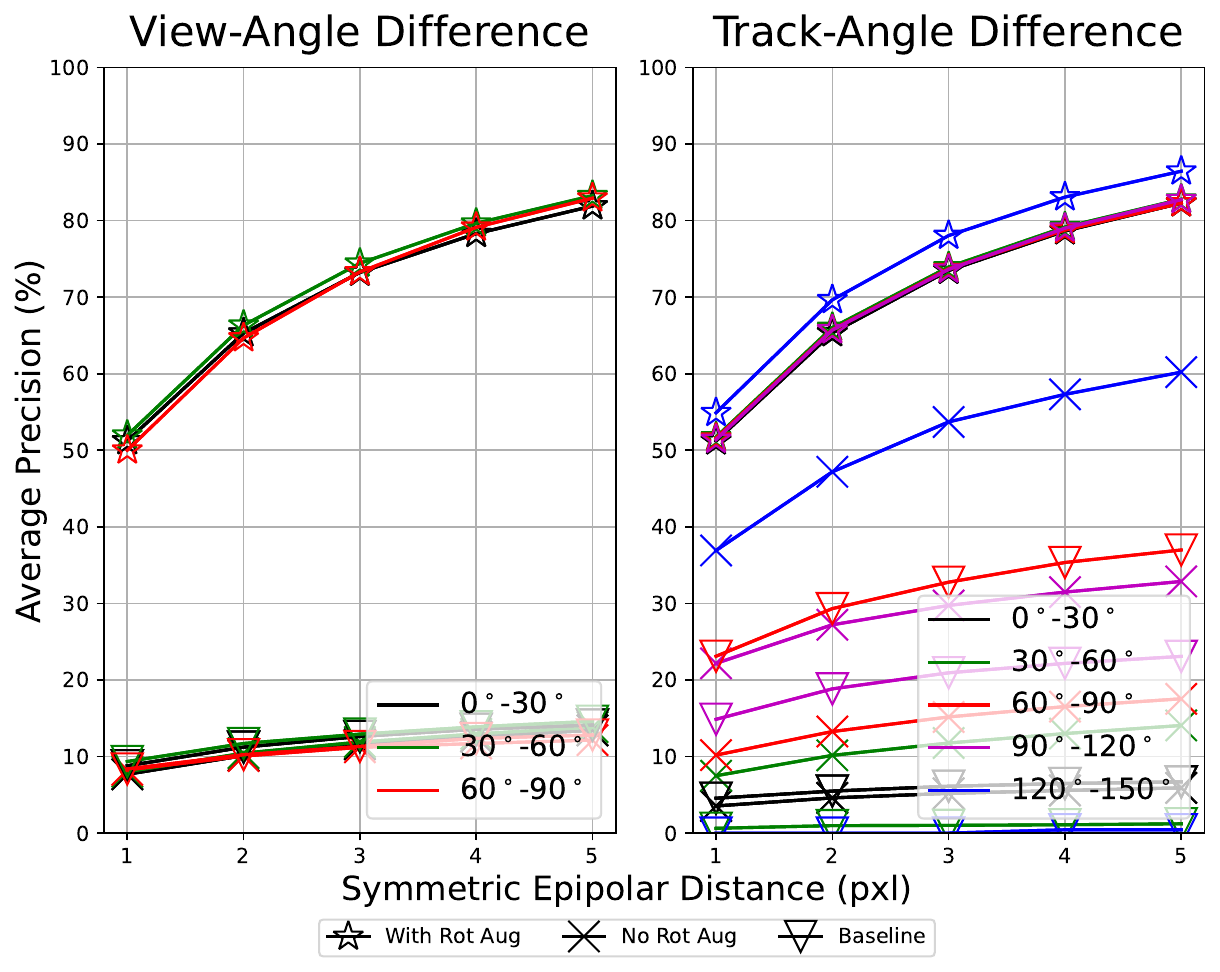}
      \caption{Jacksonville}
    \end{subfigure}
    \begin{subfigure}{0.48\linewidth}
      \centering
      \includegraphics[width=\linewidth]{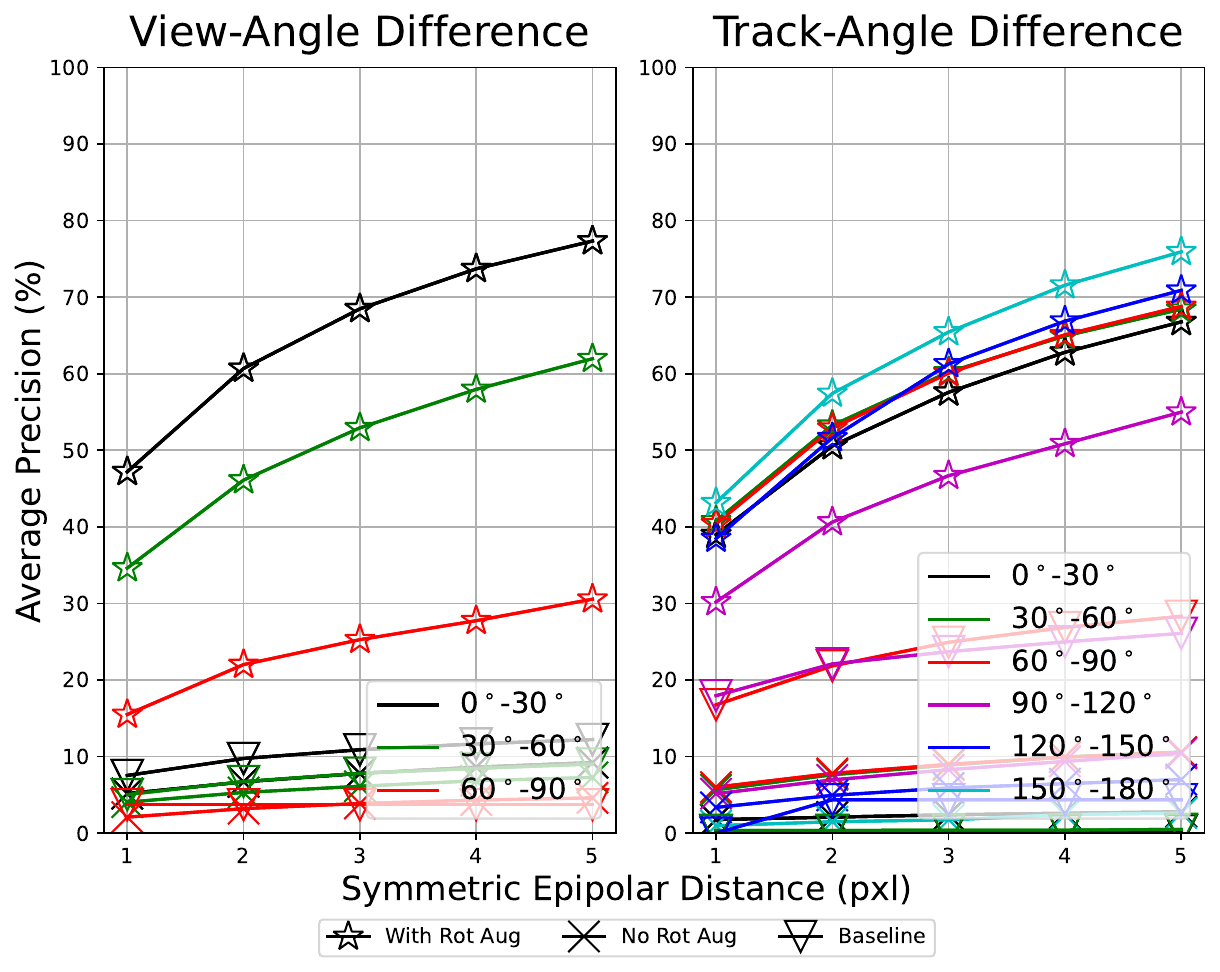}
      \caption{San Fernando}
    \end{subfigure}
    \hfill
  \caption{Average Precision for different configurations of LoFTR w.r.t. $\alpha^t$ and $\alpha^v$ for simulated rotation experiments.}
  \label{fig:matcher_performance_main}
\end{figure}

We draw a few key insights from \cref{tab:matcher_performance_main}: (1) Models trained on SatDepth perform better than the baseline models;
 (2) Transformer based models \cite{Sun_CVPR21_LOFTR,wang2022matchformer} learned better matches than the purely convolutional networks \cite{Li_NIPS20_DUALRC,Qianqian_ECCV20_CAPS} for the same number of training epochs; and 
 (3) For out-of-distribution testing (\ie San Fernando in \cref{tab:matcher_performance_main}), models trained on SatDepth with rotation augmentation perform the best but the same is not true for the testing set of Jacksonville. 
This anomaly occurs because the models trained without rotation augmentation overfits to the track-angle distribution in the Jacksonville training set. Consequently, it performs well on the Jacksonville testing set, which is spatially different but has the same angle distribution, while performing poorly on other AOIs due to the lack of generalization to different angle distributions. To substantiate this claim, we carry out experiments with simulated rotation during testing.

\vspace*{1ex}
\noindent \textbf{Simulated Rotation Experiment}: The goal of this experiment is to evaluate model generalization on unseen track-angle differences. To achieve this, we crop random patches from all testing pairs across all testing AOIs and randomly rotate the right image using our ``crop-rotate-crop'' procedure, then assess model performance for all configurations. \Cref{fig:matcher_performance_main} shows performance of LoFTR architecture over two AOIs as a function of $\alpha^v$ and $\alpha^t$ values. Due to simulated rotations, this setup produces a wider range of $\alpha^t$ values than the original distribution (\cref{fig:pair_distribution_main}). 

We draw the following insights from this experiment: (1) Models trained on SatDepth with rotation augmentation can generalize to unseen track-angle differences and to unseen AOIs; and (2) Training with rotation augmentation also helps improve performance w.r.t. view-angle differences, with gains up to 40\% for average precision. We provide further evidence in the supplementary material.

\section{Discussion}\label{sec:discussion}
In this paper, we introduced a novel dataset designed for training image matching networks specifically for satellite images. We presented our dataset generation pipeline for generating the SatDepth Maps and carried out extensive accuracy evaluation of our dataset, 
which enabled reliable learning of satellite image matching. 
To demonstrate the effectiveness of our dataset, we modified the training and evaluation protocols of four image matching models, which were initially designed for ground-based cameras, 
in order to tailor them specifically for satellite images.
Our experiments demonstrated that the models trained on SatDepth can generalize to unseen track-angles and to unseen AOIs.

We also designed a rotation augmentation procedure which allows for the discovery of corresponding pixels despite large rotational differences between the images. Handling such differences is crucial for areas where satellite coverage is sparse. Furthermore, when pooling together the images from different satellites it is highly likely that the image pairs can have a wider range of track angles, thus requiring the models to generalize well to unseen track angles.

A limitation of our work is that SatDepth Maps do not account for false matches caused by scene changes, such as construction or transient weather conditions. In future work, we plan to explore methods to detect these changes and create masks for them with minimal human input.

\ifthenelse{\equal{\useAck}{False}}{}{

\textbf{Acknowledgement}: This research is based upon work supported in part by
the Office of the Director of National Intelligence (ODNI),
Intelligence Advanced Research Projects Activity (IARPA),
via Contract \#2021-21040700001. The views and conclusions contained 
herein are those of the authors and should
not be interpreted as necessarily representing the official
policies, either expressed or implied, of ODNI, IARPA, or
the U.S. Government. The U.S. Government is authorized
to reproduce and distribute reprints for governmental purposes
notwithstanding any copyright annotation therein.
}


\clearpage  

\onecolumn  
\begin{center}
    \textbf{\LARGE Supplementary Material for \\ SatDepth: A Novel Dataset for Satellite Image Matching}\\[10pt]
    Rahul Deshmukh \quad Avinash Kak\\[1mm]
    {\tt\small deshmuk5@purdue.edu} \quad {\tt\small kak@purdue.edu}\\[1mm]
    Purdue University, West Lafayette
\end{center}

\bigskip
\hrule
\bigskip

\setcounter{tocdepth}{2}
\tableofcontents
\clearpage


\section{Overview}
In this supplementary material, we provide additional details related to SatDepth, including how to access the dataset, the dataset generation process, the accuracy assessment, the benchmarking experiments, and additional results.
In \cref{sec:sat_cam_appendix}, we present the satellite camera model. In \cref{sec:dataset_gen_appendix}, we provide further details related to the dataset generation process, including the selection of the geographic extent of the AOIs, the image alignment algorithm, and the depth map generation algorithm. We also cover the GCP annotation process and the accuracy assessment of the dataset using these GCPs. In \cref{sec:benchmarking_appendix}, we detail the benchmarking experiments conducted on the SatDepth dataset, including the dataset splits, the rotation augmentation procedure, and implementation details of the benchmarked models. In \cref{sec:additional_results_appendix}, we present additional results for the benchmarked models. Finally, in \cref{sec:datasheet}, we provide the datasheet for the SatDepth dataset.

\section{Satellite Camera Model}\label{sec:sat_cam_appendix}
Since the satellite camera model plays a critical role in the construction of the SatDepth dataset, this section provides further essential details regarding the camera model.
The RPC (Rational Polynomial Coefficient) camera model facilitates the mapping of a 3D world point $(\boldsymbol{X}\in\mathbb{R}^3)$ to a pixel location $(\boldsymbol{x}\in\mathbb{R}^2)$ in the image. This mapping is also known as the forward projection and is denoted by $\boldsymbol{x}=\mathcal{P}(\boldsymbol{X})$.
The RPC camera model for satellite images consists of 80 coefficients that form a third-order rational polynomial. 
Additionally, the RPC model has 5 scale parameters and 5 offset parameters for row, column, latitude, longitude, and height respectively.
The RPC model is shared in the standard RPC00B (RPB) format as per the NITF standard \cite{nitf_std}. 

\subsection{Forward Projection}
The RPC forward projection $\mathcal{P}: \mathbb{R}^3 \rightarrow \mathbb{R}^2 $ is defined as a mapping from world coordinates $\boldsymbol{X}=(lat,\, lon,\, h)$ to pixel coordinates $\boldsymbol{x}=(r,\,c)$ as shown in \cref{eq:rpc}.
\begin{align}
\boldsymbol{x} = \begin{bmatrix}r\\c\end{bmatrix} &= \mathcal{P}(lat,\,lon,\,h) = \mathcal{P}(\boldsymbol{X})\label{eq:rpc}
\end{align}

The RPC forward projection starts with the normalization of the world coordinates $(lat,\,lon,\, h)$ to $(\phi_n,\,\lambda_n,\, h_n)$ using the scale and offset parameters as shown in \cref{eq:rpc_norm_world}. The RPC forward projection function models the normalized pixel coordinates $(r_n,\,c_n)$ using a ratio of two third-order polynomials as shown in \cref{eq:rpc_line_samp}. The model consists of four polynomials: two for the numerator ($P^r_n\,,P^c_n$) and two for the denominator ($P^r_d\,,P^c_d$). The numerator and denominator polynomials are defined by the RPC coefficients and cubic polynomial terms as shown in \cref{eq:rpc_polys} and \cref{eq:rpc_poly_terms} respectively. Finally, the image row (line) and column (sample) coordinates are recovered by denormalizing the normalized pixel coordinates using the scale and offset parameters as shown in \cref{eq:rpc_denorm}. 

For convenience we group \cref{eq:rpc_norm_world,eq:rpc_denorm} into a single form and denote the RPC forward projection by $\mathcal{P}(.)$ as shown in \cref{eq:rpc}.

\begin{alignat}{3}
  \phi_n =\frac{lat-LAT\_OFF}{LAT\_SCALE} && \quad
  \lambda_n =\frac{lon-LONG\_OFF}{LONG\_SCALE} && \quad
  h_n =\frac{h-HEIGHT\_OFF}{HEIGHT\_SCALE} \label{eq:rpc_norm_world}
\end{alignat}

\begin{alignat}{2}
  r_n =\frac{P^r_{n}(\phi_n,\lambda_n,h_n)}{P^r_d(\phi_n,\lambda_n,h_n)} && \quad
  c_n =\frac{P^c_n(\phi_n,\lambda_n,h_n)}{P^c_d(\phi_n,\lambda_n,h_n)} \label{eq:rpc_line_samp}
\end{alignat}

\begin{equation} \label{eq:rpc_polys}
\begin{aligned}
  P^r_n(\phi_n,\lambda_n,h_n) &= \sum_{i=1}^{20} LINE\_NUM\_COEFF_i \cdot \rho_i(\phi_n,\lambda_n,h_n) \\
  P^r_d(\phi_n,\lambda_n,h_n) &= \sum_{i=1}^{20} LINE\_DEN\_COEFF_i \cdot \rho_i(\phi_n,\lambda_n,h_n) \\
  P^c_n(\phi_n,\lambda_n,h_n) &= \sum_{i=1}^{20} SAMP\_NUM\_COEFF_i \cdot \rho_i(\phi_n,\lambda_n,h_n) \\
  P^c_d(\phi_n,\lambda_n,h_n) &= \sum_{i=1}^{20} SAMP\_DEN\_COEFF_i \cdot \rho_i(\phi_n,\lambda_n,h_n) 
\end{aligned}
\end{equation}

\begin{equation} \label{eq:rpc_poly_terms}
\begin{aligned}
  \rho_1 &= 1 & \quad
  \rho_2 &= \lambda & \quad
  \rho_3 &= \phi & \quad
  \rho_4 &= h & \quad
  \rho_5 &= \lambda\phi & \quad \\
  \rho_6 &= \lambda h & \quad
  \rho_7 &= \phi h & \quad
  \rho_8 &= \lambda^2 & \quad
  \rho_9 &= \phi^2 & \quad
  \rho_{10} &= h^2 & \quad \\ 
  \rho_{11} &= \phi\lambda h & \quad
  \rho_{12} &= \lambda^3 & \quad
  \rho_{13} &= \lambda\phi^2 & \quad
  \rho_{14} &= \lambda h^2 & \quad
  \rho_{15} &= \lambda^2 \phi & \quad \\
  \rho_{16} &= \phi^3 & \quad
  \rho_{17} &= \phi h^2 & \quad
  \rho_{18} &= \lambda^2 h & \quad
  \rho_{19} &= \phi^2 h & \quad
  \rho_{20} &= h^3 \quad 
\end{aligned}
\end{equation}

\begin{alignat}{2}
  r = r_n*LINE\_SCALE + LINE\_OFF && \quad
  c = c_n*SAMP\_SCALE + SAMP\_OFF \label{eq:rpc_denorm}  
\end{alignat}

\subsection{Back Projection}
\begin{figure}[!h]
  \centering
  \includegraphics[width=0.2\linewidth]{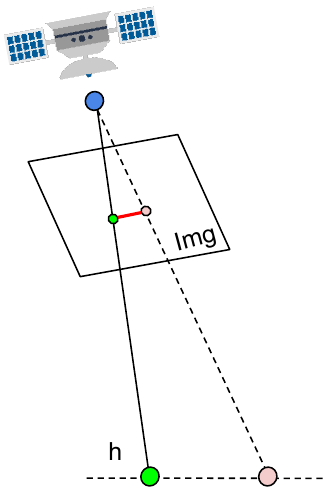}
  \caption{Given a pixel location $\boldsymbol{x}$ ({\color{green}green} dot in image plane) and the height $h$ (dashed horizontal line), we estimate the world point $\boldsymbol{X}$ (big {\color{green}green} dot) by minimizing the reprojection error (length of the {\color{red}red} line). The {\color{pink} pink} dots represent the iterative estimate of the world point (big {\color{pink} pink} dot) and its corresponding projection (small {\color{pink} pink} dot).}
  \label{fig:inv_rpc}
\end{figure}

The inverse mapping of the forward projection, known as the backprojection, results in a 3D ray joining the camera center and the pixel location ($\boldsymbol{x}$). Due to the nonlinear nature of $\mathcal{P}$, there is no closed-form solution for the inverse mapping. However, given the height ($h$) of a 3D point on this ray, we can compute the corresponding $(lat,\, lon)$ by minimization of the reprojection error as shown in  \cref{eq:inv_rpc}. The minimization process is shown pictorially in \cref{fig:inv_rpc}.

\begin{align}
  lat^*,\,lon^* = \mathcal{P}^{-1}(\boldsymbol{x},\,h) \defeq \argmin _{lat,\,lon} \norm{ \boldsymbol{x} - \mathcal{P}(lat,\, lon,\, h)}_{2}^{2}  \label{eq:inv_rpc}
\end{align}

\subsection{Affine Approximation of the RPC Camera Model}
On account of the large distance between a satellite and Earth, it is possible to approximate the RPC camera model with a computationally much more efficient local affine model for small-sized patches on the ground.

We compute the local affine approximation ($\widehat{\mathcal{P}}$) to RPC camera model ($\mathcal{P}$) centered around a given world point ($\boldsymbol{X}_0$) using the first order Taylor series expansion of the RPC model as shown in \cref{eq:rpc_affine}. 

\begin{equation} \label{eq:rpc_affine}
\begin{aligned}
  \widehat{\mathcal{P}}(\boldsymbol{X})  &= \mathcal{P}(\boldsymbol{X}_0) + \nabla \mathcal{P}(\boldsymbol{X}_0) \cdot (\boldsymbol{X} - \boldsymbol{X}_0) \\
  &= \nabla \mathcal{P}(\boldsymbol{X}_0) \boldsymbol{X} + \boldsymbol{b} \\
  \widehat{\mathcal{P}}(\boldsymbol{X}) &= \begin{bmatrix}
  \nabla \mathcal{P}(\boldsymbol{X}_0) & \boldsymbol{b}\\
  \boldsymbol{0}^T & 1
  \end{bmatrix} \begin{bmatrix}
  \boldsymbol{X}\\
  1
  \end{bmatrix}
\end{aligned}
\end{equation}

where $\boldsymbol{b}=\mathcal{P}(\boldsymbol{X}_0) - \nabla \mathcal{P}(\boldsymbol{X}_0) \boldsymbol{X}_0$ is the bias term and $\nabla \mathcal{P}(\boldsymbol{X}_0)$ is the Jacobian of the RPC model evaluated at $\boldsymbol{X}_0$. Please note that, this bias term is different from the bias correction term computed during Bundle Adjustment (``Image Alignment''section of \cite{satdepth_preprint}).

\clearpage
\section{Dataset Generation}\label{sec:dataset_gen_appendix}
In this section we present further details about our dataset and the dataset generation pipeline which were not discussed in the main manuscript \cite{satdepth_preprint}.
It was not possible to supply this information 
in the main manuscript due to page limitations.

\subsection{Deciding the Geographic Extent of a SatDepth AOI}
\begin{figure}[!h]
  \centering
  \includegraphics[width=0.5\linewidth]{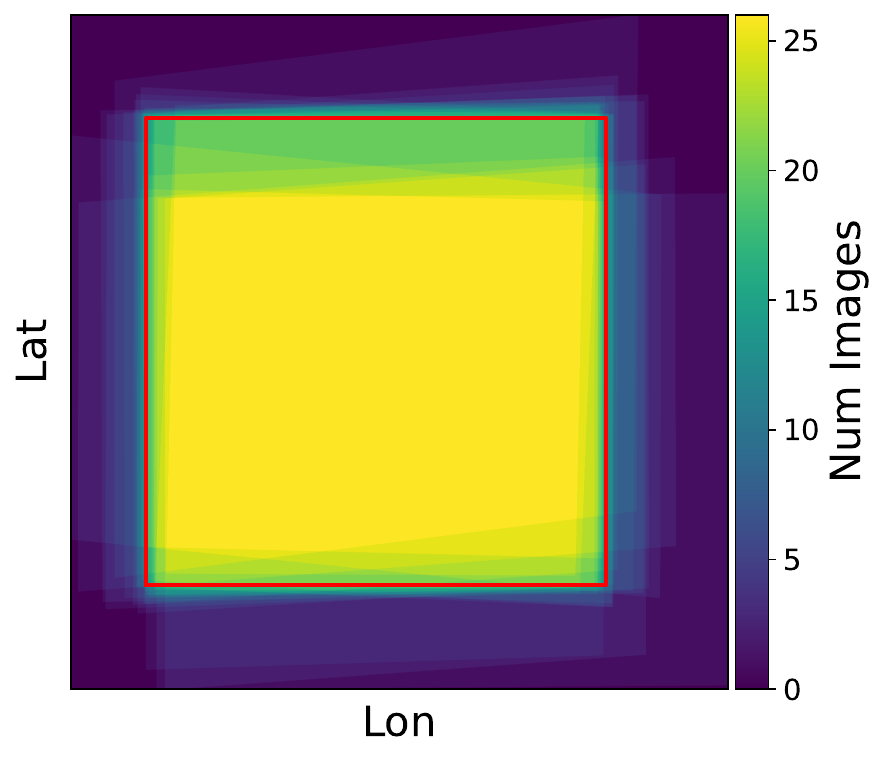}
  \caption{Jacksonville image coverage: the color intensity represents the image count at a given location and the {\color{red}red} box indicates the chosen extents of Jacksonville AOI. We can see that the coverage is sparse in the borders.}
  \label{fig:jax_img_heatmap}
\end{figure}

For constructing a training dataset for satellite image matching, it is important to take some care in choosing the AOI (Area of Interest) on the ground. What can make this problem somewhat difficult is the fact that satellite image coverage, especially for high-resolution satellites, can be highly non-uniform. The goal in choosing an area on the ground that would serve as an AOI should be that satellite coverage over all parts of the AOI are as uniform as they can be. To illustrate the non-uniformity in satellite coverage, we show a heatmap of image coverage for an area around Jacksonville in \cref{fig:jax_img_heatmap}. As one can see, the image coverage at the borders is sparse. 
For the Jacksonville AOI in our work, we chose the extent of the AOI as a rectangular area so that the image coverage over any of the tiles in the area did not drop below $N_{min}$. For Jacksonville, this resulted in an approximately 200 \sqkm AOI with $N_{min}=26$ (\ie all available images for Jacksonville). For other regions in the SatDepth dataset, we arbitrarily chose a small AOI such that the entire AOI was visible in all the images.

\begin{figure}[!h]
  \centering
  \includegraphics[width=\linewidth]{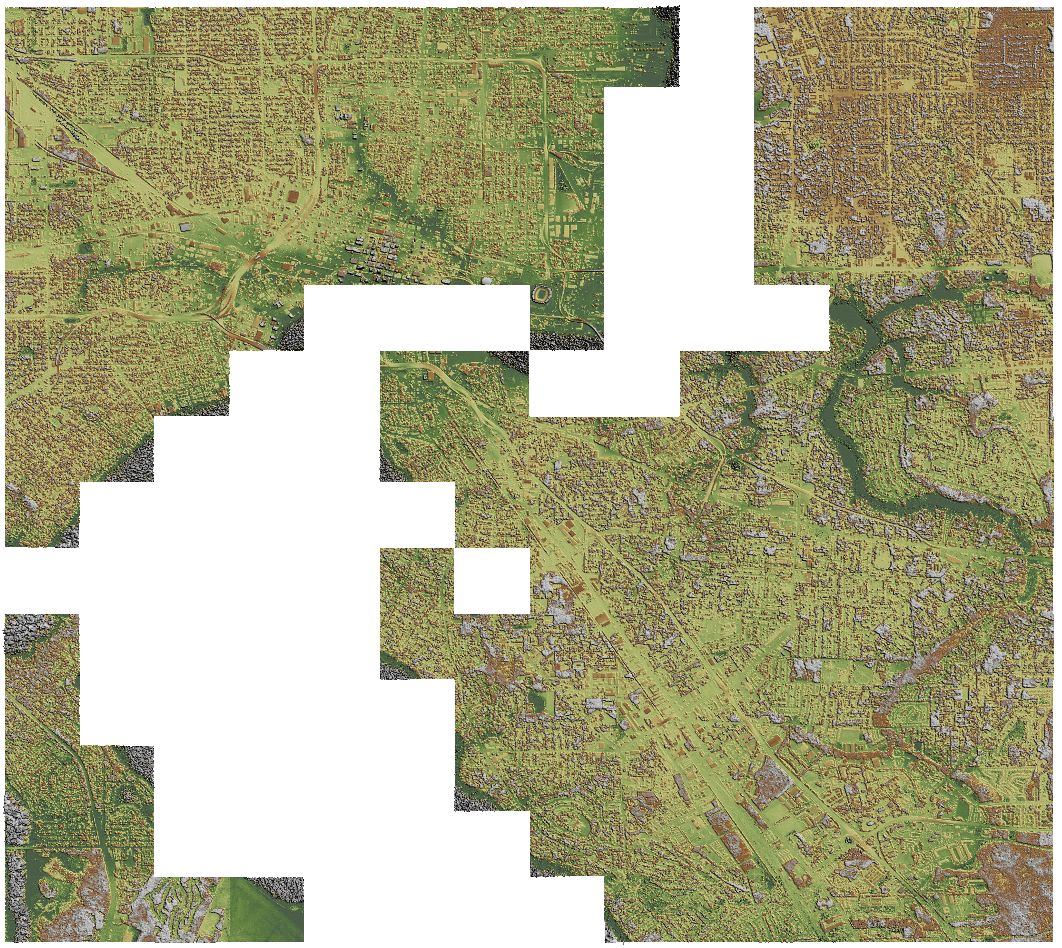}
  \caption{Composite DSM for Jacksonville AOI. The white area indicates tiles which were not processed due to water coverage or failure in bundle adjustment.}
  \label{fig:jax_dsm}
\end{figure}

\begin{figure}[!h]
  \centering
  \includegraphics[width=0.5\linewidth]{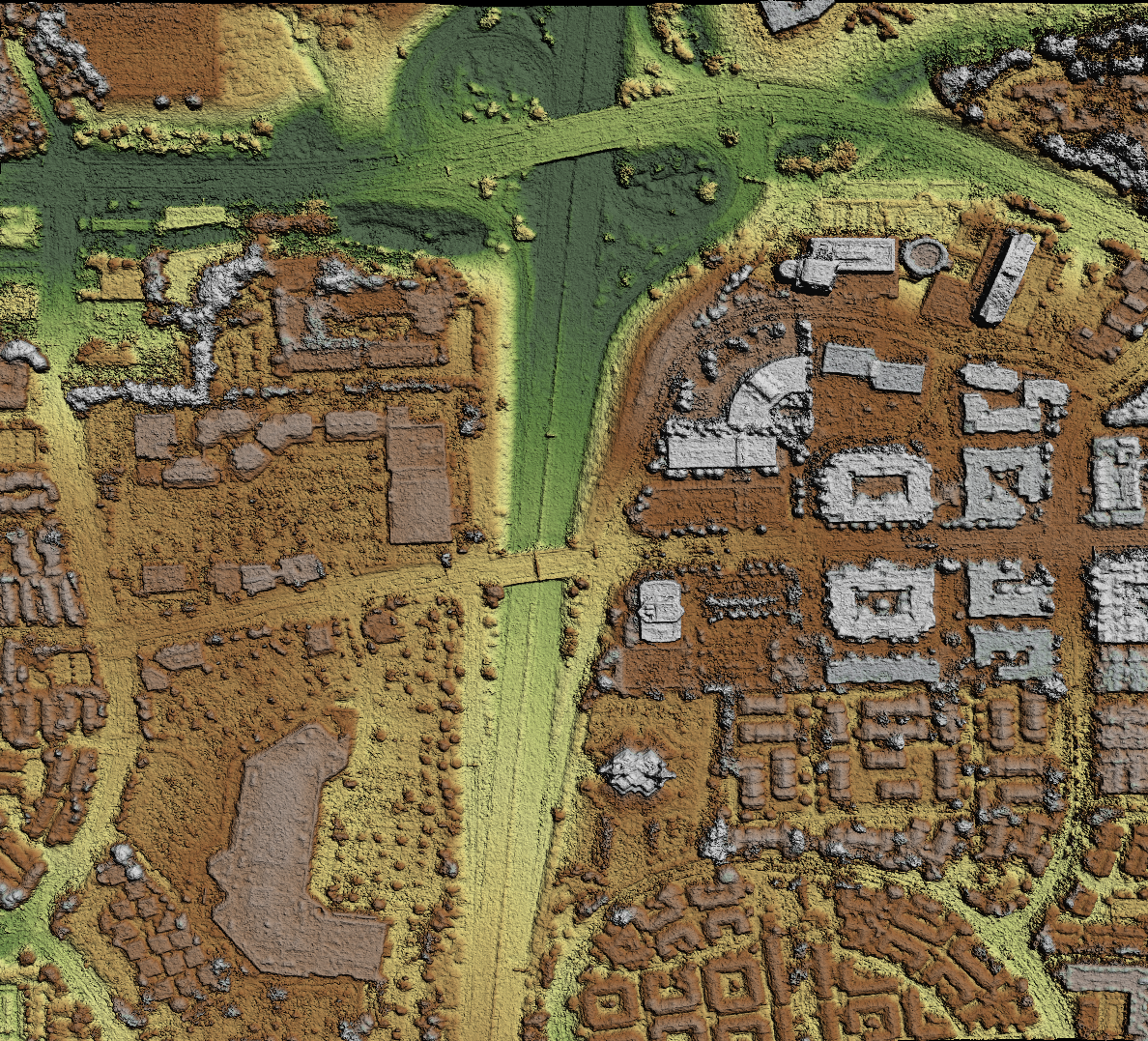}
  \caption{DSM for UCSD AOI}
  \label{fig:ucsd_dsm}
\end{figure}

For each region in SatDepth, we generate a high-resolution DSM with 0.25 m Ground sampling Distance (GSD) using our 3D reconstruction pipeline. We show the DSMs (as color plots) for each region in \cref{fig:jax_dsm,fig:ucsd_dsm}. For Jacksonville, we show the composite DSM, which is the stitched DSM for all tiles. Additionally, the Jacksonville AOI has the St. Johns river flowing through it (\cref{fig:jax_dataset_splits}) due to which several tiles have total water coverage. These water tiles were excluded from the DSM generation as they do not contain any important height information.

\begin{figure}[!h]
  \centering
  \includegraphics[width=0.5\linewidth]{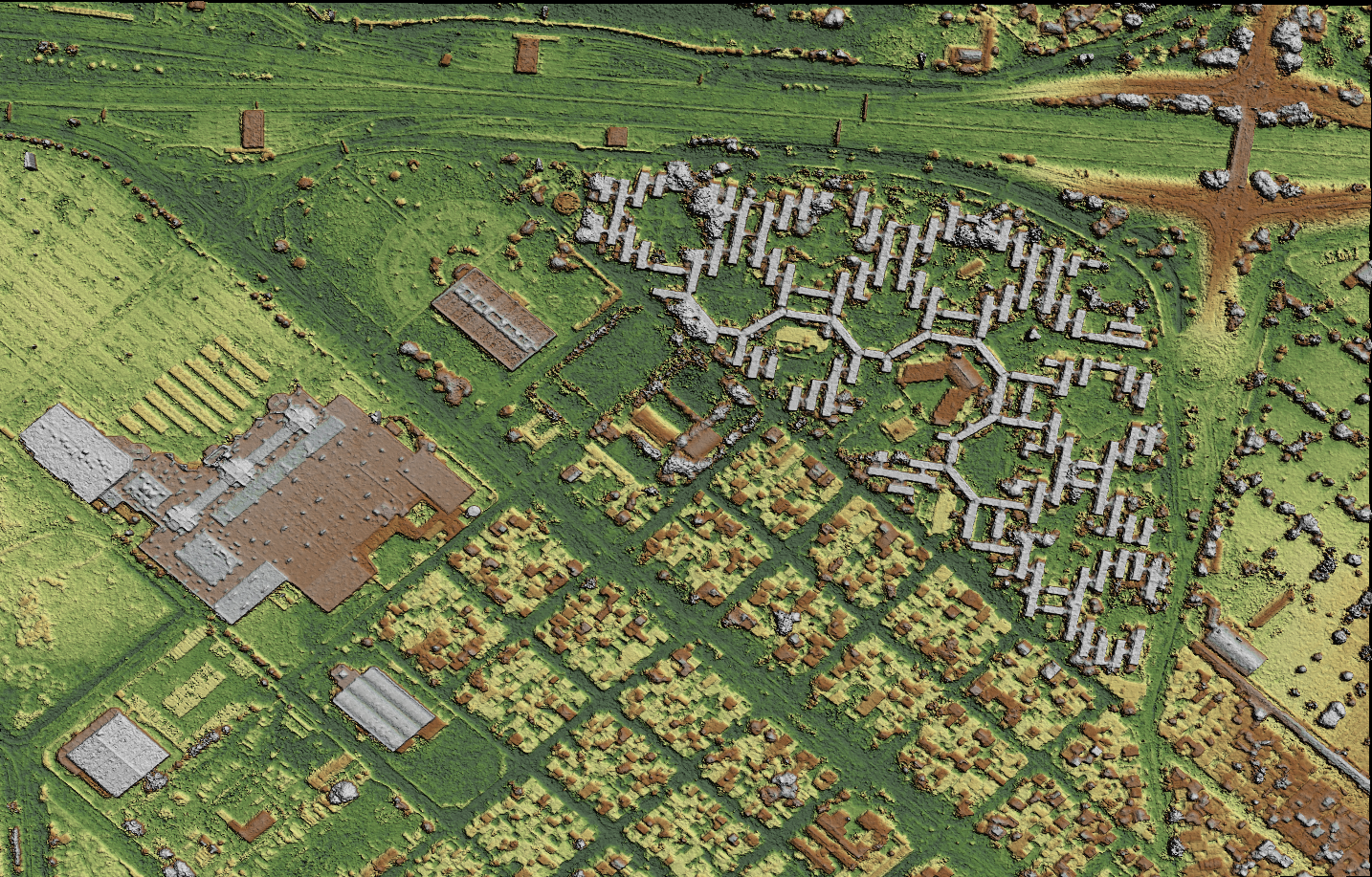}
  \caption{DSM for San Fernando AOI}
  \label{fig:argentina_dsm}
\end{figure}

\begin{figure}[!h]
  \centering
  \includegraphics[width=0.5\linewidth]{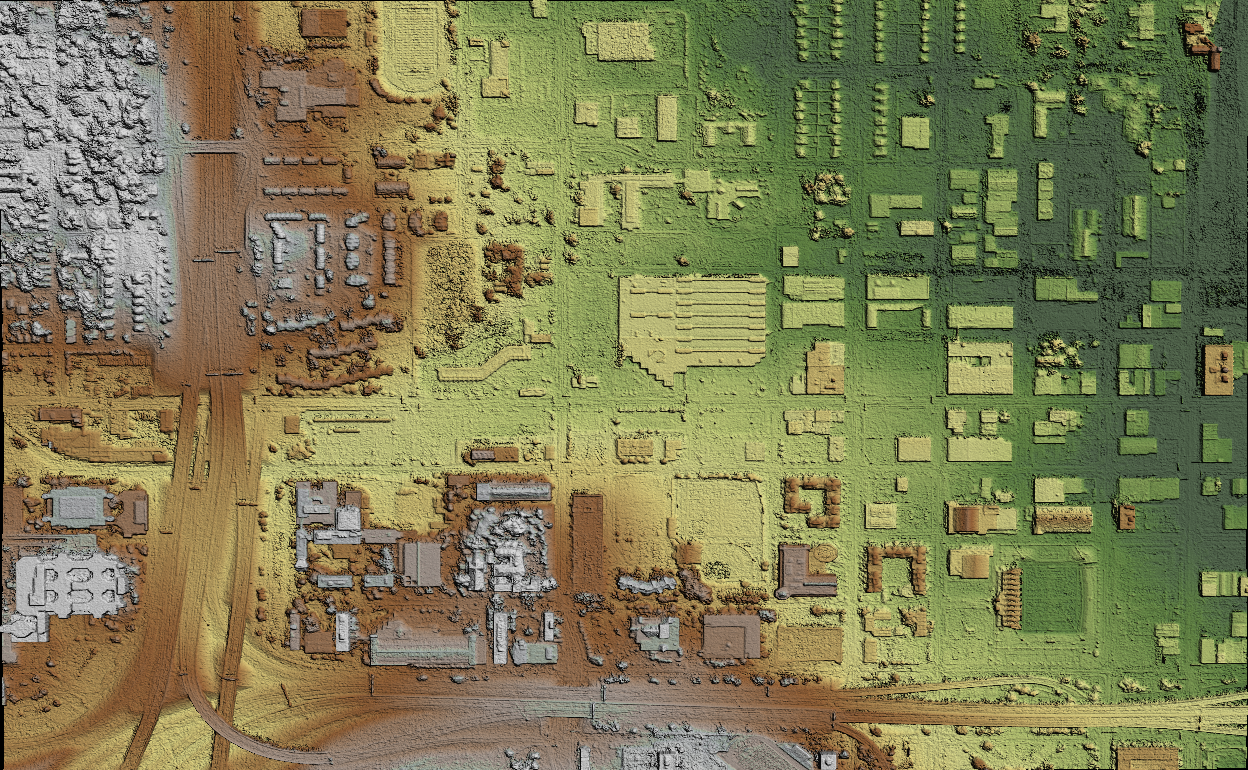}
  \caption{DSM for Omaha AOI}
  \label{fig:omaha_dsm}
\end{figure}

\clearpage
\subsection{Image Alignment}\label{sec:more_ba}
In the main manuscript, we explained how we carry out the Bundle Adjustment (BA). After BA, we compute a ``connectivity'' graph using the pairwise inlier correspondences for all the image pairs in a tile. These inlier correspondences are computed during BA. The images are represented as nodes in the ``connectivity'' graph, and an edge is created between two nodes when the number of inliers between them exceeds a threshold. Using the ``connectivity'' graph, we perform a connected components analysis, and the largest component is stored as the aligned set of images. This means that the largest component may not always include all of the initial set of images. Additionally, when the density of the largest connected component fails to meet a minimum threshold, we exclude the tile from further processing. Such tiles are declared as failures and omitted from the DSM generation process. 

In SatDepth, we provide RPCs and satellite images corresponding to the largest component for all successfully aligned tiles. There was only one tile in the Jacksonville AOI that failed to meet the density threshold and was excluded from the DSM generation process.

\subsection{Depth Map Generation}
\begin{figure}[!h]
  \centering
  \includegraphics[width=0.5\linewidth]{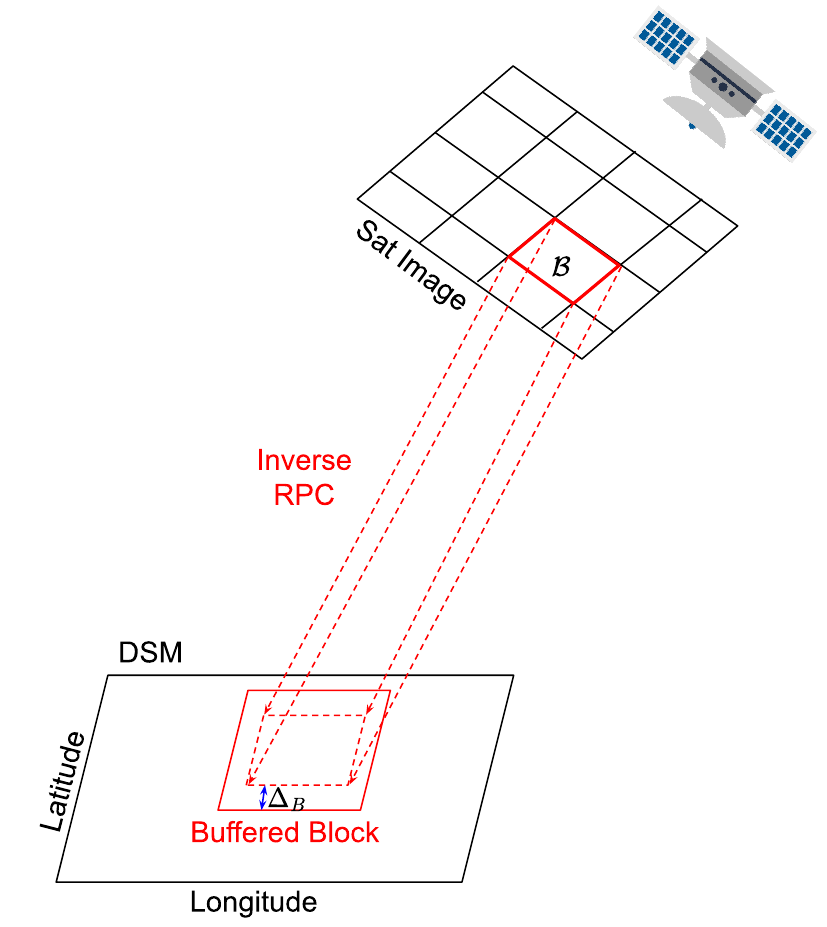}
  \caption{Tiling procedure for \textbf{depthifypp}}
  \label{fig:depthifypp_tiling}
\end{figure}

We gave a short description of \textbf{depthifypp} in the main manuscript and provide the sequential algorithm for it in Algorithm \cref{alg:seq_depth}. To generate SatDepth Maps, we use the ASTER Water Mask \cite{aster_water_mask} to avoid generating grids of points over bodies of water. Depth generation is carried out by internally tiling the large satellite image, as shown in \cref{fig:depthifypp_tiling}. We first create a small block, $\mathcal{B}$, in the satellite image and compute its corresponding world coordinates using $\mathcal{P}^{-1}$ and the DSM as the height source. A buffer ($\Delta_{B}$) is then added to these world coordinates, and depth map generation is performed for this buffered block, as shown in \cref{fig:depthifypp_tiling}. 
Since adjacent blocks might have a tall building visible in the satellite image due to the oblique viewing angle, but not in the DSM (DSM is a top-down view), this can cause discontinuities at the block boundaries.
The buffer is added to ensure that we account for such effects.
SatDepth Maps are computed independently for each block. Finally, we write the block-stitched SatDepth Map.

\begin{algorithm}[!h]
  \caption{Sequential algorithm for \textbf{depthifypp}}\label{alg:seq_depth}
  \begin{algorithmic}
  \Require Satellite Image ($Img$), Camera Model ($RPC$), Digital Surface Model ($DSM$), 
  Digital Elevation Model ($DEM$), DEM ortho camera ($DEM\_CAM$), Water Mask ($Water\_Mask$), Grid spacing for height ($\Delta_z$), and Output SatDepth Maps ($Lat, Lon, Ht$)
  \State
  \State $Lat \gets 0$ \, $Lon \gets 0$ \, $Ht \gets 0$
  \For{$dsm\_r=0$, \, $dsm\_r<ROW\_SIZE$, \, $dsm\_r++$ }
    \For{$dsm\_c=0$, \, $dsm\_c<COL\_SIZE$, \, $dsm\_c++$ }
    \State $lat, \, lon = DEM\_CAM.get\_latlon(dsm\_r, \, dsm\_c)$
    \State $isWater=Water\_Mask[lat,\, lon]$
    \If{$isWater$}
      \State Continue
    \EndIf
    \State $Z\_UB = DSM[dsm\_r, \, dsm\_c]$
    \State $dem\_r, \, dem\_c = DEM\_CAM.get\_rowcol(lat,\, lon)$
      \State $Z\_LB = DEM[dem\_r, \, dem\_c]$ \Comment{This involves interpolation}

        \For{$Z=Z\_LB$, \, $Z<Z\_UB$,\, $Z += \Delta_z$ }
      \State $Img\_r, \, Img\_c = RPC.get\_rowcol(lat, \, lon, \, Z)$
      \If{$0\leq Img\_r \leq Img\_H \quad \&\& \quad 0\leq Img\_c \leq Img\_W$ }
        \If{$Ht[Img\_r,\, Img\_c] < Z$}
        \State $Ht[Img\_r,\, Img\_c] = Z$    
        \State $Lat[Img\_r,\, Img\_c] = lat$    
        \State $Lon[Img\_r,\, Img\_c] = lon$    
        \EndIf
      \EndIf
        \EndFor
    \EndFor
  \EndFor
  \end{algorithmic}
\end{algorithm}

\clearpage
\subsection{Dataset Accuracy Assessment}
In this section we will give further details related to how GCPs were annotated and a short discussion on assessment of accuracy of the SatDepth dataset using these GCPs.
As mentioned earlier, a GCP is a Ground Control Point that has been identified by a surveyor.  GCPs comes with associated visual imagery for their identification in remotely sensed images. 

\subsubsection{GCP Annotation}\label{sec:gcp_annotation_appendix}
\begin{figure}[!h]
  \centering
  \includegraphics[width=0.7\linewidth]{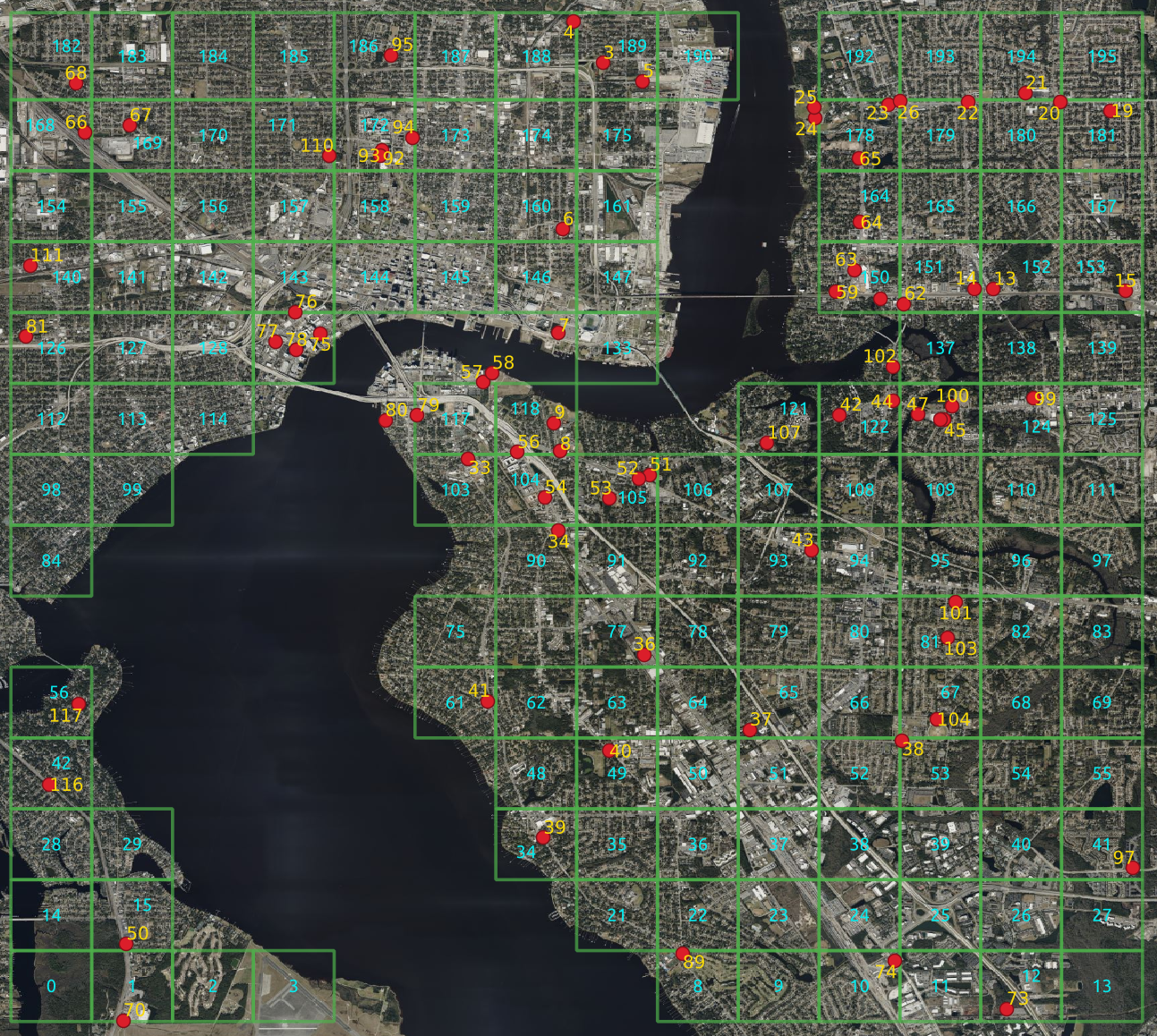}
  \caption{GCPs available for Jacksonville: Each GCP ({\color{red}red} dot) has a ID ({\color{yellow}yellow} number) associated with it. We also show the tile borders using \green{green} lines and tile numbers are displayed at the center of each tile in {\color{cyan} cyan}.}
  \label{fig:jax_all_gcps}
\end{figure}

We used Ground Control Points (GCPs) collected by \cite{accenture_gcps} for calculating SatDepth Maps accuracy. We used a total of 76 GCPs spanning 50 tiles of our Jacksonville AOI as shown in \cref{fig:jax_all_gcps}.
The most fundamental information regarding a GCP consists of its world coordinates ($\boldsymbol{X}^{GCP}$).  
However, the GCPs are accompanied by photographic images of the locations, aiding in their annotation within the satellite images. 
To annotate the GCPs in the satellite images, we use the RPC camera model ($\mathcal{P}$) to project the GCPs into the image ($\boldsymbol{x}_i = \mathcal{P}_i(\boldsymbol{X}^{GCP})$). We then present to a human (the annotator) the relevant patch of a satellite image 
centered at the projected pixel $\boldsymbol{x}_i$ and ask them to annotate the correct point in the image.

\begin{figure}[!h]  
  \centering
  \begin{subfigure}{0.25\linewidth}
    \includegraphics[width=\linewidth, height=\linewidth]{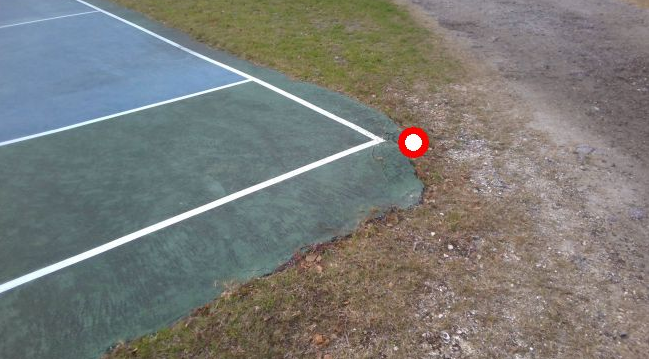}
    \caption{GCP image taken by surveyor}
    \label{fig:annotation_process_survey}
  \end{subfigure}
  \hspace*{2mm}
  \begin{subfigure}{0.25\linewidth}
    \includegraphics[width=\linewidth, height=\linewidth]{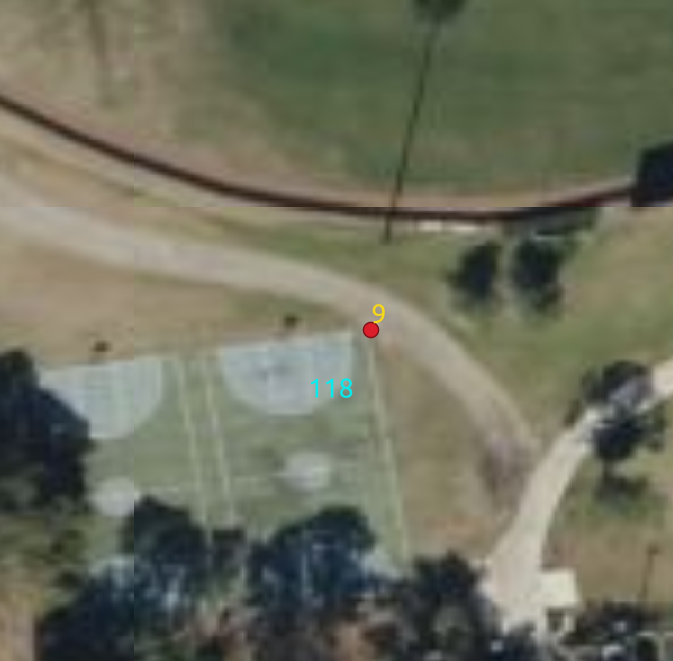}
    \caption{GCP overlayed on Bing Maps}
    \label{fig:annotation_process_qgis}
  \end{subfigure}
  \hspace*{2mm}
  \begin{subfigure}{0.25\linewidth}
    \includegraphics[width=\linewidth, height=\linewidth]{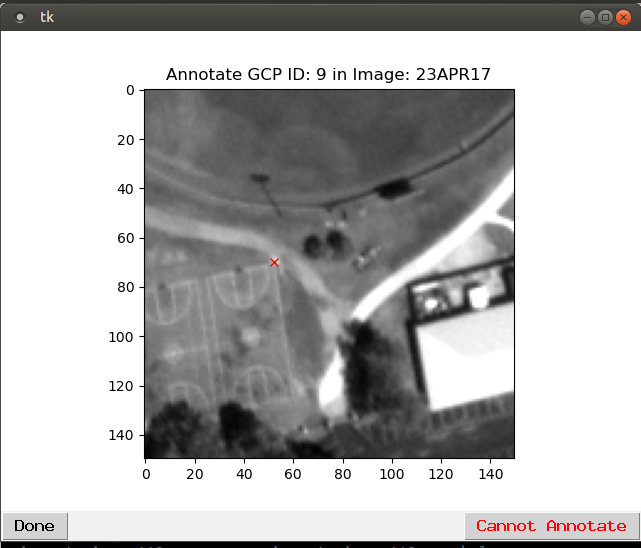}
    \caption{GCP annotation tool}
    \label{fig:annotation_process_tool}
  \end{subfigure}
\caption{GCP annotation process (for tile \# 118 of Jacksonville): A Human annotator is presented with (a) the image taken by the surveyor and (b) GCP location overlayed on Bing Maps. Then the human annotates the GCP in the satellite image using annotation tool shown in (c).}
\label{fig:annotation_process}
\end{figure}

\begin{figure}[!h]
  \centering
  \begin{subfigure}{0.3\linewidth}
    \includegraphics[width=\linewidth, trim=2cm 2cm 2cm 2cm, clip]{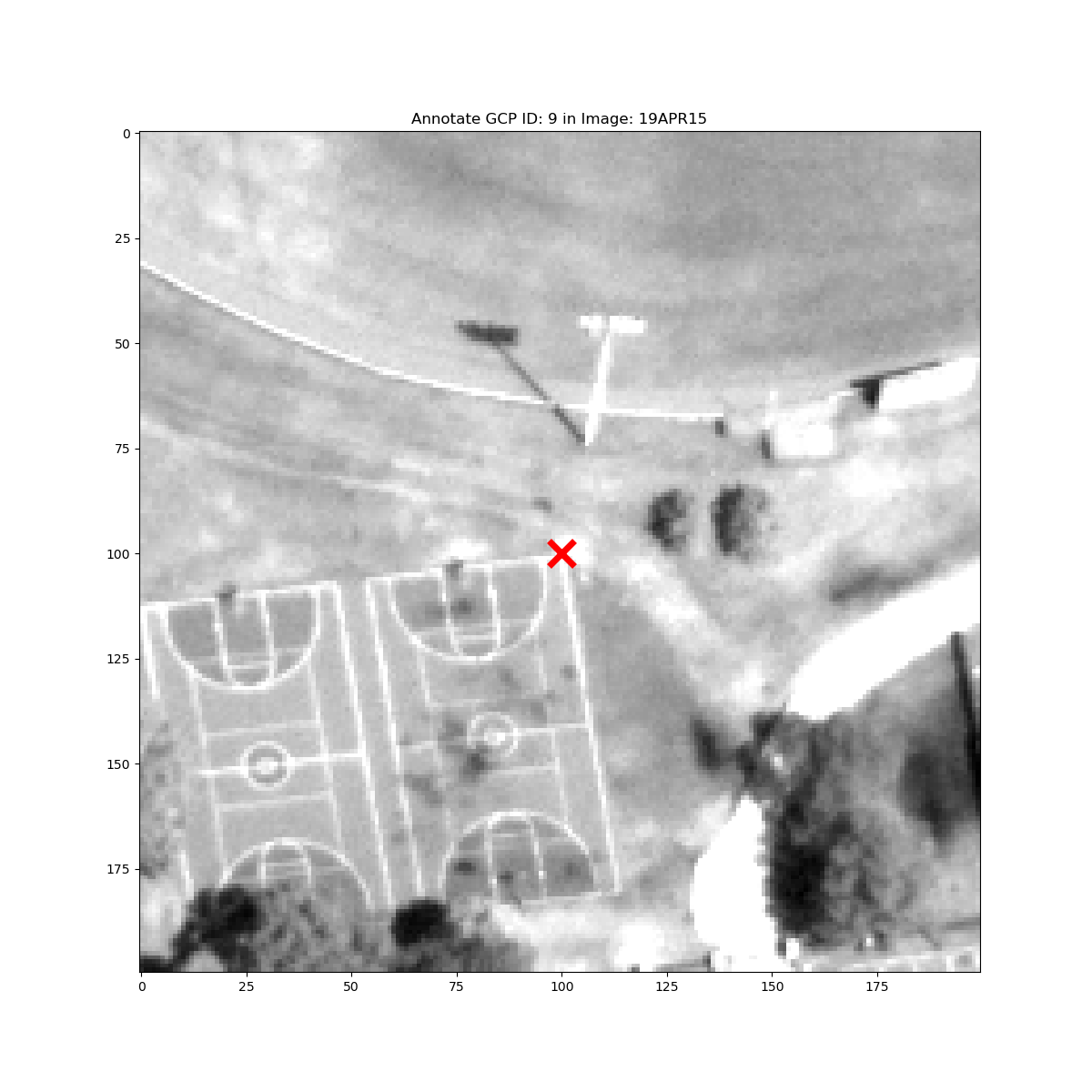}
    \caption{19APR15}
    \label{fig:jax_gcp_annotated_0}
  \end{subfigure}
  \begin{subfigure}{0.3\linewidth}
    \includegraphics[width=\linewidth,trim=2cm 2cm 2cm 2cm, clip]{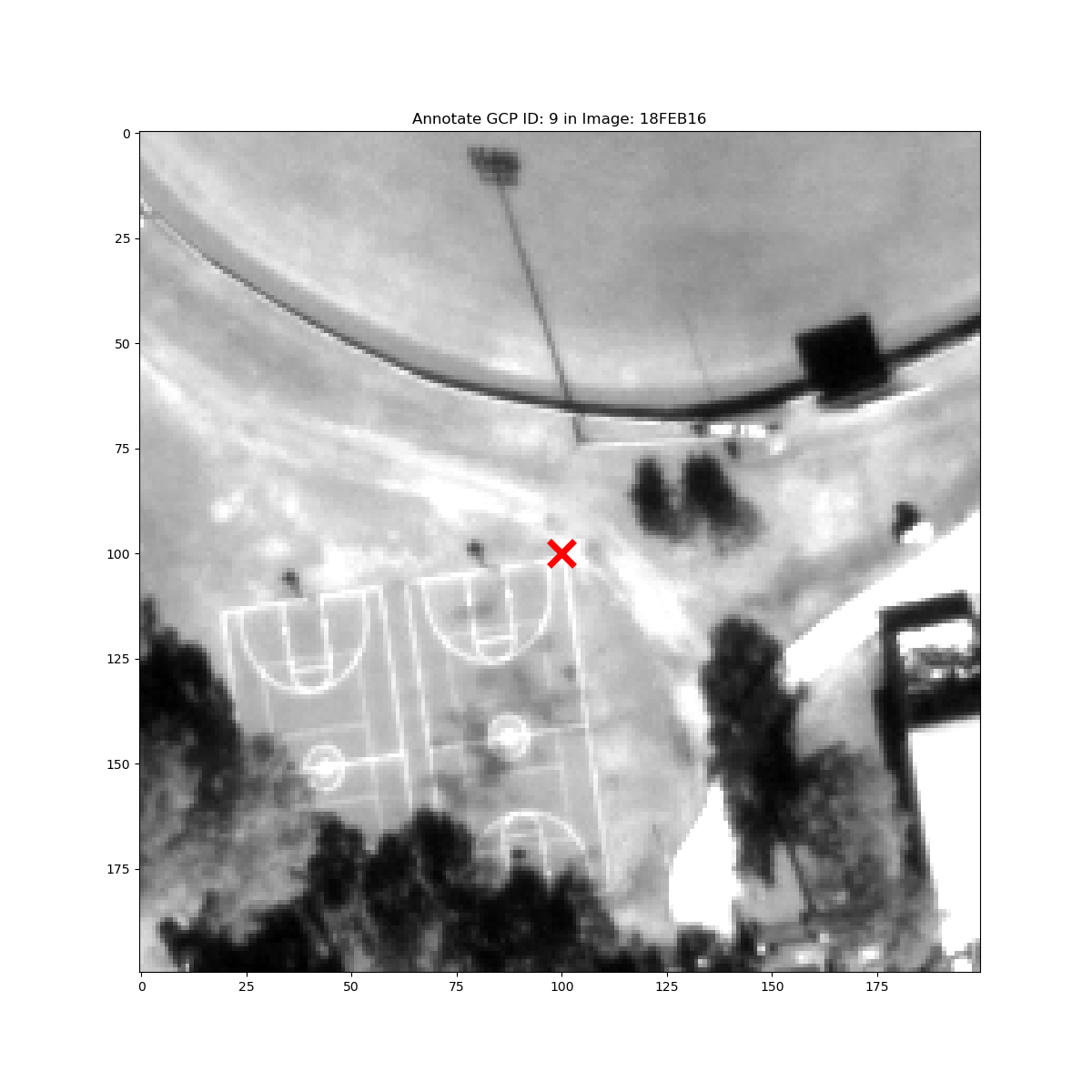}
    \caption{18FEB16}
    \label{fig:jax_gcp_annotated_1}
  \end{subfigure}
  \begin{subfigure}{0.3\linewidth}
    \includegraphics[width=\linewidth, trim=2cm 2cm 2cm 2cm, clip]{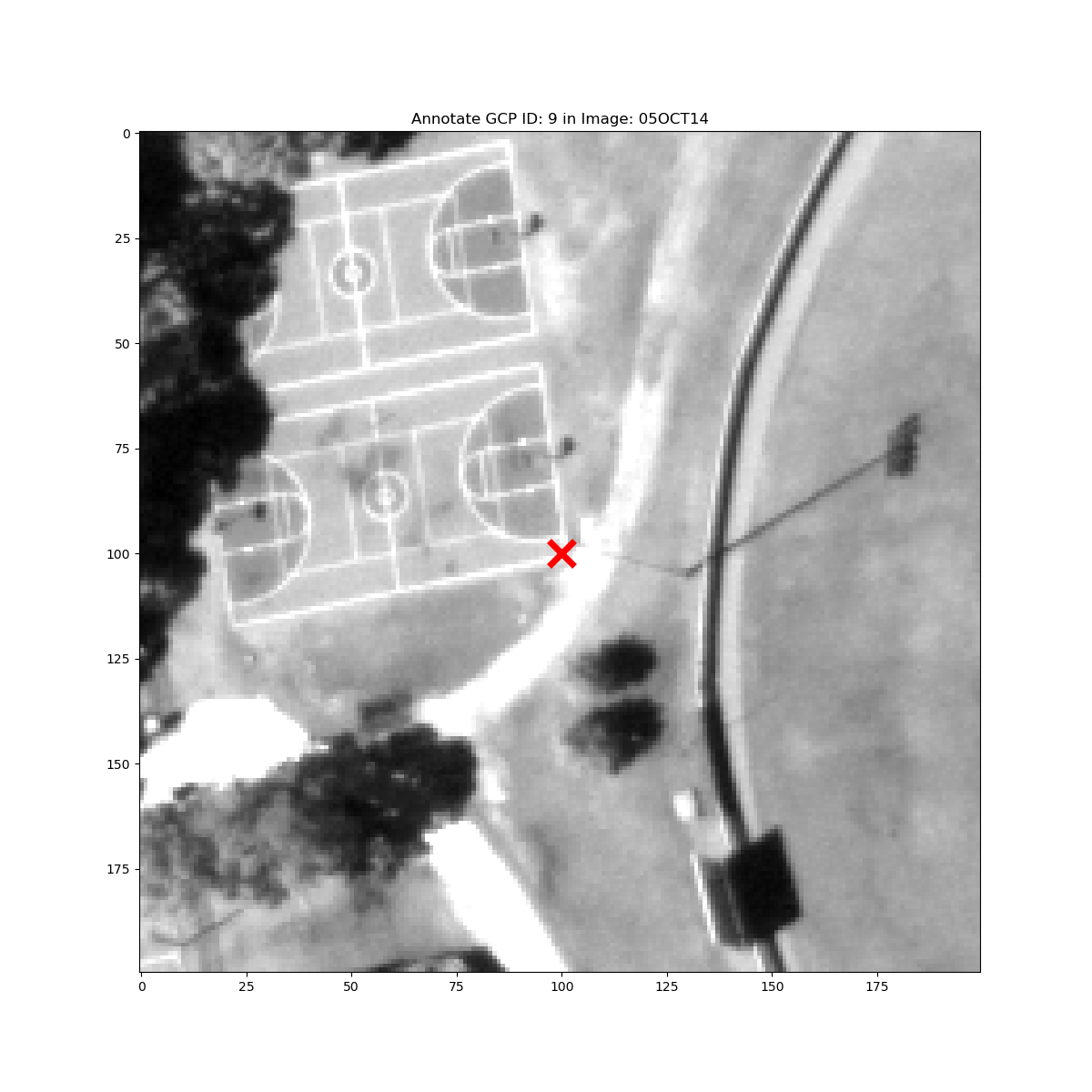}
    \caption{05OCT14}
    \label{fig:jax_gcp_annotated_2}
  \end{subfigure}
  \caption{Example GCP annotations (for tile \# 118 of Jacksonville) for three satellite images}
  \label{fig:example_annotations}
\end{figure}

To assist with the annotation, we show the annotator the image of the GCP collected by the surveyor and a rough satellite view of the same location by overlaying the GCP on Bing Maps, as shown in \cref{fig:annotation_process}. A few examples of the annotations carried out by our annotators are shown in \cref{fig:example_annotations}.

It should be noted that our GCPs were collected in January 2022, whereas the source imagery in SatDepth dates from 2014 to 2016. As a result, some GCPs could not be annotated as they were not present in the images. Additionally, in cases where clouds, haze, or occlusion made it difficult to clearly distinguish a GCP, our annotators declared that they could not annotate the GCP in the image.

\begin{figure}[!h]
  \centering
  \begin{subfigure}{0.4\linewidth}
  \includegraphics[width=\linewidth, height=\linewidth, trim=2cm 2cm 2cm 2cm, clip]{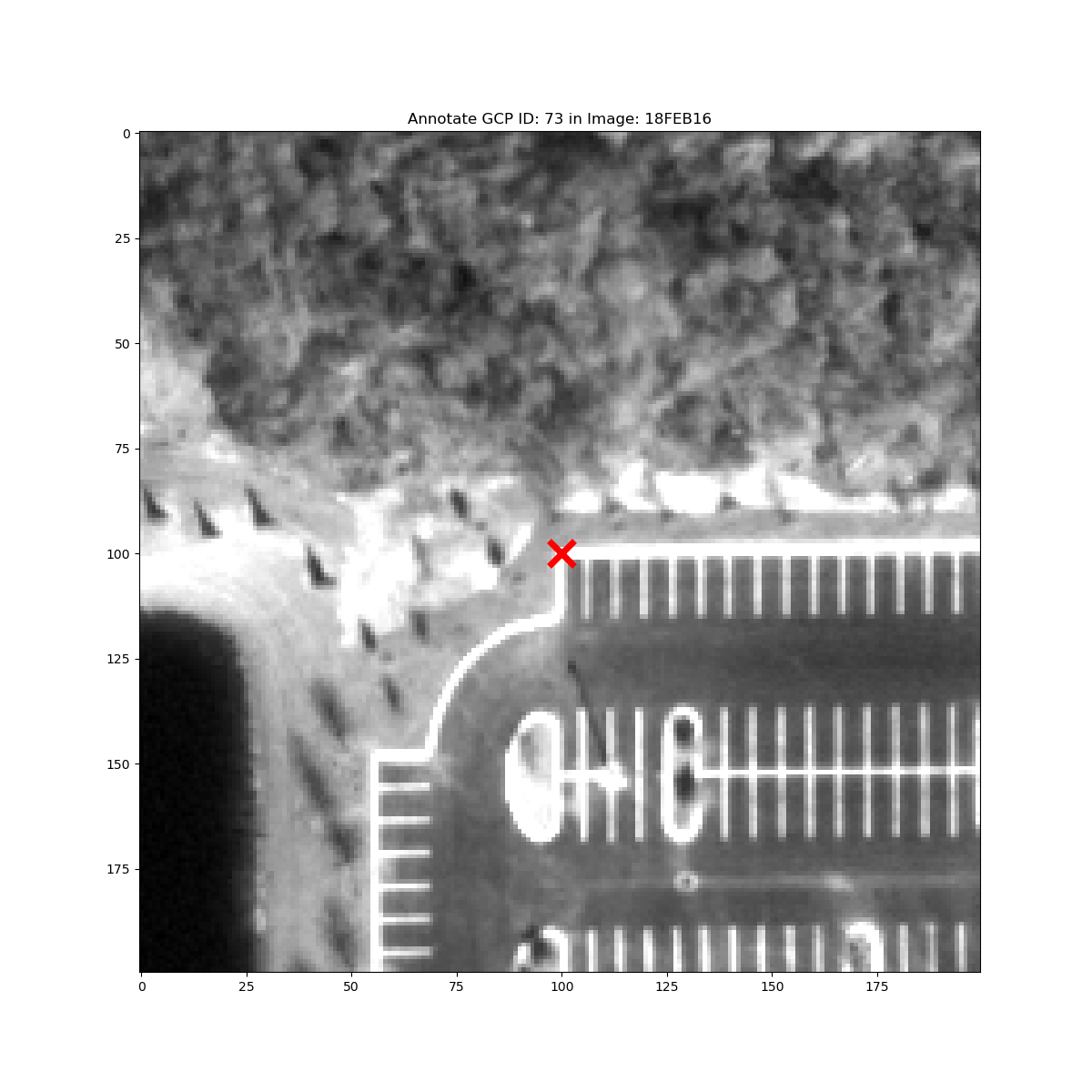}
  \caption{18FEB16}
  \end{subfigure}
  \hspace*{2mm}
  \begin{subfigure}{0.38\linewidth}
    \includegraphics[width=\linewidth, height=\linewidth]{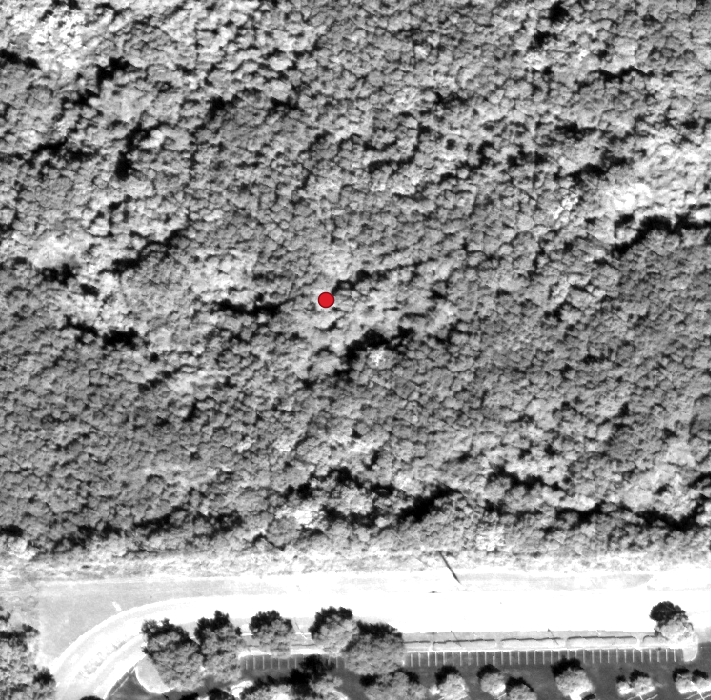}
    \caption{11OCT14}
  \end{subfigure}
  \caption{GCP \# 73 for tile \# 12 of Jacksonville: The GCP was annotated ({\color{red}red} x marker) as the parking lot corner as shown in (a), but it was not present in the 2014-2015 images as shown in (b). The expected location of the GCP is shown in (b) using a {\color{red}red} dot.}
  \label{fig:jax_gcp_73}
  \end{figure}

\clearpage
\subsubsection{SatDepth Accuracy}
To compute the accuracy of SatDepth Maps, we calculate the three error measures ($\epsilon^{a}_{3D}, \epsilon^{r}_{3D}, \epsilon^{r}_{2D} $) as mentioned in the main manuscript. Additionally, to compute the Absolute 3D error $\epsilon^{a}_{3D}\defeq \norm{\boldsymbol{X}_i - \boldsymbol{X}^{GCP}}_2$, we convert the latitude, longitude, height coordinates of both the world points ($\boldsymbol{X}_i,\,\boldsymbol{X}^{GCP}$) to the XYZ coordinates using the ECEF (Earth-Centered, Earth-Fixed) coordinate frame. We then compute the Euclidean distance between the GCP ($\boldsymbol{X^{GCP}}$) and the corresponding point in the SatDepth Map ($\boldsymbol{X}_i$).

\begin{figure}[!h]
  \centering
  \begin{subfigure}{0.25\linewidth}
      \includegraphics[width=\linewidth]{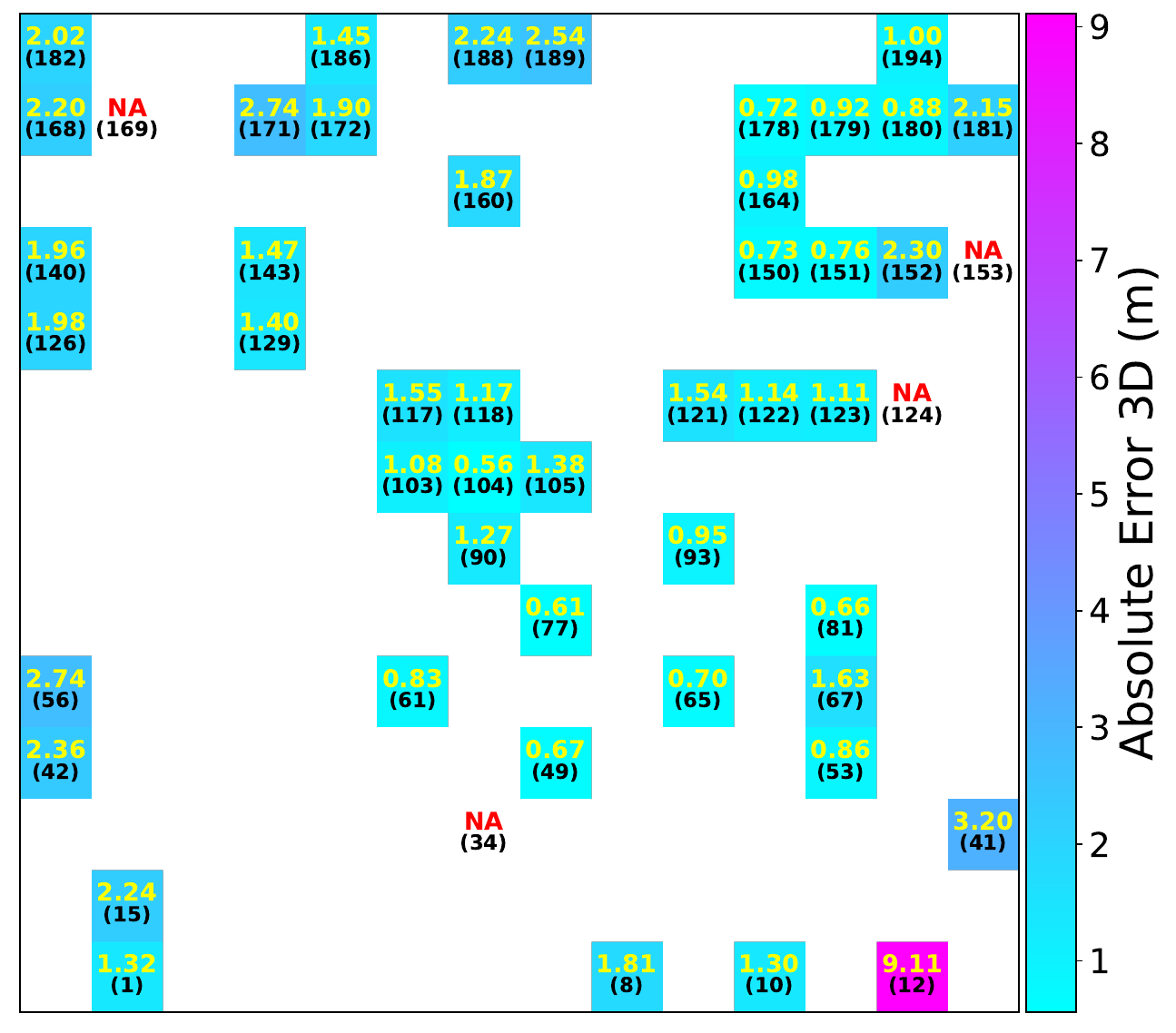}
      \caption{Absolute 3D Error}
      \label{fig:jax_abs_error_3d}
  \end{subfigure}
  \hspace*{2mm}
  \begin{subfigure}{0.25\linewidth}
    \includegraphics[width=\linewidth]{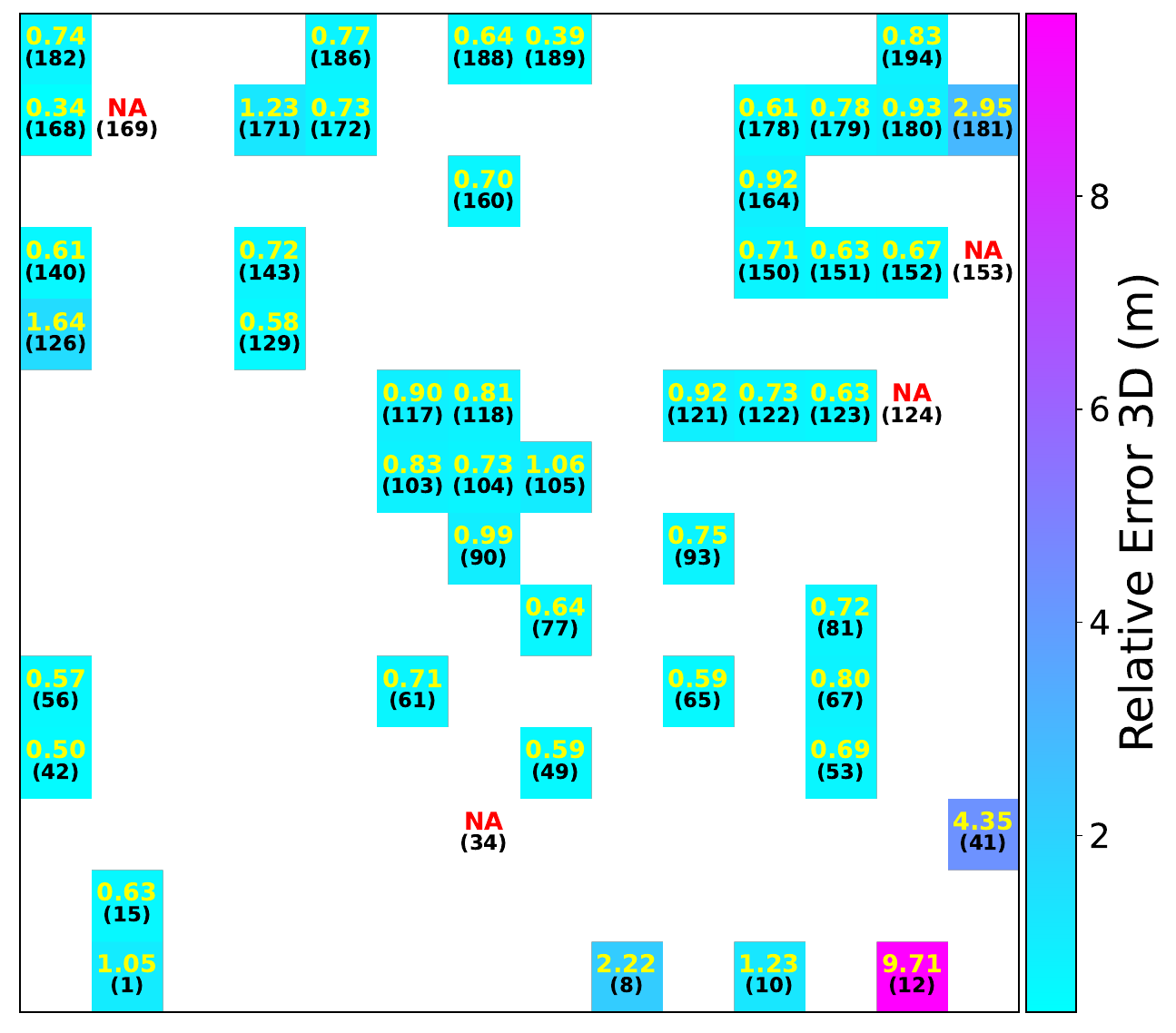}
    \caption{Relative 3D Error}
    \label{fig:jax_rel_error_3d}
  \end{subfigure}
  \hspace*{2mm}
  \begin{subfigure}{0.25\linewidth}
    \includegraphics[width=\linewidth]{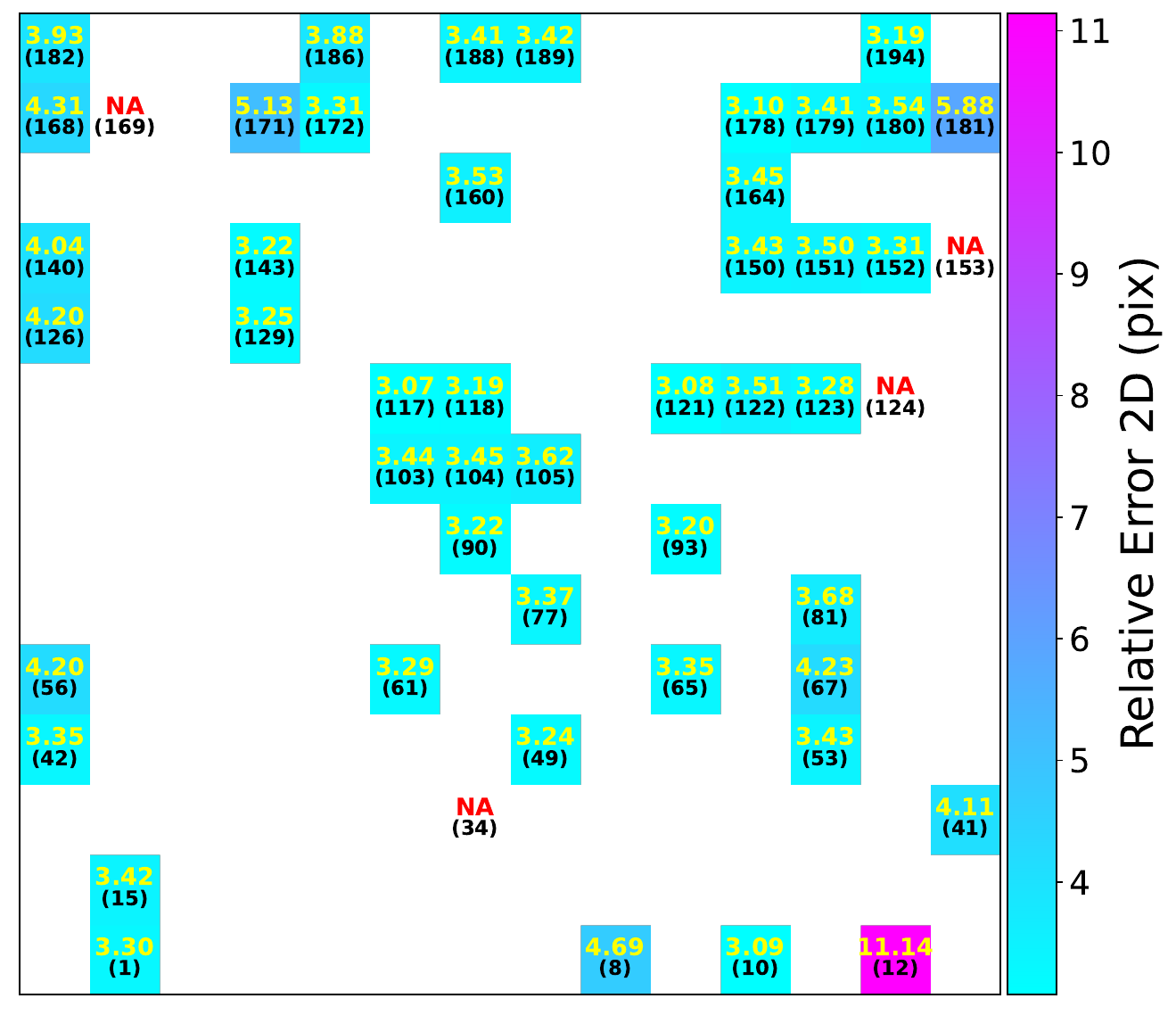}
    \caption{Relative 2D Error}
    \label{fig:jax_rel_error_2d}
  \end{subfigure}
  \hspace*{2mm}
    \caption{Averaged error plots for Jacksonville. We indicate tile numbers in brackets and use ``{\color{red}NA}'' for tiles where GCP could not be annotated.}
    \label{fig:jax_gcp_errors}
\end{figure}

\begin{figure}[!h]
  \centering
  \begin{adjustbox}{valign=t}
  \begin{subfigure}{0.3\linewidth}
    \includegraphics[width=\linewidth, trim=0 -1.5cm 0 0, clip]{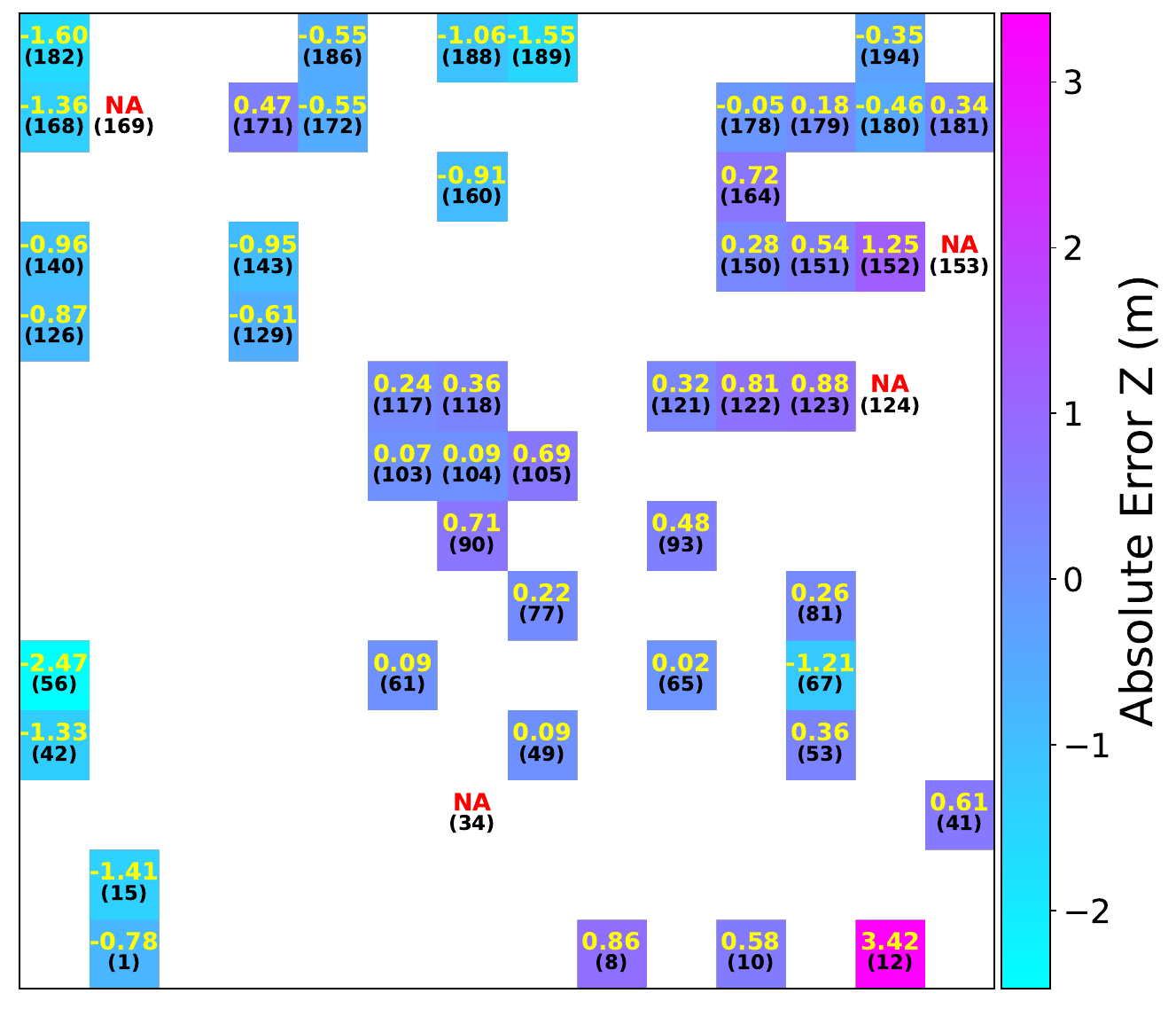}
    \caption{Z component}
    \label{fig:jax_abs_error_3d_z}
  \end{subfigure}
\end{adjustbox}
\begin{adjustbox}{valign=t}
  \begin{subfigure}{0.3\linewidth}
    \includegraphics[width=\linewidth]{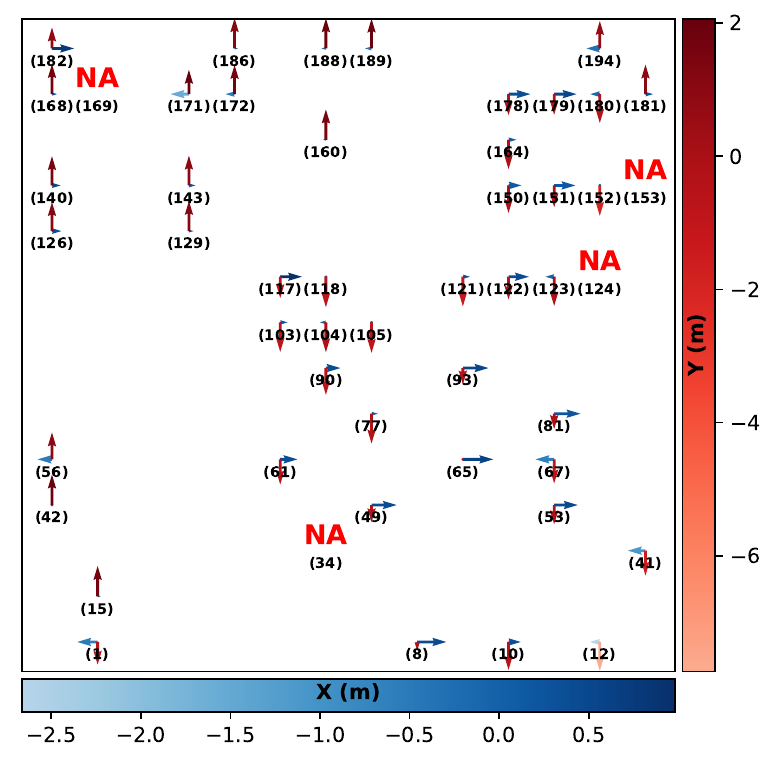}
    \caption{X \& Y components}
    \label{fig:jax_abs_error_3d_xy}
  \end{subfigure}
  \end{adjustbox}
  \caption{Individual components of Absolute 3D Error for Jacksonville: (a) Z component as a color plot and (b) X \& Y components as a vector field plot where the arrows are colored and scaled relative to their magnitudes. We indicate tile numbers in brackets and use ``{\color{red}NA}'' for tiles where GCP could not be annotated.}
  \label{fig:jax_abs_error_components}
\end{figure}

A closer examination of the individual components of Absolute 3D error, revealed a consistent error pattern, which was uniform across all tiles. This indicated that there is a constant shift error in our DSMs. We used this information to further reduce our Absolute 3D error by aligning the DSMs to the GCPs using a Monte-Carlo simulation as discussed in \cref{sec:monte_carlo_appendix}. Finally, we show the summary of errors for Jacksonville in \cref{fig:jax_gcp_errors,fig:jax_abs_error_components}. It should be noted that for extracting ground-truth matches, only the relative accuracy influences the match quality. However, achieving greater absolute accuracy remains beneficial for integrating with other datasets or potential unforeseen applications.

Additionally, we notice that tile \#12 of Jacksonville has a higher error compared to other tiles. This was visually verified to be caused by a change in the scene. Specifically, for tile \#12, the recorded GCP (GCP \#73) was a parking lot corner. We were able to annotate this GCP in the satellite images for the year 2016. However, the parking lot was not present in the 2014-2015 images. Upon visual inspection, we observed a wooded area in the place of the parking lot in the 2014-2015 images. As our DSM records the top-N median height, the DSM had recorded the higher elevation corresponding to the wooded area in the 2014-2015 date range. This caused a higher error in the absolute 3D errors for the GCP. We show the annotated GCP in the satellite images for the two time periods in \cref{fig:jax_gcp_73}.

\vspace*{-2em}

\subsubsection*{Monte-Carlo Experiment for Estimating Shift Corrections}\label{sec:monte_carlo_appendix}
\begin{figure}[!h]
  \centering
  \includegraphics[width=0.3\linewidth]{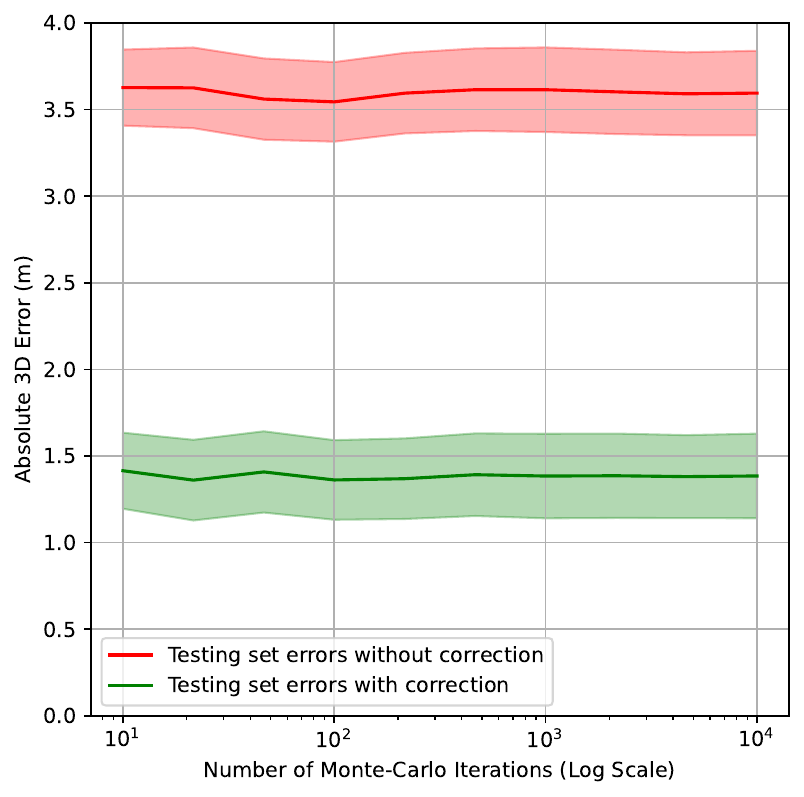}
  \caption{Absolute 3D Error of the Testing Set in the Monte-Carlo Experiment. The plot shows the average error (colored lines) along with shaded envelopes representing the single standard deviation from the mean.}   
  \label{fig:monte_carlo_gcp_shift_correction}
\end{figure}

\begin{figure}[!h]
  \begin{subfigure}{0.5\linewidth}
    \begin{adjustbox}{valign=t}
      \includegraphics[width=0.48\linewidth]{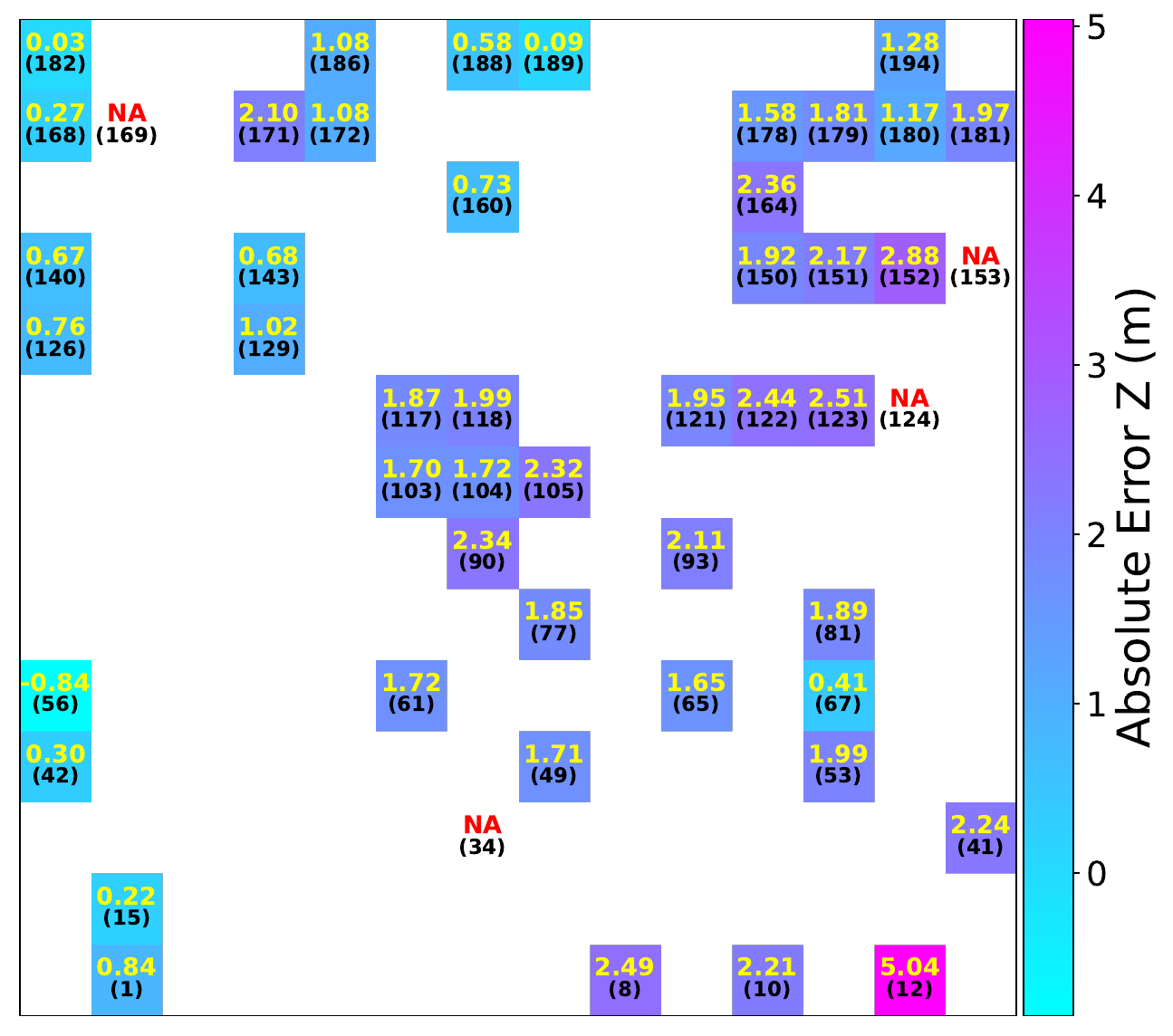}
    \end{adjustbox}
    \begin{adjustbox}{valign=t}
      \includegraphics[width=0.45\linewidth]{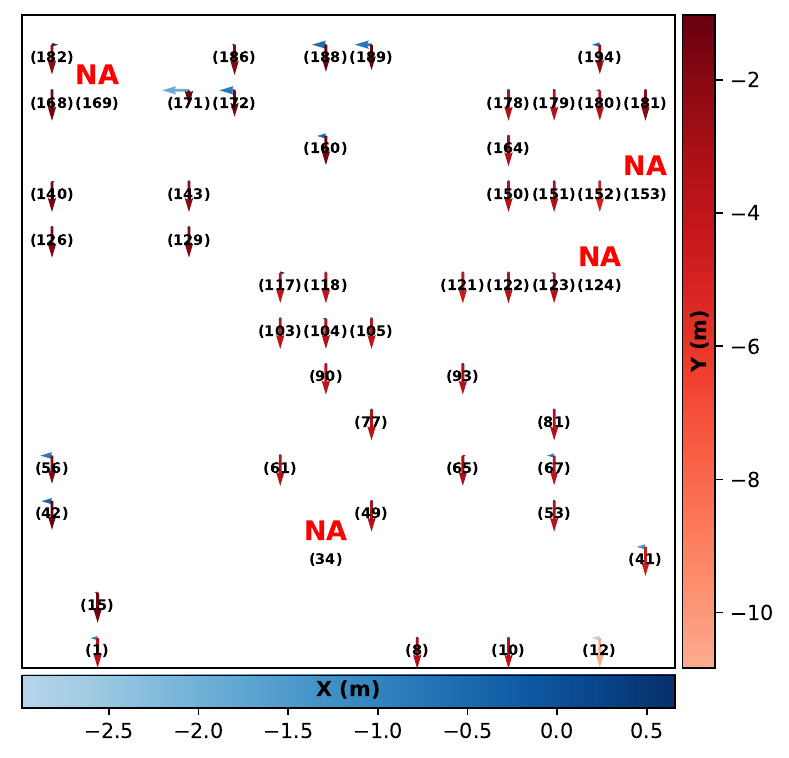}
    \end{adjustbox}
    \caption{Before Monte-Carlo Shift Correction}
  \end{subfigure}
  \begin{subfigure}{0.5\linewidth}
    \begin{adjustbox}{valign=t}
      \includegraphics[width=0.48\linewidth]{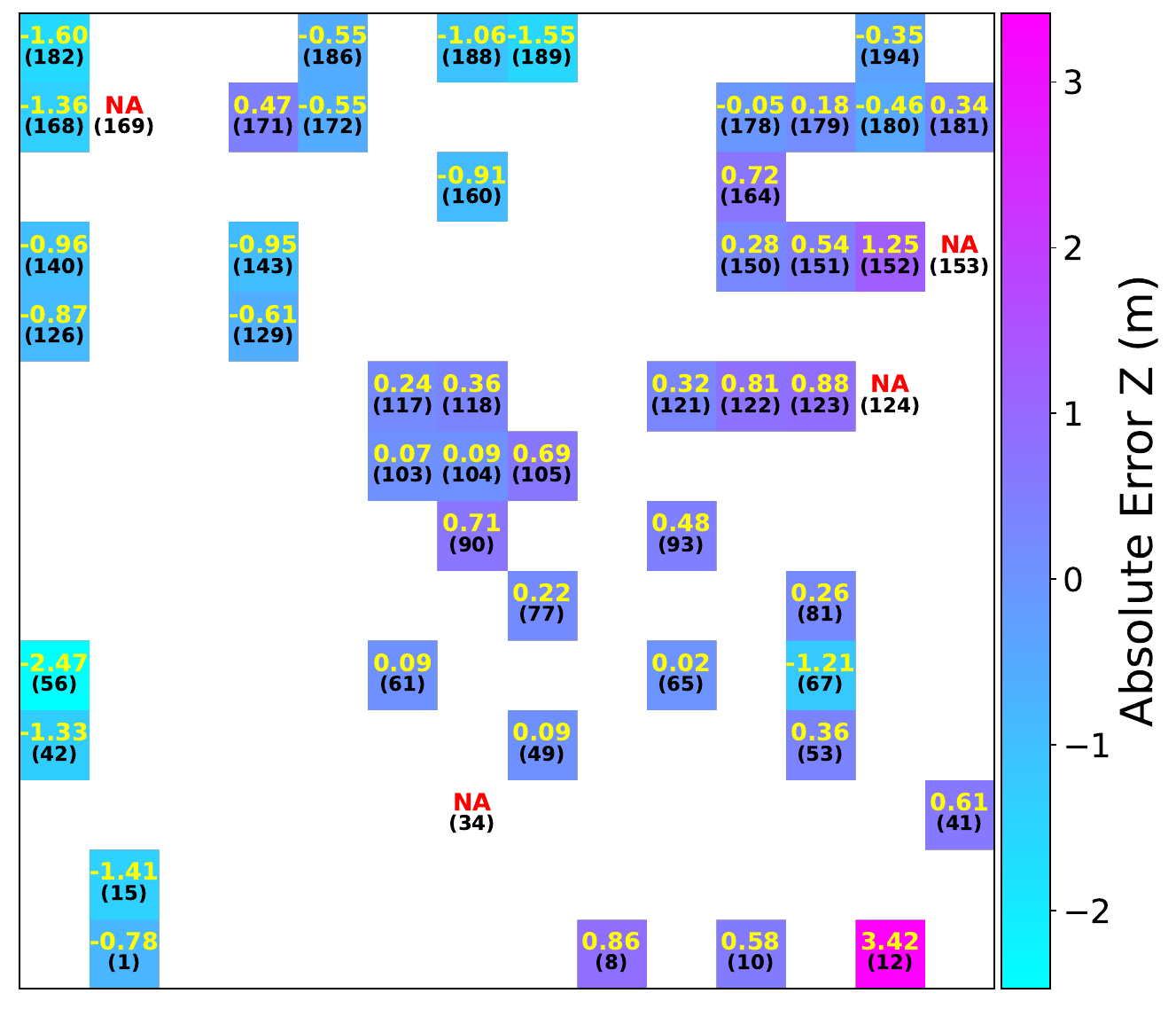}
    \end{adjustbox}
    \begin{adjustbox}{valign=t}
      \includegraphics[width=0.45\linewidth]{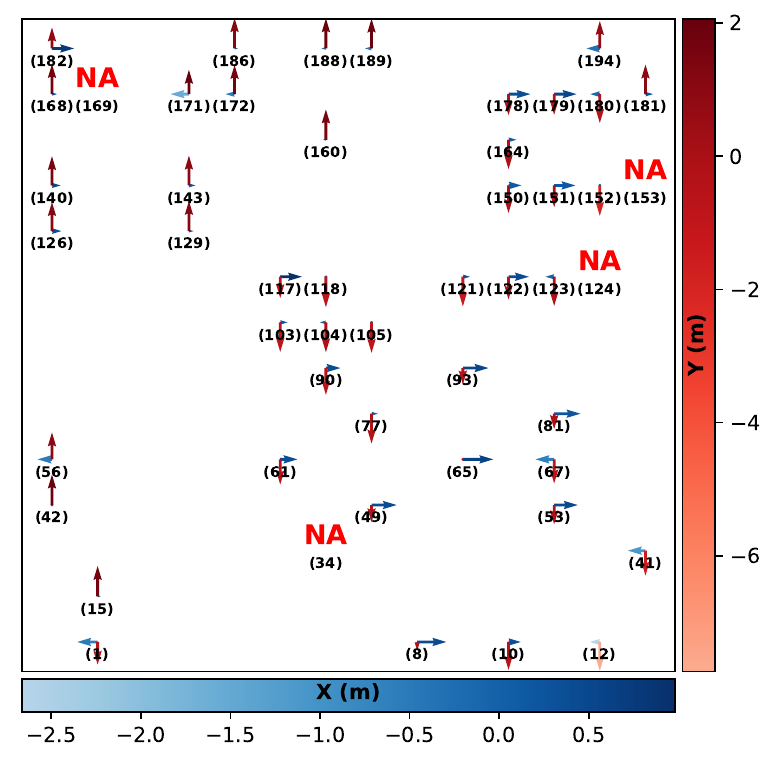}
    \end{adjustbox}
    \caption{After Monte-Carlo Shift Correction}
  \end{subfigure}
  \caption{Individual components of absolute 3D Error for Jacksonville before and after Monte-Carlo shift correction. We indicate tile numbers in brackets and use ``{\color{red}NA}'' for tiles where GCP could not be annotated.}
  \label{fig:before_after_mc_shifts}
\end{figure}

As discussed earlier, we observed a consistent error pattern in the absolute 3D errors across all tiles. This indicated that there is a constant shift error in our DSMs. To estimate these shift errors, we carried out a Monte-Carlo estimation of the absolute 3D errors vis-a-vis the GCP. In the Monte-Carlo experiments, we randomly split the GCPs into two sets (70\% train and 30\% test). We then calculate the average 3D error using the training set and applied the estimated correction to the testing set. We carried out a series of Monte-Carlo experiments by varying the number of random simulations from 10 to 10000. \Cref{fig:monte_carlo_gcp_shift_correction} shows the results for the absolute 3D error for the testing set with and without the correction estimated from the training set. We observe that by applying a correction of (X: -0.294310 m, Y: -3.096609 m, Z: 1.6104 m where XYZ are the ECEF coordinates), our absolute 3D error drops from 3.26 m (std-dev 0.32) to 1.44 m (std-dev 0.24) for the testing set. We also show the individual components of the absolute 3D error for Jacksonville before and after the Monte-Carlo shift correction in \cref{fig:before_after_mc_shifts}.

Having shown how we reduce the absolute 3D error for the entire dataset, we wish to point out that the main application we envision for our dataset only required the relative errors to be at subpixel level. The dataset is meant to help researchers experiment with new neural architectures for image matching. That requires matching precision on a pairwise basis in a relative sense, and not with respect to an absolute coordinate frame defined by the GCPs. However, using the above shifts, one can easily align the SatDepth dataset to the GCPs for applications that require higher absolute accuracy. The shifts need to be applied to the DSMs, SatDepth Maps, and the corresponding RPCs. In the case of the DSMs, the geotransforms of DSMs can be updated by subtracting the XY shifts and the DSM raster needs to be updated by subtracting the Z shift. For the SatDepth Maps, the shifts need to be converted to latitude, longitude, height and then subtracted from the SatDepth Maps. Finally, for the RPCs, the latitude, longitude, and height offsets of RPCs can be updated by subtracting the shifts.

\clearpage
\subsection{Comparison against other Stereo Pipelines}
We follow the approach of \cite{bosch2017metric} for comparing the quality of the DSM generated from our stereo pipeline against other pipelines. This approach involves comparison of the photogrammetric DSM with respect to the ground truth Lidar DSM. The IARPA MVS3D Challenge \cite{bosch2016multiple_mvs3d} provides ground truth Lidar point cloud for San Fernando which is used to calculate the evaluation metrics as reported in \cite{bosch2016multiple_mvs3d}. The evaluation metrics are defined as follows:
\begin{itemize}
  \item \textbf{Completeness}: Percentage of points in the photogrammetric DSM where the absolute difference of heights is less than 1 meter with respect to the ground truth Lidar DSM. The higher Completeness number indicates denser photogrammetric DSM.
  \item \textbf{RMSE}: Root Mean Squared Error (RMSE) of heights over valid pixels in both photogrammetric and ground truth Lidar DSM. Lower RMSE indicates higher absolute accuracy of the photogrammetric DSM.
  \item \textbf{MAE}: Median Absolute Z-Error (MAE) over valid pixels in both photogrammetric and ground truth Lidar DSM. Lower MAE indicates higher absolute accuracy of the photogrammetric DSM.
\end{itemize}

\begin{figure}[!h]
  \centering
  \begin{adjustbox}{valign=t}
  \begin{subfigure}{0.45\linewidth}
    \includegraphics[width=\linewidth]{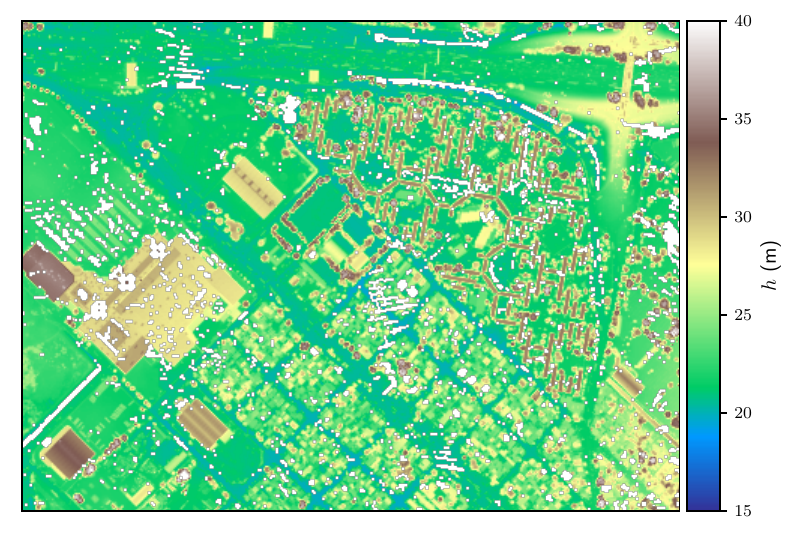}
    \caption{Ground Truth Lidar DSM over San Fernando. The colorbar indicates the heights (in meters) associated with each color. The white regions indicate non-valid (empty) pixels.}
    \label{fig:lidar_dsm}
  \end{subfigure}
  \end{adjustbox}
  \hspace{2mm}
  \begin{adjustbox}{valign=t}
  \begin{subfigure}{0.47\linewidth}
    \includegraphics[width=\linewidth]{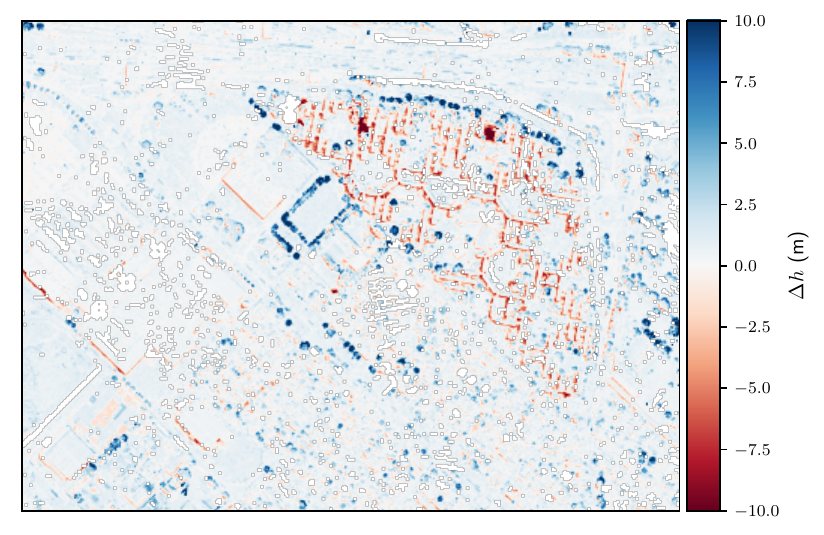}
    \caption{Difference between Lidar DSM and our DSM}
    \label{fig:diff_dsms}
  \end{subfigure}
  \end{adjustbox} 
  \caption{Qualitative comparison of our DSM with respect to ground truth Lidar}
  \label{fig:pipeline_comparison}
\end{figure}

To carry out the comparison, we crop the ground truth Lidar point clouds to our San Fernando AOI. We then convert the Lidar point cloud to DSM by retaining the maximum height at each latitude and longitude location. To enable comparison, the resolution of the Lidar DSM is kept the same as the photogrammetric DSM. The Lidar DSM is shown in \refFig{fig:lidar_dsm}. We then compute the above metrics and report them in \refTab{table:pipeline_metrics_comparison}. The metrics for other pipelines were obtained from \cite{bosch2017metric}. The metrics indicate that our pipeline's performance is similar to other stereo pipelines.
This means that other pipelines can be used to generate aligned cameras and DSMs, after which \textbf{\textit{depthifypp}} can create SatDepth Maps, allowing researchers to extend the dataset as more satellite images become publicly available.

Furthermore, the metrics used for comparison only report the averaged error for the DSMs. For more detailed analysis we show the difference of the heights between the Lidar DSM and our DSM in \refFig{fig:pipeline_comparison}. We observe that the differences are smaller for man-made structures and are larger only for areas covered by trees.

\begin{table}[h!]
  \centering
  \begin{tabular}{ccccc}
    \toprule
    Metric & 
    MAE (m)$\downarrow$ &
    RMSE (m)$\downarrow$ &
    Completeness (\%) $\uparrow$
    \\ 
    \midrule 
    JHU/APL\cite{bosch2017metric} & 0.47 & 2.20 & 64.1 \\ 
    RSP\cite{qin2016rpc} & 0.39 & 2.31 & 68.7 \\ 
    ASP\cite{nasa_ames} & 0.35 & 2.27 & 69.4 \\
    S2P\cite{Franchis_ISPRS14_S2P} & 0.37 & 2.59 & 73.2 \\
    \textbf{Our} & 0.59 & 2.38 & 70.53 \\
    \bottomrule
\end{tabular}
  \caption{Comparison of the quality of the DSM generated from different stereo pipelines over San Fernando.}
  \label{table:pipeline_metrics_comparison}
\end{table}

\vspace*{-1em}

This concludes our discussion on the Dataset Generation part of SatDepth. We will now proceed with our discussion on how to use the SatDepth dataset for training and benchmarking in the next section.

\clearpage
\section{Dataset Benchmarking}\label{sec:benchmarking_appendix}

\subsection{Training and Testing Details}

\subsubsection*{Patch Extraction}
\begin{figure}[!h]
  \centering
  \begin{subfigure}{0.4\linewidth}
    \includegraphics[width=\linewidth]{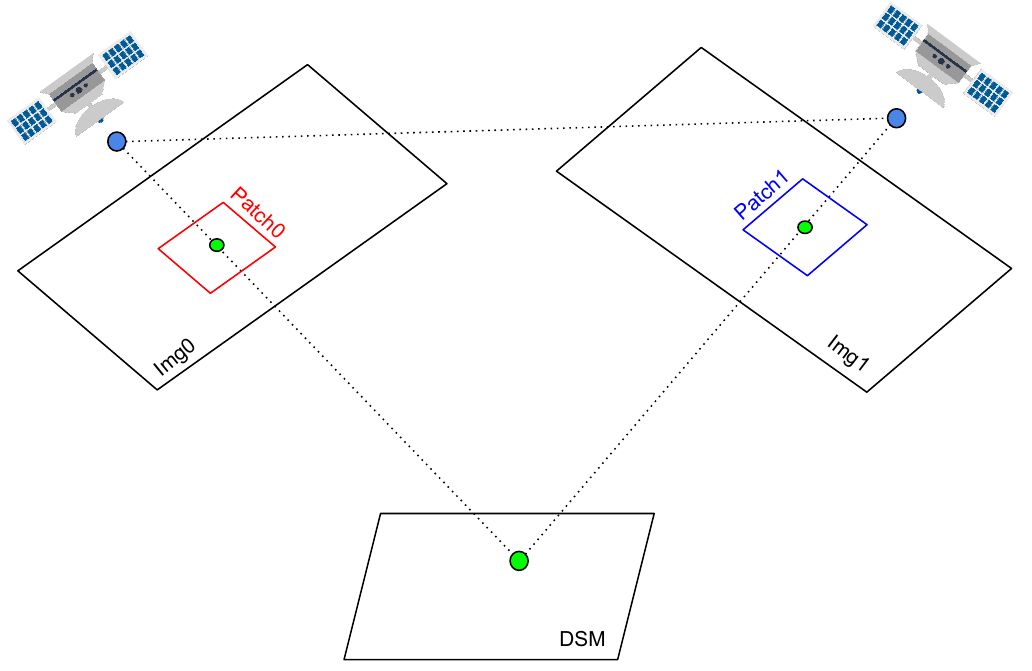}
    \caption{Training}
    \label{fig:patch_extract_train}
  \end{subfigure}
  \begin{subfigure}{0.4\linewidth}
  \includegraphics[width=\linewidth]{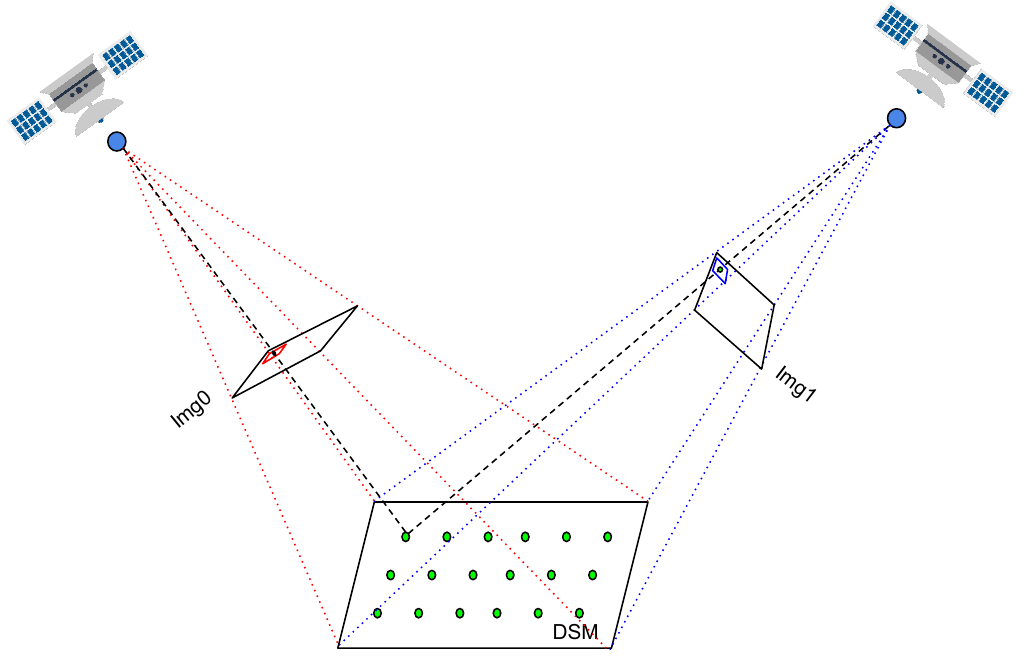}
  \caption{Testing}
  \label{fig:patch_extract_test}
  \end{subfigure}
  \caption{Patch extraction procedure for training and testing. We display the extracted pair of patches using {\color{red} red} and {\color{blue} blue} boxes in the two satellite images.}
  \label{fig:patch_extraction}
\end{figure}

In our experiments, during training and validation, we extract pairs of image patches by projecting randomly chosen 3D points (using the DSM) into a pair of satellite images. Each projected point thus obtained serves as the center of a $p \times p$ image patch as shown in \cref{fig:patch_extract_train}.  In this manner, we obtain what we may loosely refer to as ``matched pairs of corresponding patches'' in the two images. Subsequently, it would be up to the neural network to discover the pixel-based correspondences in such pairs of corresponding patches.

Conversely, during testing, we generate a uniform grid of 3D points (using the DSM) and project them into a pair of images to obtain the $p \times p$ image patches centered at the projected points. This process is illustrated in \cref{fig:patch_extract_test}. Subsequently, the neural network extracts the pixel-based correspondences in these pairs of patches and then concatenate the matches to form matches for the entire image.

When working with the image patches and their corresponding affine cameras $\widehat{\mathcal{P}}$, extracting ground-truth correspondences becomes efficient due to the use of affine approximations. This efficiency arises because $\widehat{\mathcal{P}}$ is a linear operator (a $3 \times 4$ matrix), unlike the non-linear function $\mathcal{P}$. To extract the ground-truth correspondences, we transfer a grid of world points ($\boldsymbol{X}^k_i$) from $I_i$ to $I_j$ using $\widehat{\mathcal{P}}_j$ for the forward projection. Since this forward projection involves a simple matrix multiplication operation, it can be efficiently performed directly on the GPU.

\subsubsection*{Workflow}
\begin{figure}[!h]
  \centering
  \includegraphics[width=0.7\linewidth]{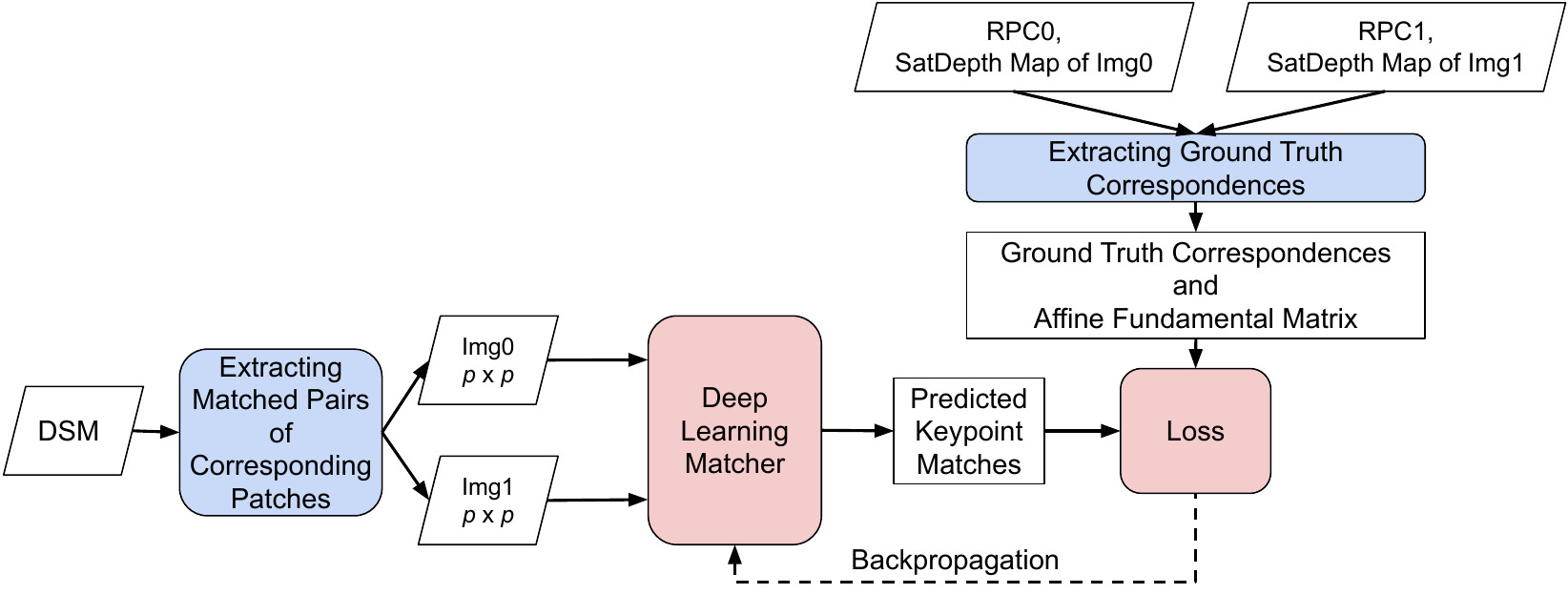}
  \caption{Model workflow for training of benchmarked models} 
  \label{fig:model_traning_workflow}
\end{figure}

The training workflow for the benchmarked models can be summarized using the workflow shown in \cref{fig:model_traning_workflow}. In this workflow, we first extract the image patches ($I_i, I_j$) from the satellite images using the DSM. We also compute the affine cameras for the image patches and compute a set of ground-truth correspondences ($\{\boldsymbol{x}^{k}_{i},\leftrightarrow \boldsymbol{x}^{k}_{j}\}$) using the SatDepth Maps as explained in the ``Depth Map Generation'' section of our paper \cite{satdepth_preprint}. 
The image patches are fed to the network and the ground-truth correspondences are used for supervision. 
The deep-learning based matcher predicts a set of correspondences between the patch pairs, and a loss is computed by comparing the set of predicted correspondences with the set of ground-truth correspondences. This loss is backpropagated through the network to update the weights, and the process is repeated over a fixed number of epochs until the model is trained. 
Subsequently, during testing, we generate a uniform grid of 3D points, project them onto an image pair to obtain the image patches, extract matches using the trained network, and then join the sets of predicted matches to form the final set of matches for the image pair. This workflow is used for all four benchmarked models, with minor deviations due to the different forms of supervision employed by each network. 
We provide model specific details in \cref{sec:implementation_details}.

Finally, we evaluate the performance of the trained model on the test set using two error measures -- (1) \textbf{Precision of Matches} using Symmetric epipolar distance, and (2) \textbf{Pose Estimation Errors}. 

To calculate the symmetric epipolar distance, we first calculate the ground-truth Affine Fundamental Matrix $\widehat{F}_{GT}$ using the ground-truth Affine Cameras for the image patches. In other words, the Affine Fundamental Matrix for satellite images is a patch based concept and not an image based concept. For each patch in an image, its camera model depends on the location of the patch as shown in \cref{eq:rpc_affine}, therefore, the Affine Fundamental Matrix is a function of the location of the two image patches. It should be noted that the calculation of $\widehat{F}_{GT}$  does not involve any BA or RANSAC based logic involving the predicted matches. We calculate the symmetric epipolar distance using the predicted matches and the ground-truth fundamental matrix using \cref{eq:symmetric_epipolar_distance}. The precision of matches is then calculated as the percentage of predicted matches with symmetric epipolar distance less than $\delta_{epi}$.

To calculate the Pose Estimation Errors, we first estimate the Affine Fundamental Matrix using predicted matches with the help of RANSAC based logic. We then compare the estimated Affine Fundamental Matrix with the ground-truth Affine Fundamental Matrix. The comparison between the two matrices is done using affine camera motion parameters \cite{hartley2003multiple} : cyclotorsion ($\widehat{\theta}$), out-of-plane rotation ($\widehat{\phi}$), and scaling ($\widehat{s}$). We report the area under the cumulative curve (AUC) of the affine pose error as the maximum of angular error in ($\widehat{\theta},\,\widehat{\phi}$) for multiple thresholds.

\subsection{View-Angle and Track-Angle Differences}
In this section, we will explain what we mean by view-angle and track-angle differences between a pair of images ($I_i,\,I_j$).

\textbf{View-Angle Difference} ($\alpha^v$): This is the angle between the viewing vectors of the two satellite images. To compute this angle, we first create a normalized viewing vector $\boldsymbol{v}_i \in \mathbb{R}^3$ using the image metadata (IMD files). The image metadata stores the satellite azimuth and elevation angles. We create the viewing vector ($\boldsymbol{v}_i$) using \cref{eq:sat_view_vector}. We then compute $\alpha^v_{ij}$ using \cref{eq:view_angle}. The resulting angle is in the range $\alpha^v \in [0,\,\pi]$.

\begin{align}
\boldsymbol{v}_i &\defeq [cos(El_i)cos(Az_i),\,cos(El_i)sin(Az_i),\, sin(El_i)]^T \label{eq:sat_view_vector}  \\
\alpha^v_{ij} &= cos^{-1}(\boldsymbol{v}_i \cdot \boldsymbol{v}_j)  \label{eq:view_angle}  
\end{align}

\begin{figure}[!h]
  \centering
  \includegraphics[width=0.3\linewidth]{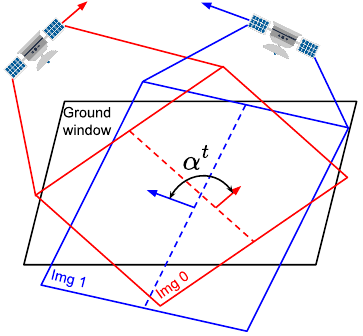}
  \caption{Track-Angle Difference ($\alpha^{t}$) for two satellite images is depicted by the angle made by the two {\color{red}red} and {\color{blue}blue} arrows. The arrows indicate the track direction of the two satellites and the dotted lines indicate the middle row of the images.}
  \label{fig:relative_track_angle}
\end{figure}

\textbf{Track-Angle Difference} ($\alpha^{t}$): Satellites capture images from various tracks through multiple revisits over a region. For a pair of images, the difference in track-angles introduces a ``relative rotation'' between the two images, as illustrated in \cref{fig:relative_track_angle}. We define the track-angle of the satellite as the angle formed by its track with respect to the True North direction. For a pair of images, we define the track-angle difference ($\alpha^{t}$) as the angle made by the two tracks.

To compute this angle, we use the SatDepth Maps of the pair of images ($I_i,\,I_j$). We first read a set of world points $\{\boldsymbol{X}_i^k,\,\boldsymbol{X}_j^k\}$ corresponding to the middle row of the two images. Then we convert the $(lat,\,lon)$ angular coordinates to $XY$ projected coordinates using the Universal Transverse Mercator (UTM) coordinate system. The $XY$ coordinates are subsequently used to estimate the equation of a 2D line (using linear least squares) for both the images. Then we compute the direction vector of this line, $\boldsymbol{t}_i$, such that it is oriented along the positive x-axis of the image. Finally, we compute the angle between the two line direction vectors to get $\alpha^{t}_{ij}$. The resulting angle is in the range $\alpha^t \in [0,\,\pi]$.

\newpage
\subsection{Dataset Splits}\label{sec:dataset_splits}
\begin{figure}[!h]
  \centering
  \includegraphics[width=0.5\linewidth]{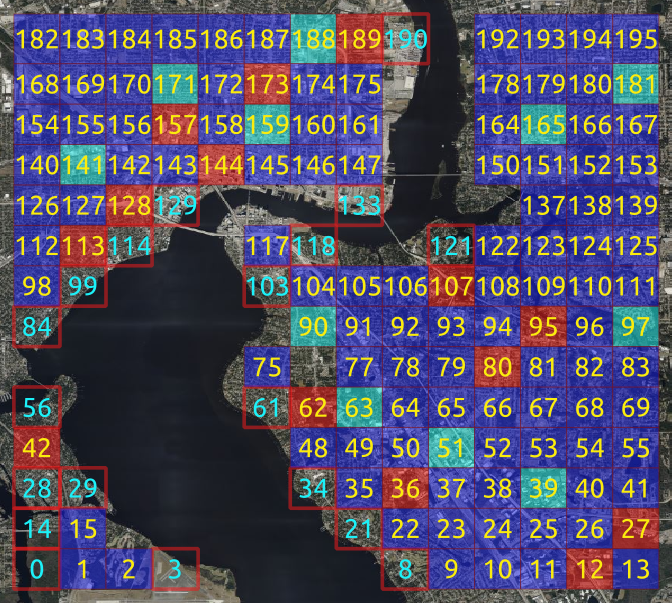}
  \caption{Jacksonville dataset split: The dataset is split into training ({\color{blue}blue} fill), validation ({\color{cyan}cyan} fill) and test ({\color{red}red} fill) sets. We don't use the partial water tiles (empty fill with {\color{cyan}cyan} text) for training, validation or testing.}
  \label{fig:jax_dataset_splits}
\end{figure}

\begin{figure}[!h]
  \centering
  \begin{subfigure}{0.23\linewidth}
    \includegraphics[width=\linewidth]{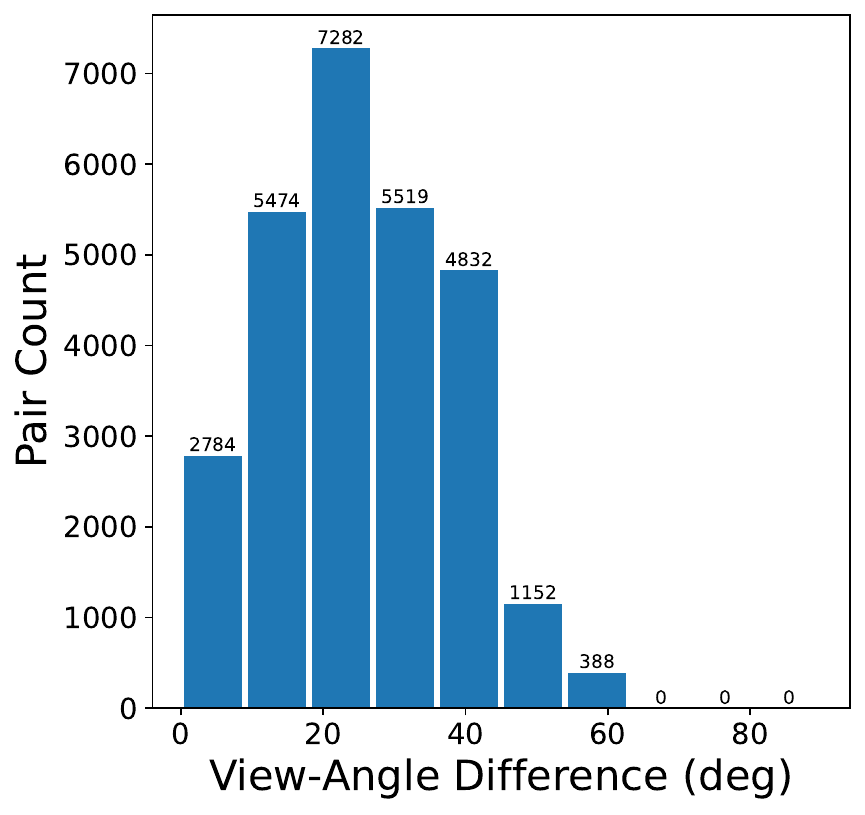}
    \caption{Jacksonville}
  \end{subfigure}
  \begin{subfigure}{0.23\linewidth}
    \includegraphics[width=\linewidth]{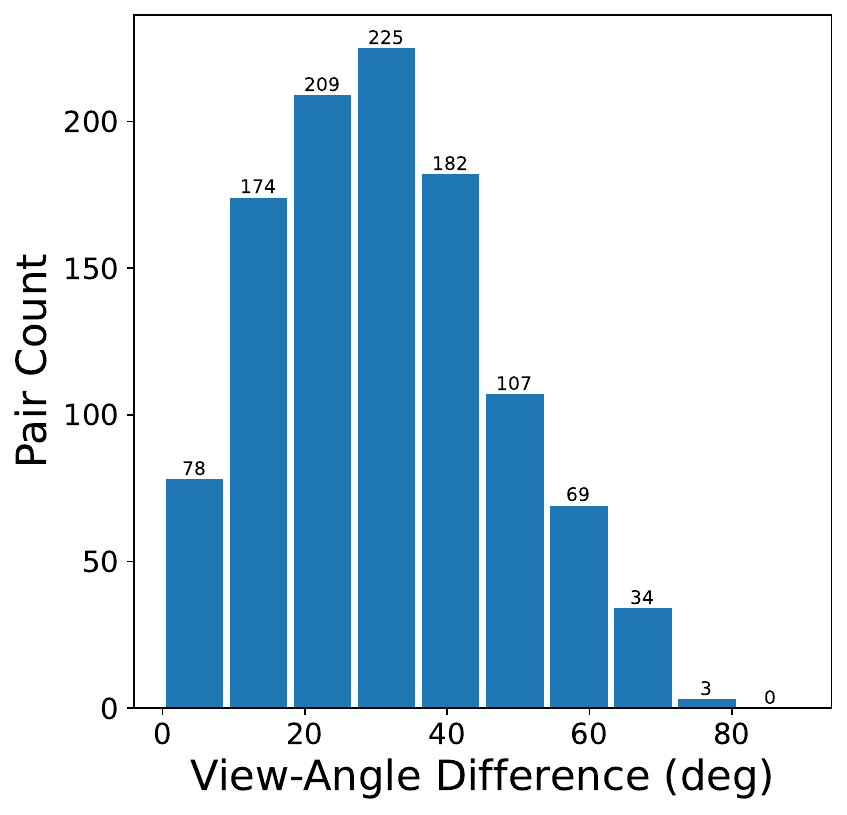}
    \caption{San Fernando}
  \end{subfigure}
  \begin{subfigure}{0.23\linewidth}
    \includegraphics[width=\linewidth]{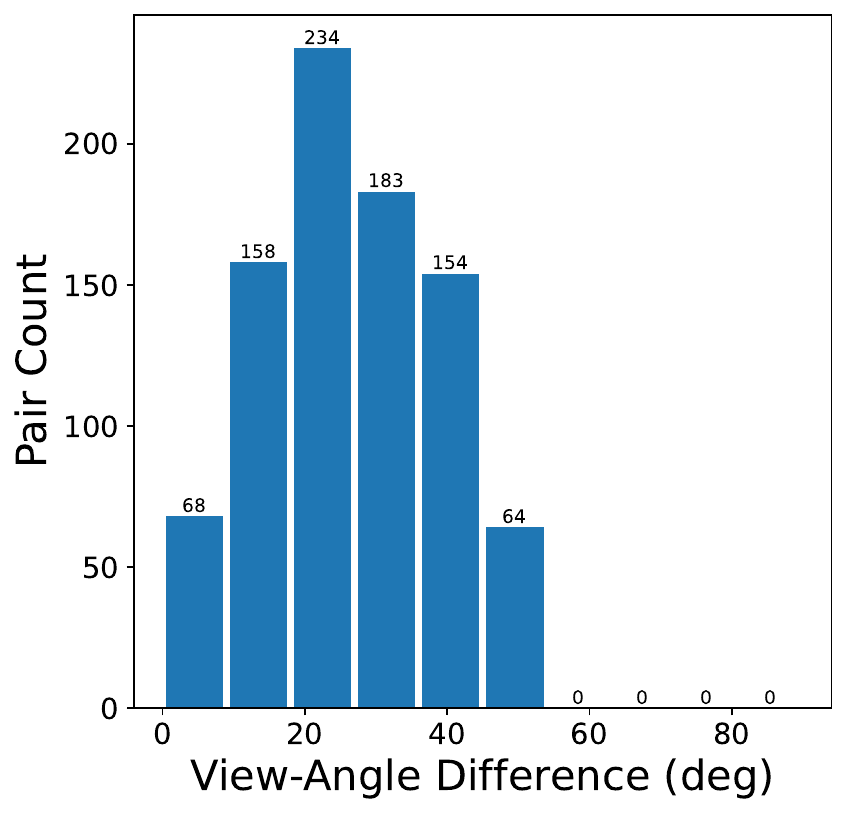}
    \caption{Omaha}
  \end{subfigure}
  \begin{subfigure}{0.23\linewidth}
    \includegraphics[width=\linewidth]{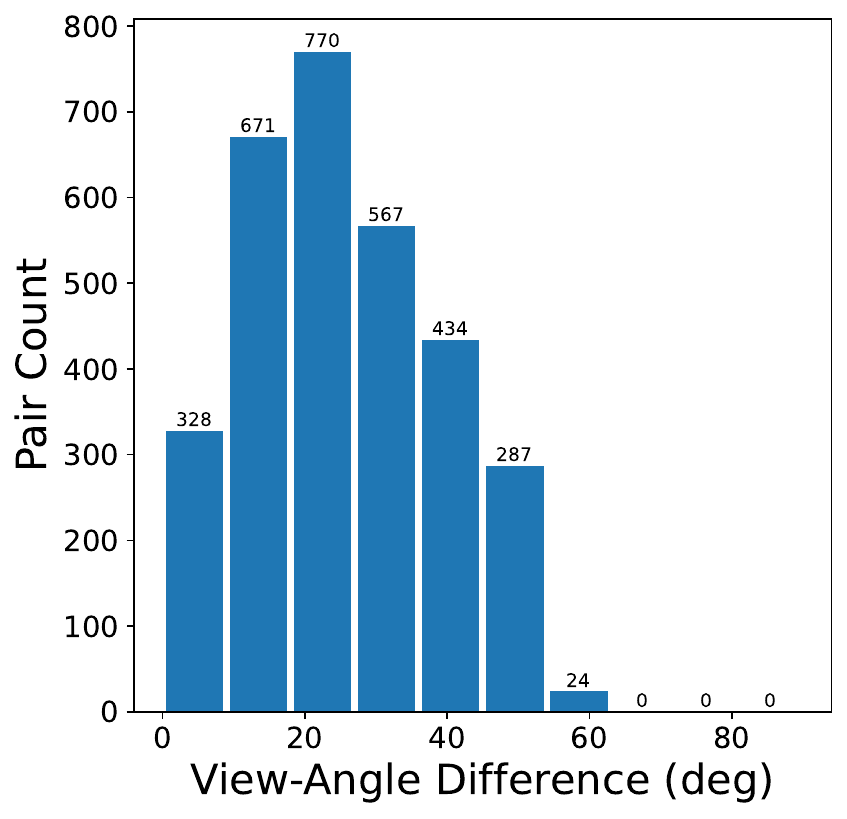}
    \caption{UCSD}
  \end{subfigure}
\caption{Raw image pair distribution with respect to view-angle difference.}
\label{fig:view_int_pairs}
\end{figure}

In the main manuscript, we discussed how Jacksonville tiles were split into three groups (training, validation, and testing). We provide further details in this section. For Jacksonville, we initially split the $\sim200$ \sqkm AOI into 196 (14\cross 14 grid) 1 km \cross 1 km tiles. Out of these 196 tiles, 53 tiles were discarded either due to total water coverage (52 tiles) or failure during bundle adjustment (1 tile). The remaining 143 tiles included 19 tiles with partial water coverage, which were not used for training, validation, or testing. We split the remaining 124 tiles into training, validation, and testing sets with a ratio of roughly 80:10:10. The tile split is visually shown in \cref{fig:jax_dataset_splits}. Additionally, we provide a summary of the raw image pairs that can be constructed using this dataset split in \cref{tab:jax_split_summary}. However, the raw image pairs are highly imbalanced with respect to view-angle and track-angle differences, as shown in \cref{fig:view_int_pairs} and \cref{fig:rel_track_pairs}.

\begin{figure}[!h]
  \centering
  \begin{subfigure}{0.3\linewidth}
    \includegraphics[width=\linewidth]{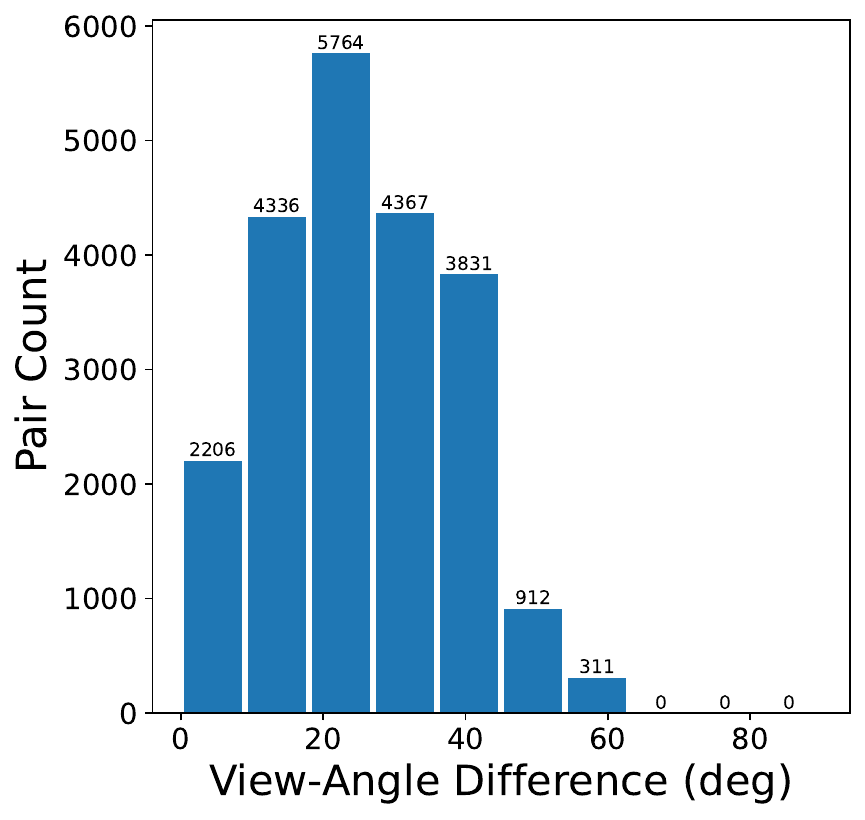}
    \caption{Raw Pairs}
    \label{fig:jax_raw_pairs}
  \end{subfigure}
  \begin{subfigure}{0.3\linewidth}
    \includegraphics[width=\linewidth]{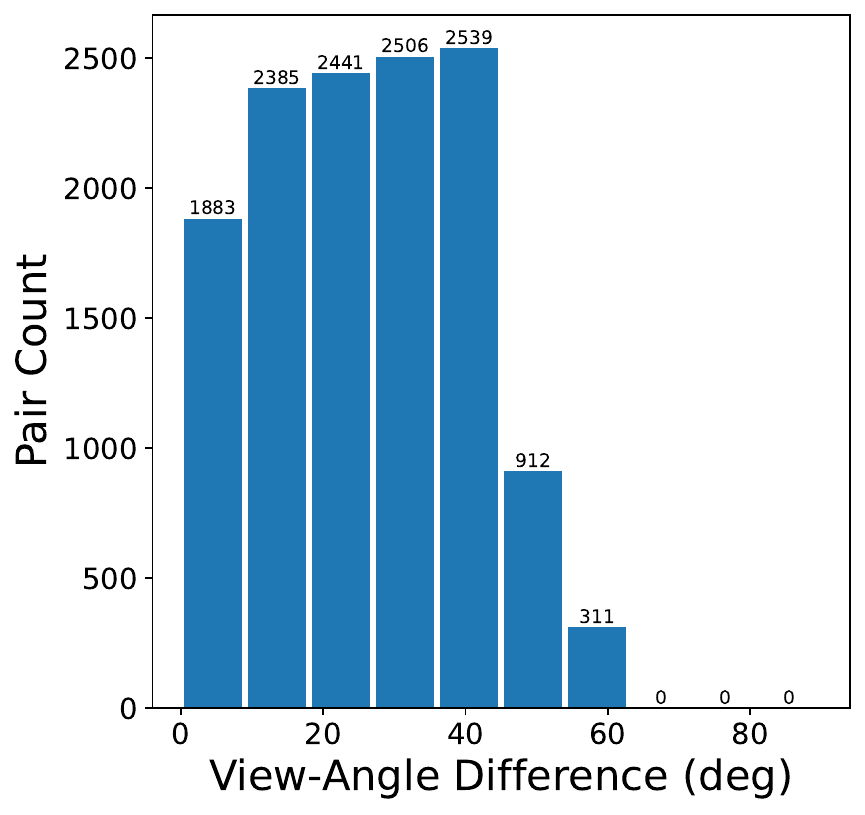}
    \caption{Balanced Pairs}
    \label{fig:jax_balanced_pairs}
  \end{subfigure}
  \caption{Image pair balancing with respect to view-angle difference for Jacksonville training set.}
  \label{fig:jax_pairs_balancing}
\end{figure}

To mitigate the imbalance arising from view-angle differences, we first create a histogram of the view-angle differences with a fixed number of bins ($N_{bin}=10$) using all the raw image pairs. Using the histogram, we randomly select a ``target'' number of pairs from each bin. If a bin has fewer pairs than the target, we include all pairs from that bin to our selection. This strategy of uniformly sampling from a histogram results in a more balanced set of pairs, as shown in \cref{fig:jax_pairs_balancing}.

This balancing procedure is applied to all the image pairs per tile because each tile may have a different number of images, based on which images passed bundle adjustment. Therefore, a perfectly uniform distribution is not observed in \cref{fig:jax_balanced_pairs}. Additionally, it's important to note that the limited number of images for Jacksonville impacts the balancing procedure, which would achieve a more uniform distribution if there were a greater number of images.

Finally, to mitigate the imbalance arising from track-angle differences, we use rotation augmentation during training. We discuss this procedure in detail in the next section.

\begin{table}[!h]
  \centering
  \begin{tabular}{cccc}
  \toprule
  Dataset Split & \# Tiles & \# Raw Image Pairs &\# Balanced Image Pairs\\
  \midrule
  Training  & 99 & 21727 & 12977\\
  Validation & 11 & 2466 & 1471\\
  Testing & 14 & 3238 & -\\
  Partial Water & 19 & - & -\\
  \midrule
  Total & 143 & 27431 & 14448\\
  \bottomrule
  \end{tabular}
  \caption{Jacksonville dataset split summary}
  \label{tab:jax_split_summary}
\end{table}

\clearpage
\subsection{Rotation Augmentation} 
\begin{figure}[!h]
  \centering
  \begin{subfigure}{0.22\linewidth}
    \includegraphics[width=\linewidth]{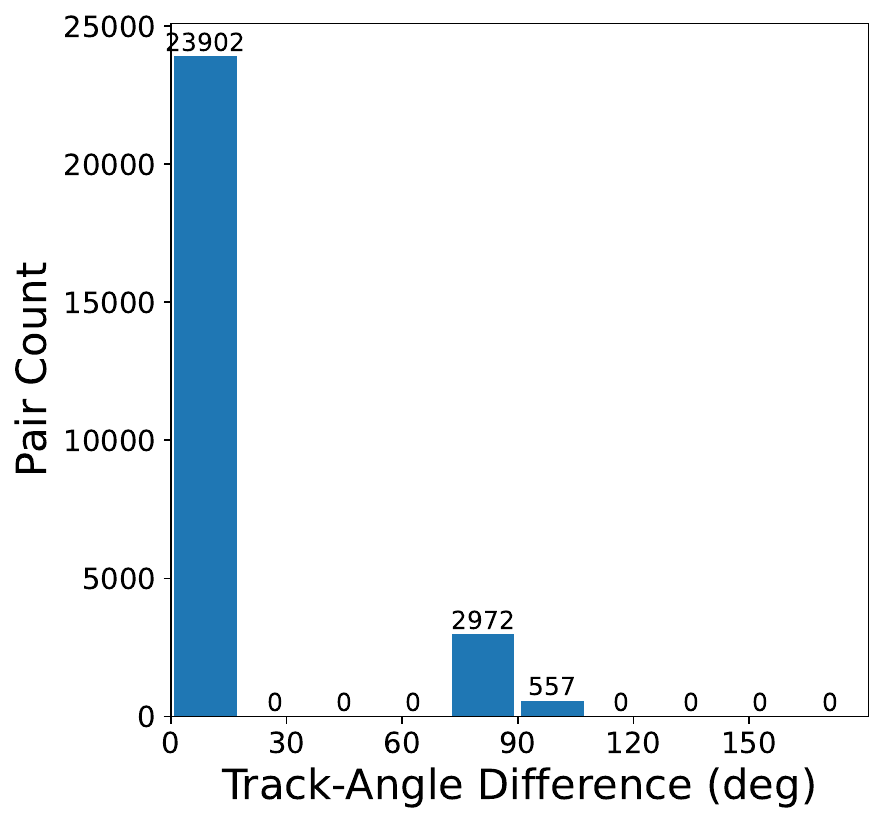}
    \caption{Jacksonville}
    \label{fig:rel_track_jax}
  \end{subfigure}
  \begin{subfigure}{0.22\linewidth}
    \includegraphics[width=\linewidth]{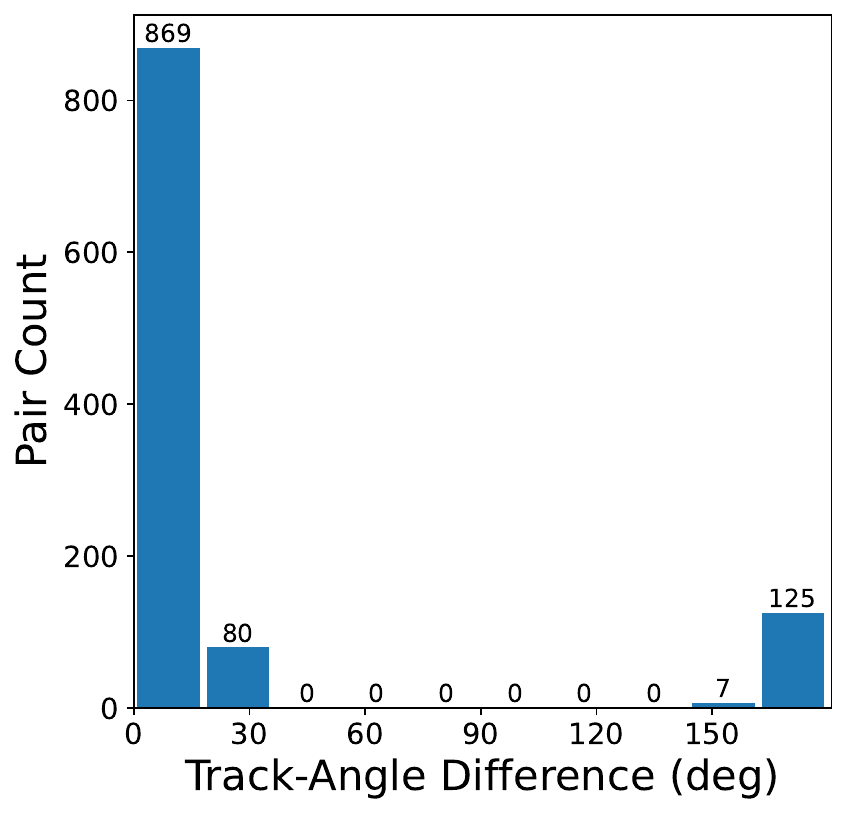}
    \caption{San Fernando}
    \label{fig:rel_track_argentina}
  \end{subfigure}
  \begin{subfigure}{0.22\linewidth}
    \includegraphics[width=\linewidth]{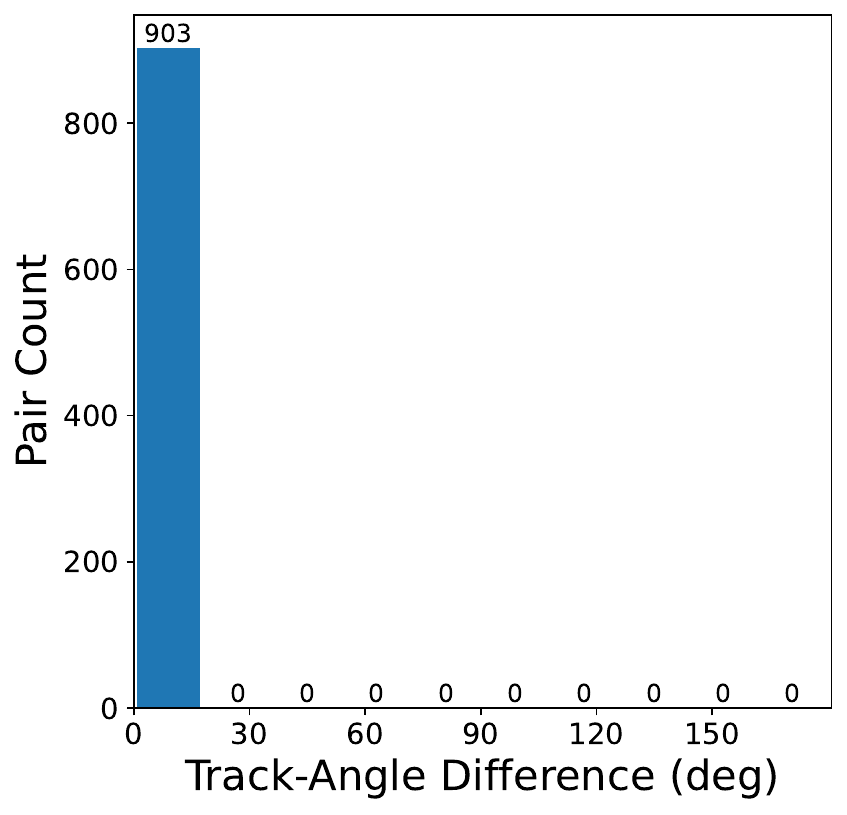}
    \caption{Omaha}
    \label{fig:rel_track_omaha}
  \end{subfigure}
  \begin{subfigure}{0.22\linewidth}
    \includegraphics[width=\linewidth]{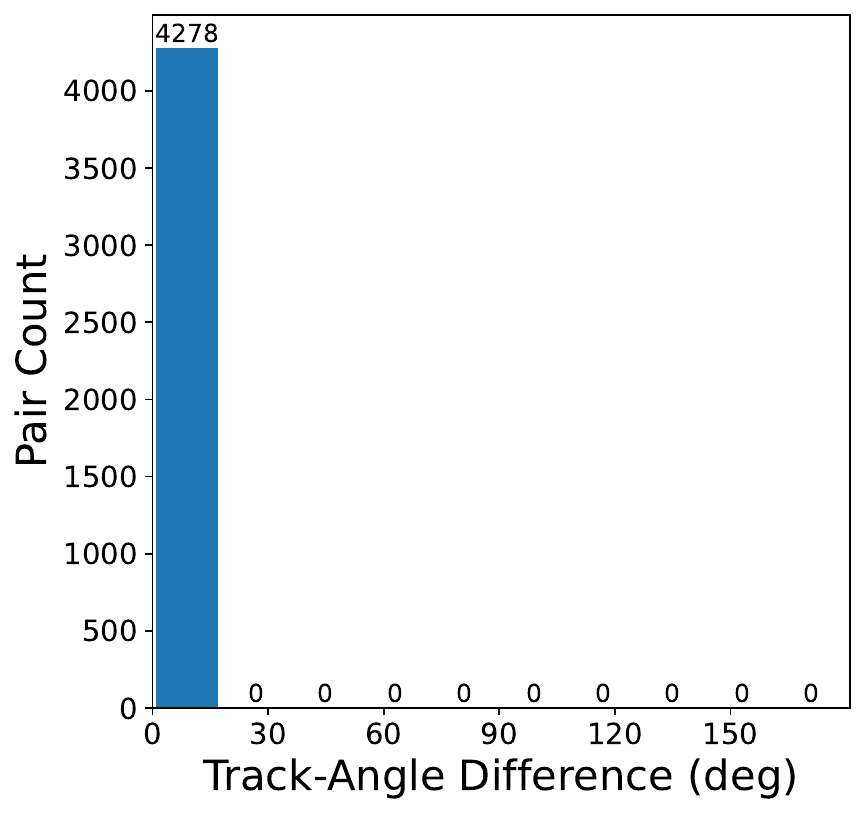}
    \caption{UCSD}
    \label{fig:rel_track_ucsd}
  \end{subfigure}
  \caption{Raw image pair distribution with respect to track-angle difference.}
\label{fig:rel_track_pairs}
\end{figure}

On account of how the high-resolution imaging satellites are operated, the available satellite images are not likely to be uniformly distributed over all possible track-angle differences. This results in training data that is highly imbalanced with respect to the track-angle differences, 
as shown in \cref{fig:rel_track_pairs}. To tackle this issue, we employ rotation augmentation during training.

Our objective with rotation augmentation is to simulate the rotation of a satellite (by an angle $\theta_r$) about its camera viewing axis. 
The rotation augmentation procedure involves transforming the image, its corresponding SatDepth Maps, and the camera.
Since we train the networks with image patches, we utilize the affine camera $\widehat{\mathcal{P}}$ associated with the image patches.

\begin{figure}[!h]
  \centering
  \includegraphics[width=0.7\linewidth]{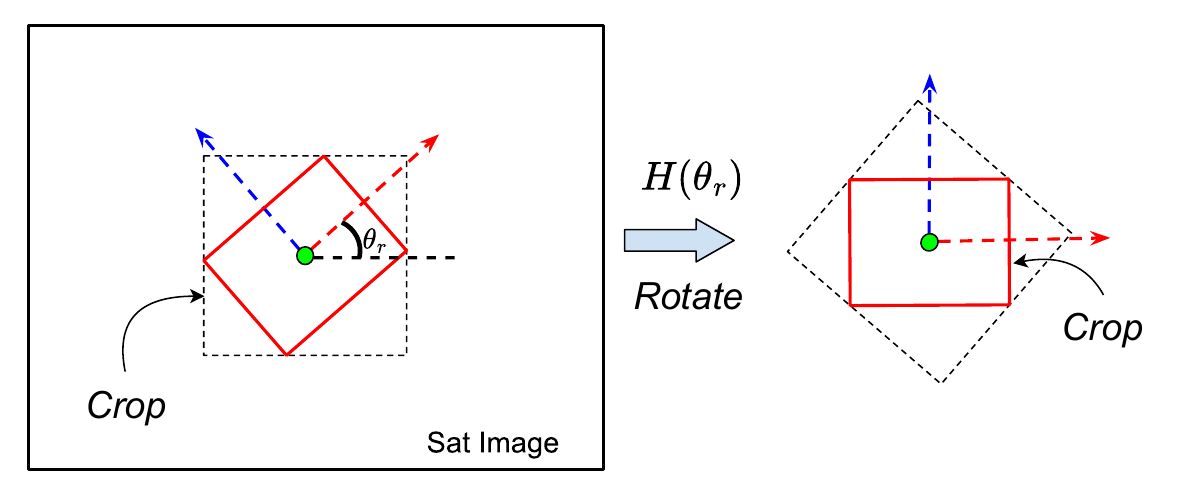}
  \caption{The \textit{``crop-rotate-crop''} procedure for rotation augmentation: Given a patch center ({\color{green}green} dot) and the rotation angle ($\theta_r$), we create a rotated window ({\color{red}red} box) centered at the patch center and compute its bounding box (dotted black box). We crop the image (solid black box) to the bounding box, rotate the cropped image using $\theta_r$ and then crop the new image using the {\color{red}red} box.}
  \label{fig:rotation_aug_crop}
\end{figure}

In rotation augmentation, we begin by sampling a random 3D point ($\boldsymbol{X}$) using the DSM and project it into an image to obtain the corresponding patch center $\boldsymbol{x} = \mathcal{P}(\boldsymbol{X})$. Subsequently, we create a $p \times p$ rotated window centered at $\boldsymbol{x}$ as shown in \cref{fig:rotation_aug_crop}, determine its bounding box, and \textit{crop} the image and corresponding SatDepth Maps to this bounding box. Concurrently, we compute the affine camera $\widehat{\mathcal{P}}$ centered at $\boldsymbol{X}$ according to \cref{eq:rpc_affine}. Next, we pre-multiply $\widehat{\mathcal{P}}$ with a homography ($T_1$) that simulates shifting the origin to the top-left corner of the bounding box. The camera for the bounding box is now represented by $T_1\, \widehat{\mathcal{P}}$. We then \textit{rotate} the image and corresponding SatDepth Maps using a rotation homography ($H(\theta_r)$). The camera for the rotated image becomes $H(\theta_r)\, T_1\, \widehat{\mathcal{P}}$.

\begin{figure}[!h]
  \centering
  \begin{subfigure}{0.45\linewidth}
    \includegraphics[width=\linewidth]{images/rot_aug/pt_matches_0to1_theta_45}
    \caption{45\degree}
    \label{fig:rot_aug_45}
  \end{subfigure}
  \hfill
  \begin{subfigure}{0.45\linewidth}
    \includegraphics[width=\linewidth]{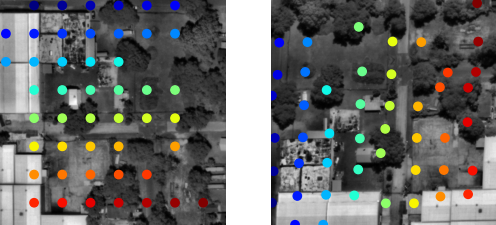}
    \caption{90\degree}
    \label{fig:rot_aug_90}
  \end{subfigure}

  \begin{subfigure}{0.45\linewidth}
    \includegraphics[width=\linewidth]{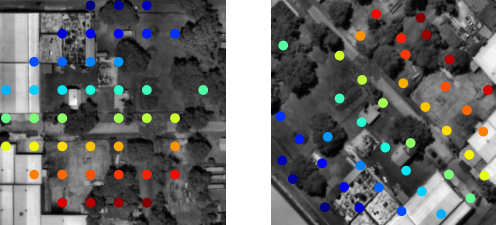}
    \caption{135\degree}
    \label{fig:rot_aug_135}
  \end{subfigure}
  \hfill
  \begin{subfigure}{0.45\linewidth}
    \includegraphics[width=\linewidth]{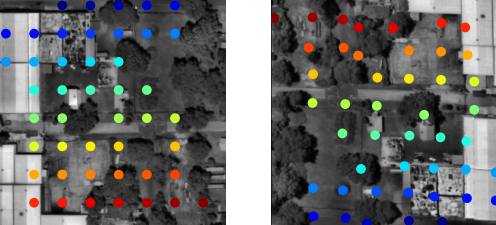}
    \caption{180\degree}
    \label{fig:rot_aug_180}
  \end{subfigure}

  \begin{subfigure}{0.45\linewidth}
    \includegraphics[width=\linewidth]{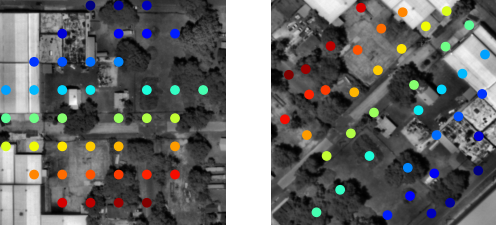}
    \caption{225\degree}
    \label{fig:rot_aug_225}
  \end{subfigure}
  \hfill
  \begin{subfigure}{0.45\linewidth}
    \includegraphics[width=\linewidth]{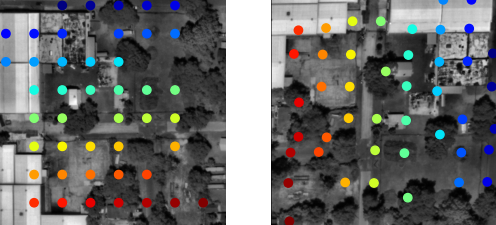}
    \caption{270\degree}
    \label{fig:rot_aug_270}
  \end{subfigure}

  \begin{subfigure}{0.45\linewidth}
    \includegraphics[width=\linewidth]{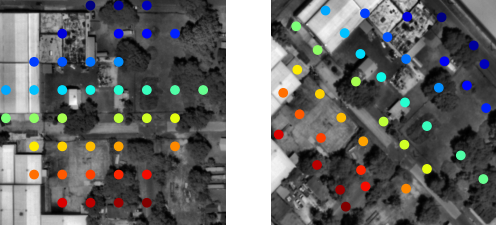}
    \caption{315\degree}
    \label{fig:rot_aug_315}
  \end{subfigure}
  \hfill
  \begin{subfigure}{0.45\linewidth}
    \includegraphics[width=\linewidth]{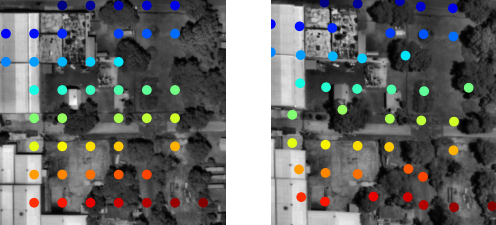}
    \caption{360\degree}
    \label{fig:rot_aug_360}
  \end{subfigure}
\caption{Rotation augmentation examples for different angles. For each plot, the left image is the reference image and the right image is the rotated image. We display the point correspondences, computed using the camera models and the SatDepth Maps, between the two images with the help of colored dots.}
\label{fig:rot_aug_all}
\end{figure}

Finally, we \textit{crop} the rotated image using the $p \times p$ window and simultaneously pre-multiply the camera with another homography ($T_2$) to simulate shifting the origin to the top-left corner of the window. The final camera is given by $T_2\, H(\theta_r)\, T_1\, \widehat{\mathcal{P}}$. We refer to this procedure as \textit{``crop-rotate-crop''} and visually depict the procedure in \cref{fig:rotation_aug_crop}. A few examples of rotation augmented image pairs along with ground-truth correspondences extracted from them are shown in \cref{fig:rot_aug_all} along with average relative 3D error associated with the matches \cref{tab:rot_aug_error_for_correspondences}.

Please note that the three homographies ($T_1$, $H(\theta_r)$ and $T_2$) used in our explanation above belong to the Euclidean transformation group of homographies \cite{hartley2003multiple}. 

\begin{table}[!h]
  \centering
  \begin{tabular}{ccccccccc}
    \toprule
    $\theta_r$ & 45\degree & 90\degree & 135\degree & 180\degree & 225\degree & 270\degree & 315\degree & 360\degree \\ 
    $\epsilon_{3D}^r$(m) & 0.2679 & 0.3192 & 0.3403 & 0.3210 & 0.3613 & 0.3480 & 0.3087 & 0.2821 \\ 
    \bottomrule
  \end{tabular}
  \caption{Average relative 3D error for matches shown in \cref{fig:rot_aug_all}}
  \label{tab:rot_aug_error_for_correspondences}
\end{table}

\clearpage
\subsection{Symmetric Epipolar Distance}
\begin{figure}[!h]
  \centering
  \includegraphics[width=0.6\linewidth]{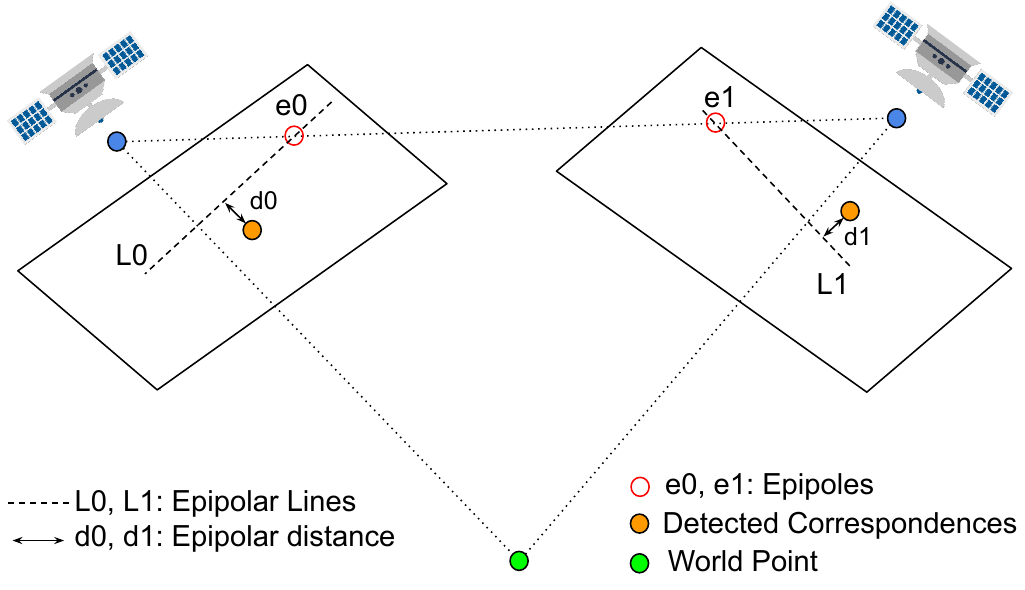}
  \caption{Symmetric epipolar distance is given by $d_{epi}= (d_0 + d_1)/2$, where $d_0,\,d_1$ are the epipolar distances in the images $I_0,\, I_1$ respectively as shown in the figure}
  \label{fig:symmetric_epipolar_distance}
\end{figure}

For computing the precision of matches, we use the symmetric epipolar distance $d_{epi}$ with a threshold $\delta_{epi}$. To compute $d_{epi}$ for a correspondence ($\boldsymbol{x}_i \leftrightarrow \boldsymbol{x}_j$) over a pair of image patches ($I_i,\, I_j$) with cameras ($\widehat{\mathcal{P}}_i,\,\widehat{\mathcal{P}}_j$), we first compute the affine fundamental matrix ($\widehat{F}$) \cite{hartley2003multiple}. Then using the affine fundamental matrix, the symmetric epipolar distance is calculated using \cref{eq:symmetric_epipolar_distance}. This distance is also shown pictorially in \cref{fig:symmetric_epipolar_distance}.

\begin{equation}
  d_{epi} = \frac{1}{2}(\boldsymbol{x}_i^T\widehat{F}\boldsymbol{x}_j)^2 \Bigl( \frac{1}{ (\widehat{F}\boldsymbol{x}_i)_1^2 + (\widehat{F}\boldsymbol{x}_i)_2^2} + \frac{1}{ (\widehat{F}\boldsymbol{x}_j)_1^2 + (\widehat{F}\boldsymbol{x}_j)_2^2} \Bigr)
  \label{eq:symmetric_epipolar_distance}
\end{equation}

where $\boldsymbol{x}_i$ is the pixel coordinate (homogeneous coordinates) of the correspondence in image $I_i$ and $(\widehat{F}\boldsymbol{x}_i)_k$ indicates the $k^{th}$ component of the vector $\widehat{F}\boldsymbol{x}_i$. 

\subsection{Implementation Details}\label{sec:implementation_details}
In the main manuscript \cite{satdepth_preprint}, we gave a brief description of the implementation details for training the three benchmarked networks \cite{Sun_CVPR21_LOFTR,Li_NIPS20_DUALRC,Qianqian_ECCV20_CAPS,wang2022matchformer}. In this section, we provide additional details on the changes made to the three networks to adapt their training and evaluation procedure to the SatDepth dataset.

For fair comparison, we trained the three models with a patch size of $p=448$ for 30 epochs without any hyperparameter tuning. We indicate the models trained on SatDepth by prepending ``sat'' to the model name. Although the original models were trained with image sizes of 640\cross480 on multiple GPUs, we chose the patch size as 448 based on our computing constraints (2 RTX A5000 GPUs). 

We provide a detailed list of the changes made for each model below. The rest of the training and evaluation procedure was kept the same as  in the original implementations.

To adapt the three networks for training with SatDepth dataset, we had to make changes to the data loading, batch size, matching layer, supervision, and training duration. We provide a detailed list of the changes made for each model below. The rest of the training and evaluation procedure was kept the same as  in the original implementations.

\begin{figure}[!h]
  \centering
  \begin{subfigure}{0.25\linewidth}
    \includegraphics[width=\linewidth]{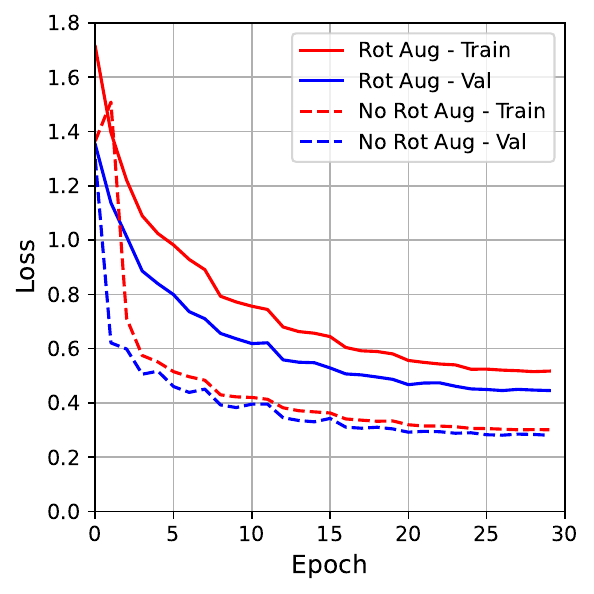}
    \caption{Training and Validation Loss for satLoFTR}
    \label{fig:loftr_convergence}
  \end{subfigure}
  \hspace{2em}
  \begin{subfigure}{0.25\linewidth}
    \includegraphics[width=\linewidth]{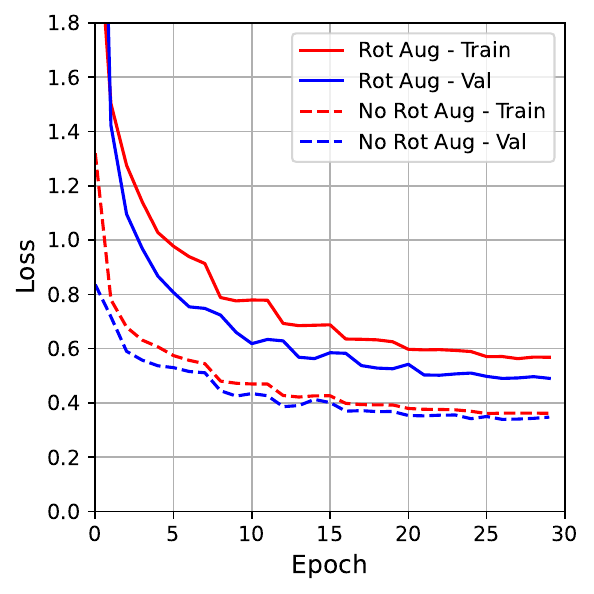}
    \caption{Training and Validation Loss for satMatchFormer}
    \label{fig:matchformer_convergence}
  \end{subfigure}
  \caption{Training and Validation Loss convergence for satLoFTR and satMatchFormer for both with and without rotation augmentation.}
  \label{fig:convergence_plots}
\end{figure}

\vspace*{1em}
\noindent\textbf{LoFTR} \cite{Sun_CVPR21_LOFTR}:
\begin{itemize}
  \item \textbf{Data Loading}: LoFTR was trained on grayscale versions of images from the MegaDepth dataset \cite{MegaDepthLi18}. We train satLoFTR on the grayscale PAN band of the satellite images using the SatDepth dataset.
  \item \textbf{Batch Size}: To train satLoFTR, we used a total batch size of 6 (\ie 3 per RTX A5000 GPU).
  \item \textbf{Matching Layer}: LoFTR has the option to use either \textit{dual-softmax} or \textit{optimal-transport} for the coarse matching layer. For benchmarking, we trained satLoFTR using the \textit{optimal-transport} layer. 
  \item \textbf{Supervision}: To train satLoFTR, we extract matching correspondences using SatDepth Maps with $\delta_{3D}=1.0$ m. We explained the correspondence extraction procedure in the ``Depth Map Generation'' section of our paper \cite{satdepth_preprint}.
  \item \textbf{Baseline Model}: We use the outdoor model of LoFTR which was trained on the MegaDepth dataset as the baseline. This baseline model was also trained using the \textit{optimal-transport} coarse matching layer.
  \item \textbf{Model Convergence} : We provide the convergence plots for satLoFTR in \cref{fig:loftr_convergence}.
\end{itemize}

\vspace*{1em}
\noindent\textbf{MatchFormer} \cite{wang2022matchformer}:
\begin{itemize}
  \item MatchFormer has a different architecture than LoFTR but MatchFormer uses LoFTR codebase for tranining. As a result we use similar choices as that were made for satLoFTR for training satMatchFormer. We provide the details below.
  \item \textbf{Model Config}: MatchFormer provides four different model configurations -- ``LA\_small'', ``LA\_large'', ``SA\_small'', and ``SA\_large''. We use the ``LA\_large'' model for training satMatchFormer.
  \item \textbf{Data Loading}: MatchFormer was trained on grayscale versions of images from the MegaDepth dataset \cite{MegaDepthLi18}. We train satMatchFormer on the grayscale PAN band of the satellite images using the SatDepth dataset.
  \item \textbf{Batch Size}: To train satMatchFormer, we used a total batch size of 4 (\ie 2 per RTX A5000 GPU).
  \item \textbf{Matching Layer}: satMatchFormer was trained using \textit{dual-softmax} matching layer.
  \item \textbf{Supervision}: To train satMatchFormer, we extract matching correspondences using SatDepth Maps with $\delta_{3D}=1.0$ m. We explained the correspondence extraction procedure in the ``Depth Map Generation'' section of our paper \cite{satdepth_preprint}.
  \item \textbf{Baseline Model}: We use the ``LA\_large'' model of MatchFormer as the baseline.
  \item \textbf{Model Convergence} : We provide the convergence plots for satMatchFormer in \cref{fig:matchformer_convergence}.
\end{itemize}

\vspace*{1em}
\noindent\textbf{DualRC-Net} \cite{Li_NIPS20_DUALRC}:
\begin{itemize}
  \item \textbf{Data Loading}: DualRC-Net was trained using RGB images of the MegaDepth dataset. To train satDualRC-Net, we adapted the data loading process to grayscale PAN band satellite images without changing the architecture by repeating the grayscale input three times.
  \item \textbf{Batch Size}: satDualRC-Net was trained using a batch size of 24 on a single RTX A5000 GPU.
  \item \textbf{Supervision}: To train satDualRC-Net, we extract matching correspondences using SatDepth Maps with $\delta_{3D}=1.0$ m. We explained the correspondence extraction procedure in the ``Depth Map Generation'' section of our paper \cite{satdepth_preprint}. satDualRC-Net was trained using 128 randomly collected correspondences.
  \item \textbf{Training Duration}: We train satDualRC-Net for 30 epochs.
  \item \textbf{Baseline Model}: We use the DualRC-Net model which was trained on the MegaDepth dataset as the baseline.
\end{itemize}

\vspace*{1em}
\noindent\textbf{CAPS} \cite{Qianqian_ECCV20_CAPS}:
\begin{itemize}
  \item \textbf{Data Loading}: CAPS was trained using RGB images of MegaDepth \cite{MegaDepthLi18}. To train satCAPS, we adapted the data loading process to grayscale PAN band satellite images without changing the architecture by repeating the grayscale input three times.
  \item \textbf{Batch Size}: satCAPS was trained using a batch size of 16 on a single RTX A5000 GPU.
  \item \textbf{Supervision}: satCAPS was trained using a 50:50 ratio of SIFT to random query points. We use a total of 500 number of query points for each image pair.
  \item \textbf{Training Duration}: We train satCAPS for 30 epochs.
  \item \textbf{Baseline Model}: We use the CAPS model which was trained on the MegaDepth dataset as the baseline.
\end{itemize}

\clearpage
\section{Additional Results}\label{sec:additional_results_appendix}
\begin{figure*}[ht!]
    \centering
    \input{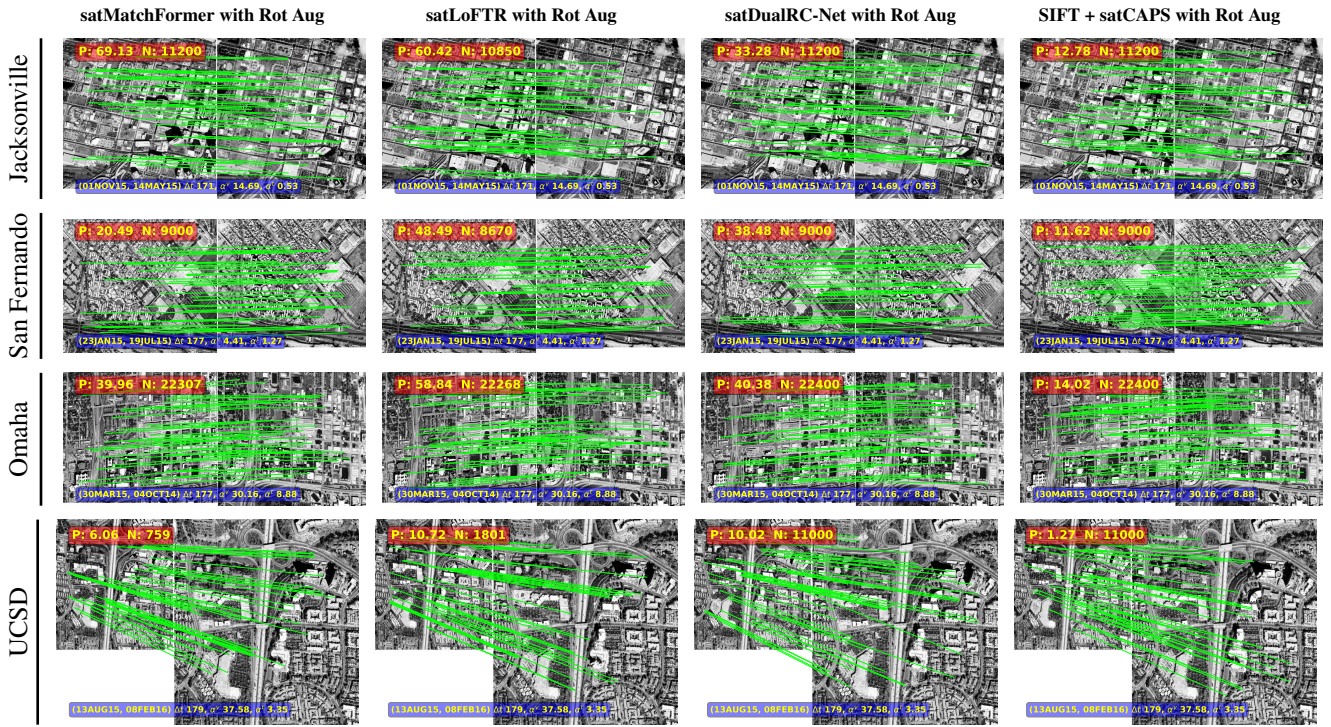}
  \caption{Qualitative results for \textbf{large time difference} ($\Delta t$) for models trained on SatDepth with rotation augmentation. Precision (P) and number of matches (N) are displayed at the top of each plot. Image pair names, time difference ($\Delta t$), view-angle difference ($\alpha^v$), and track-angle difference ($\alpha^t$) are displayed at the bottom. The \flgreen{green} lines depict 40 randomly chosen \flgreen{true} matches.}
  \label{fig:qualitative_matching_plots_time_diff}
  \end{figure*}

We present additional qualitative results in \cref{fig:qualitative_matching_plots_intersection} and \cref{fig:qualitative_matching_plots_time_diff}. A summary of the performance comparisons of the different models on Omaha and UCSD AOIs is provided in \cref{tab:matcher_performance_appendix}. 

\begin{table}[!h]
  \centering
    \resizebox{0.45\linewidth}{!}{%
    \setlength\tabcolsep{4.0pt}
    \begin{tabular}{cccccccc}
    \toprule
    \multicolumn{8}{c}{\centering \textbf{Omaha}}\\
    \midrule
    \multirow{2}{*}{\rotatebox[origin=c]{90}{Config}}
    & \multirow{2}{1cm}[-.4em]{Method} 
    & \multirow{2}{*}[-.4em]{\begin{tabular}[c]{@{}c@{}}Rot.\\ Aug.\end{tabular}}
    & \multicolumn{3}{c}{Pose estimation AUC$\uparrow$} 
    & \multirow{2}{*}[-.4em]{\begin{tabular}[c]{@{}c@{}}Precision $\uparrow$\\ \end{tabular}} 
    & \multirow{2}{*}[-.4em]{\begin{tabular}[c]{@{}c@{}} \# Matches $\uparrow$\\ ({\scriptsize{TP}})\end{tabular}}\\
    \cmidrule(lr){4-6}
    & & & @5\degree & @10\degree & @20\degree & &\\
    \midrule   
    \multirow{4}{*}{\rotatebox[origin=c]{90}{Baseline}} 
    & SIFT + CAPS\cite{Qianqian_ECCV20_CAPS}  &\cross & 70.41 & 73.29 & 77.20 & 12.60 & 25 \\
    & DualRC-Net\cite{Li_NIPS20_DUALRC}  &\cross & 69.82 & 72.29 & 75.76 & 26.72 & 53 \\
    & LoFTR\cite{Sun_CVPR21_LOFTR}  &\cross & 82.24 & 86.73 & 90.32 & 49.51 & 91 \\
    & MatchFormer\cite{wang2022matchformer}  &\cross & 86.10 & 90.23 & 93.20 & 53.95 & 107 \\
    \midrule
    \multirow{8}{*}{\rotatebox[origin=c]{90}{Trained on SatDepth}}  
    & SIFT + satCAPS & \cross & 72.16 & 76.01 & 80.74 & 16.13 & 32 \\
    & satDualRC-Net & \cross & 70.69 & 73.70 & 77.66 & 31.55 & 63 \\
    & satLoFTR & \cross & 77.95 & 82.09 & 85.78 & 43.16 & 86 \\
    & satMatchFormer & \cross & 83.24 & 87.97 & 91.56 & 50.11 & 99 \\
    \cmidrule{2-8}
    & SIFT + satCAPS & \checkmark & 69.42 & 71.77 & 75.24 & 10.64 & 21 \\
    & satDualRC-Net & \checkmark & 71.10 & 74.35 & 78.56 & 28.31 & 57 \\
    & satLoFTR & \checkmark & 82.41 & 87.36 & 91.18 & 46.85 & 90 \\
    & satMatchFormer & \checkmark & 81.83 & 86.65 & 90.49 & 44.10 & 86 \\    
    \bottomrule
\end{tabular}
    }
    \resizebox{0.45\linewidth}{!}{%
    \setlength\tabcolsep{4.0pt}
    \begin{tabular}{cccccccc}
    \toprule
    \multicolumn{8}{c}{\centering \textbf{UCSD}}\\
    \midrule
    \multirow{2}{*}{\rotatebox[origin=c]{90}{Config}}
    & \multirow{2}{1cm}[-.4em]{Method} 
    & \multirow{2}{*}[-.4em]{\begin{tabular}[c]{@{}c@{}}Rot.\\ Aug.\end{tabular}}
    & \multicolumn{3}{c}{Pose estimation AUC$\uparrow$} 
    & \multirow{2}{*}[-.4em]{\begin{tabular}[c]{@{}c@{}}Precision $\uparrow$\\ \end{tabular}} 
    & \multirow{2}{*}[-.4em]{\begin{tabular}[c]{@{}c@{}} \# Matches $\uparrow$\\ ({\scriptsize{TP}})\end{tabular}}\\
    \cmidrule(lr){4-6}
    & & & @5\degree & @10\degree & @20\degree & &\\
    \midrule

    \multirow{4}{*}{\rotatebox[origin=c]{90}{Baseline}} 
    & SIFT + CAPS\cite{Qianqian_ECCV20_CAPS} &\cross & 70.14 & 72.77 & 76.35 & 11.32 & 23 \\
    & DualRC-Net\cite{Li_NIPS20_DUALRC}  &\cross & 70.74 & 73.67 & 77.51 & 26.73 & 53 \\
    & LoFTR\cite{Sun_CVPR21_LOFTR}  &\cross & 82.88 & 87.58 & 91.24 & 51.70 & 71 \\
    & MatchFormer\cite{wang2022matchformer} &\cross & 88.12 & 92.41 & 95.25 & 64.23 & 128 \\
    \midrule
    \multirow{8}{*}{\rotatebox[origin=c]{90}{Trained on SatDepth}}  
    & SIFT + satCAPS & \cross & 70.85 & 73.85 & 77.63 & 12.41 & 25 \\
    & satDualRC-Net & \cross & 71.18 & 74.40 & 78.52 & 28.55 & 57 \\
    & satLoFTR & \cross & 70.25 & 72.71 & 75.99 & 28.99 & 57 \\
    & satMatchFormer & \cross & 83.62 & 87.88 & 91.02 & 39.24 & 70 \\
    \cmidrule{2-8}
    & SIFT + satCAPS & \checkmark & 68.68 & 70.47 & 73.25 & 7.86 & 16 \\
    & satDualRC-Net & \checkmark & 70.85 & 73.98 & 78.06 & 24.68 & 49 \\
    & satLoFTR & \checkmark & 76.82 & 80.52 & 83.94 & 38.00 & 66 \\
    & satMatchFormer & \checkmark & 80.10 & 83.96 & 87.15 & 34.75 & 63 \\
    \bottomrule
\end{tabular}
    }
  \caption{Precision and Pose error averaged over all testing image patches for Omaha and UCSD AOIs. Note that these results are for the original data distribution (ref \cref{fig:rel_track_pairs}) and not the simulated rotations. Both Omaha and UCSD do not have large range of track-angle differences.}
  \label{tab:matcher_performance_appendix}
  \end{table}

  \begin{figure*}[ht!]
    \centering
    \input{hard_pairs_intersection_matching_plots_table.tex}
    
    \caption{Qualitative results for \textbf{large view-angle difference} ($\alpha^v$) for models trained on SatDepth with rotation augmentation. Precision (P) and number of matches (N) are displayed at the top of each plot. Image pair names, time difference ($\Delta t$), view-angle difference ($\alpha^v$), and track-angle difference ($\alpha^t$) are displayed at the bottom. The \flgreen{green} lines depict 40 randomly chosen \flgreen{true} matches.}
    \label{fig:qualitative_matching_plots_intersection}
    \end{figure*}

\clearpage
\subsection{Simulated Rotation Experiment}
\begin{figure}[h!]
  \centering
  \input{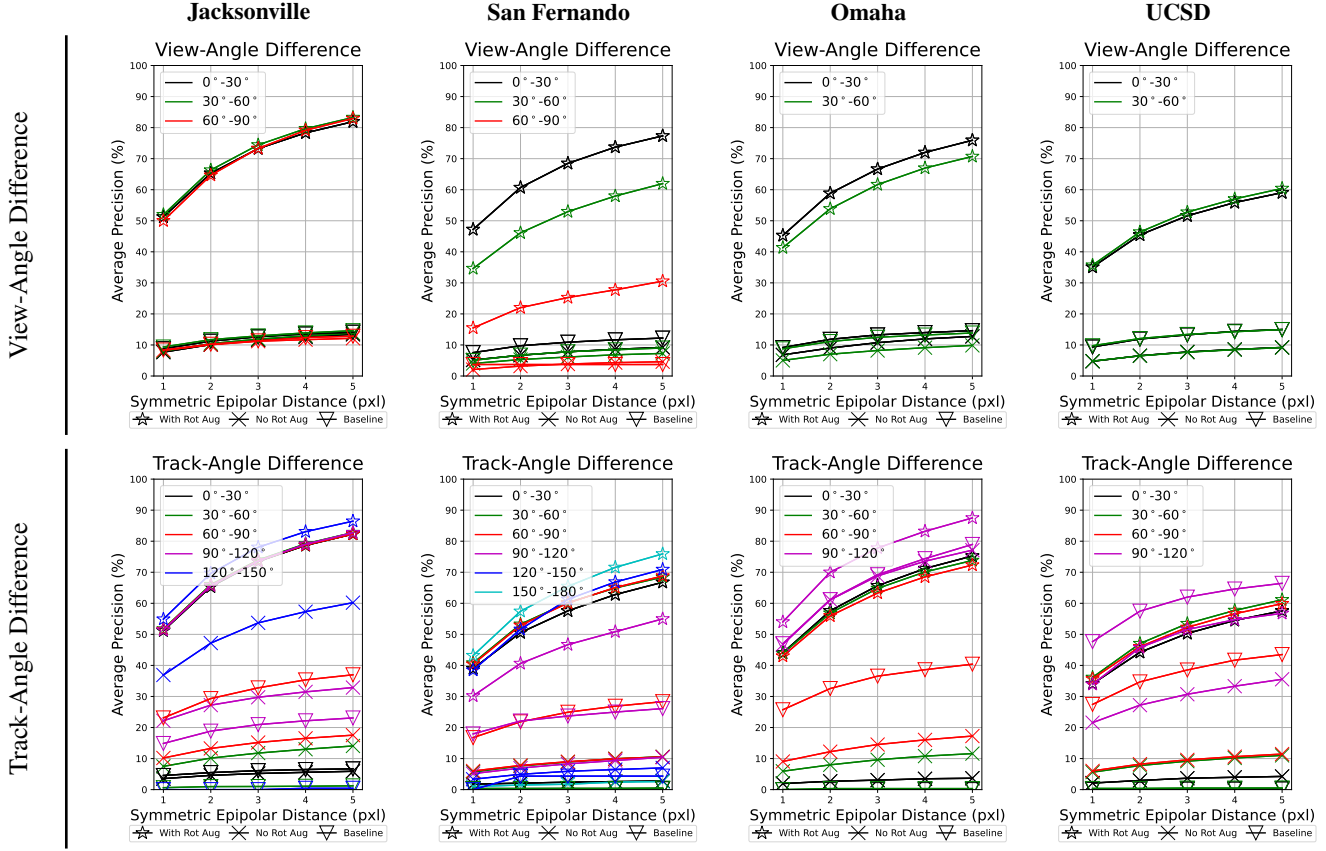}
  \caption{Average Precision for different configurations of LoFTR \cite{Sun_CVPR21_LOFTR} for various view-angle ($\alpha^v$) and randomly simulated track-angle ($\alpha^t$) differences for all testing image pairs. The satLoFTR model trained with rotation augmentation outperforms other configurations across all AOIs, for both View-Angle and Track-Angle Differences.}
  \label{fig:loftr_angle_performance_all_aois_simulated_rot}
\end{figure}

In this section, we present additional plots for the performance of the models for the simulated rotation experiment. We provide the average precision for different configurations of MatchFormer and LoFTR for various view-angle ($\alpha^v$) and randomly simulated track-angle ($\alpha^t$) differences for all testing image pairs in \cref{fig:matchformer_angle_performance_all_aois_simulated_rot} and \cref{fig:loftr_angle_performance_all_aois_simulated_rot} respectively.

\begin{figure}[h!]
  \centering
  \input{./figure_tables/simulated_rot/matchformer_angle_performance_all_aois.tex}
  \caption{Average Precision for different configurations of MatchFormer \cite{wang2022matchformer} for various view-angle ($\alpha^v$) and randomly simulated track-angle ($\alpha^t$) differences for all testing image pairs. The satMatchFormer model trained with rotation augmentation outperforms other configurations across all AOIs, for both View-Angle and Track-Angle Differences.}
  \label{fig:matchformer_angle_performance_all_aois_simulated_rot}
\end{figure}
\clearpage
\section{Datasheet}\label{sec:datasheet}

\definecolor{darkblue}{RGB}{46,25, 110}

\newcommand{\dssectionheader}[1]{%
\phantomsection
   \noindent\framebox[\columnwidth]{%
      {\fontfamily{phv}\selectfont \textbf{\textcolor{darkblue}{#1}}}
   }
  \addcontentsline{toc}{subsection}{#1}
}

\newcommand{\dsquestion}[1]{%
\medskip
\phantomsection
    {\noindent \fontfamily{phv}\selectfont \textcolor{darkblue}{\textbf{#1}}}
  \addcontentsline{toc}{subsubsection}{#1}
\smallskip
    }

\newcommand{\dsquestionex}[2]{%
\medskip
\phantomsection
    {\noindent \fontfamily{phv}\selectfont \textcolor{darkblue}{\textbf{#1} #2}}
  \addcontentsline{toc}{subsubsection}{#1}
    \smallskip
}

This Datasheet for Dataset follows the template from \cite{cornebise2022open}.

\dssectionheader{Motivation}
\dsquestionex{For what purpose was the dataset created?}{Was there a specific task in mind? Was there a specific gap that needed to be filled? Please provide a description.}

The dataset was created to enable research in satellite image matching -- \ie given a pair of satellite images, learn to detect matching pixel correspondences. The dataset was created intentionally with this task in mind as there was no existing satellite image matching dataset.

\dsquestion{Who created this dataset (e.g., which team, research group) and on behalf of which entity (e.g., company, institution, organization)?}

\ifthenelse{\equal{\anonSub}{True}}{
  \TBD
}{
The dataset was created by the authors at Purdue University.
}

\dsquestionex{Who funded the creation of the dataset?}{If there is an associated grant, please provide the name of the grantor and the grant name and number.}

\ifthenelse{\equal{\anonSub}{True}}{
  \TBD
}{
Funding was provided by the Office of the Director of National Intelligence (ODNI),
Intelligence Advanced Research Projects Activity (IARPA),
via Contract \#2021-21040700001.
}

\dsquestion{Any other comments?}

\ifthenelse{\equal{\anonSub}{True}}{
  \TBD
}{
We would like to thank the former members of Robot Vision Lab (RVL), Purdue University and the Applied Research Associates (ARA) team for their collective work towards the IARPA CORE-3D project. Their collective contributions to the project facilitated the development of the processing pipeline used to create this dataset. We would also like to thank the efforts of Accenture Team of IARPA SMART project which led to the collection\cite{accenture_gcps} of the Ground Control Points (GCPs). These GCPs were used to assess the accuracy of the SatDepth dataset.

Finally, we would also like to thank the following members of our lab who assisted with GCP annotations:
\begin{itemize}
    \item Amith Kashyap, PhD student, RVL, Purdue University.
    \item Adrien Jean Guy Dubois, PhD student, RVL, Purdue University.
\end{itemize}
}

\bigskip
\dssectionheader{Composition}
\dsquestionex{What do the instances that comprise the dataset represent (e.g., documents, photos, people, countries)?}{ Are there multiple types of instances (e.g., movies, users, and ratings; people and interactions between them; nodes and edges)? Please provide a description.}

The SatDepth dataset consists of satellite images from four Area of Interests (AOIs). The data has been divided into tiles of size under 2 \sqkm. These tiles can be thought of as instances in the dataset. Each instance contains a collection of satellite images (along with cameras and metadata) and SatDepth Maps.

\dsquestion{How many instances are there in total (of each type, if appropriate)?}

There are a total of 146 tiles (\ie instances) in the dataset as detailed in \cref{tab:satdepth_summary}.

\begin{table}[!h]
  \caption{SatDepth dataset coverage}
  \label{tab:satdepth_summary}
  \centering
  \begin{tabular}{lcccl}
    \toprule    
    AOI Name     & Area (\sqkm) &\# Tiles & \# Images     & Date-Range \\
    \midrule
    Jacksonville, FL & 143 & 143 &26 & 2014-2016 \\
    San Fernando, Argentina    & 1 &1  & 50 & 2014-2016 \\
    Omaha, NE    & 2.2 & 1  & 43 & 2014-2015 \\
    UCSD, CA    & 1  &1 & 79 & 2014-2016 \\
    \bottomrule
  \end{tabular}
\end{table}

\dsquestionex{Does the dataset contain all possible instances or is it a sample (not necessarily random) of instances from a larger set?}{ If the dataset is a sample, then what is the larger set? Is the sample representative of the larger set (e.g., geographic coverage)? If so, please describe how this representativeness was validated/verified. If it is not representative of the larger set, please describe why not (e.g., to cover a more diverse range of instances, because instances were withheld or unavailable).}

\begin{figure}[!h]
  \centering
    \begin{subfigure}[b]{0.22\linewidth}
      \includegraphics[width=\linewidth, height=\linewidth]{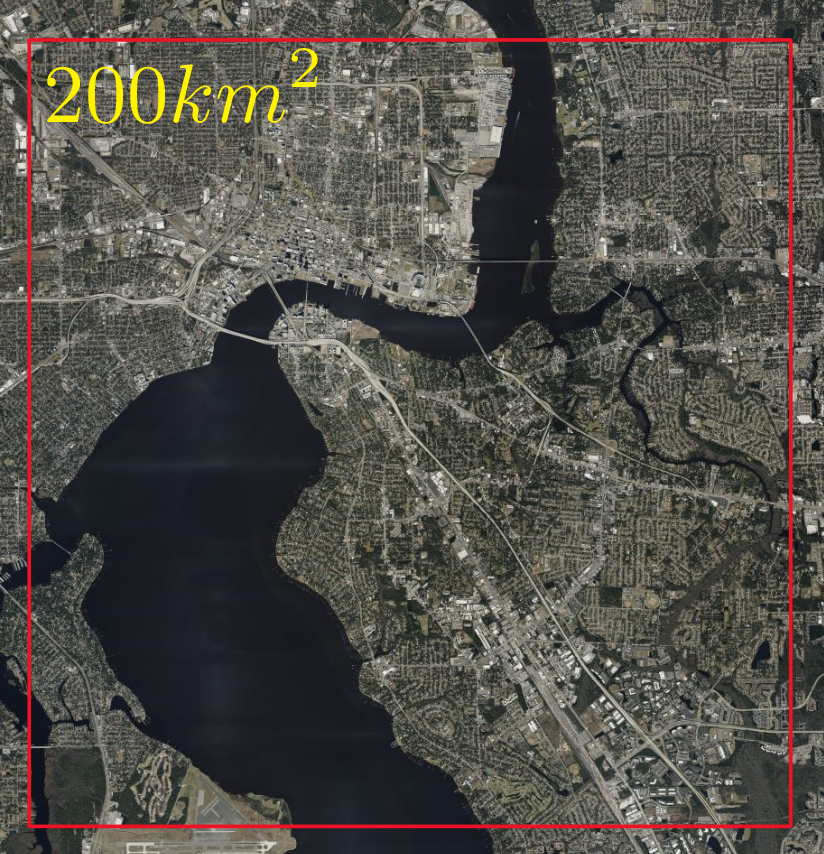}
      \caption{Jacksonville}
    \end{subfigure}
    \begin{subfigure}[b]{0.22\linewidth}
      \includegraphics[width=\linewidth, height=\linewidth]{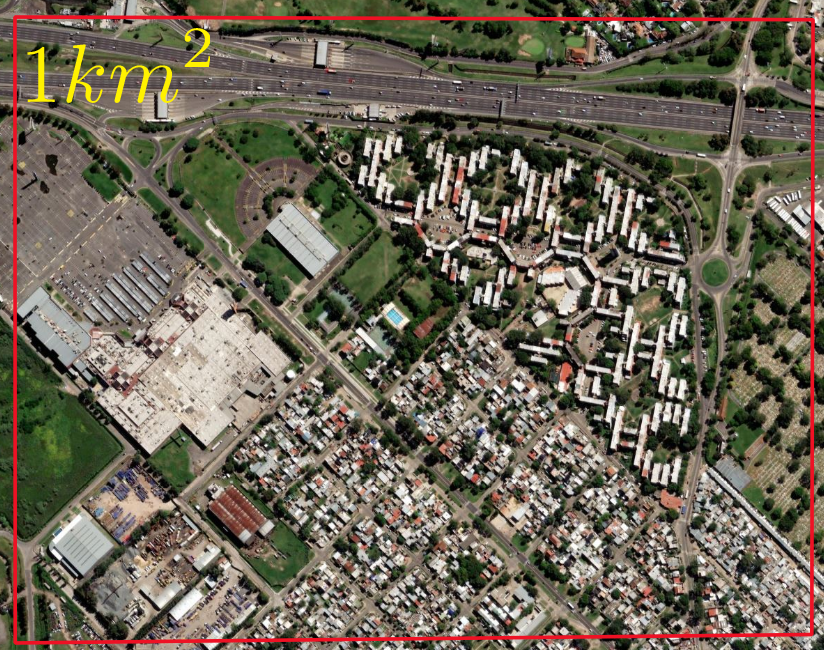}
      \caption{San Fernando}
    \end{subfigure}
    \begin{subfigure}[b]{0.22\linewidth}
      \includegraphics[width=\linewidth, height=\linewidth]{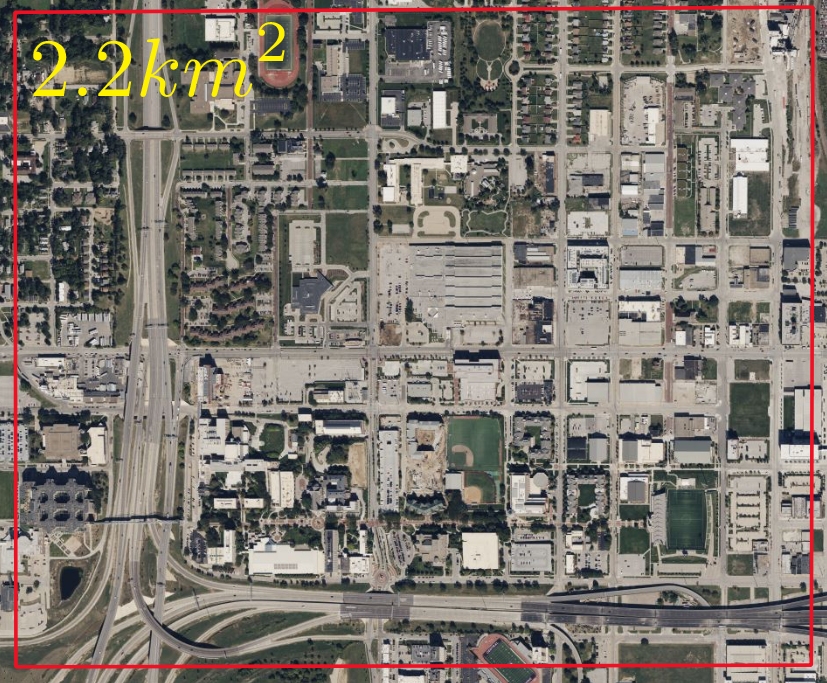}
      \caption{Omaha}
    \end{subfigure}
    \begin{subfigure}[b]{0.22\linewidth}
      \includegraphics[width=\linewidth, height=\linewidth]{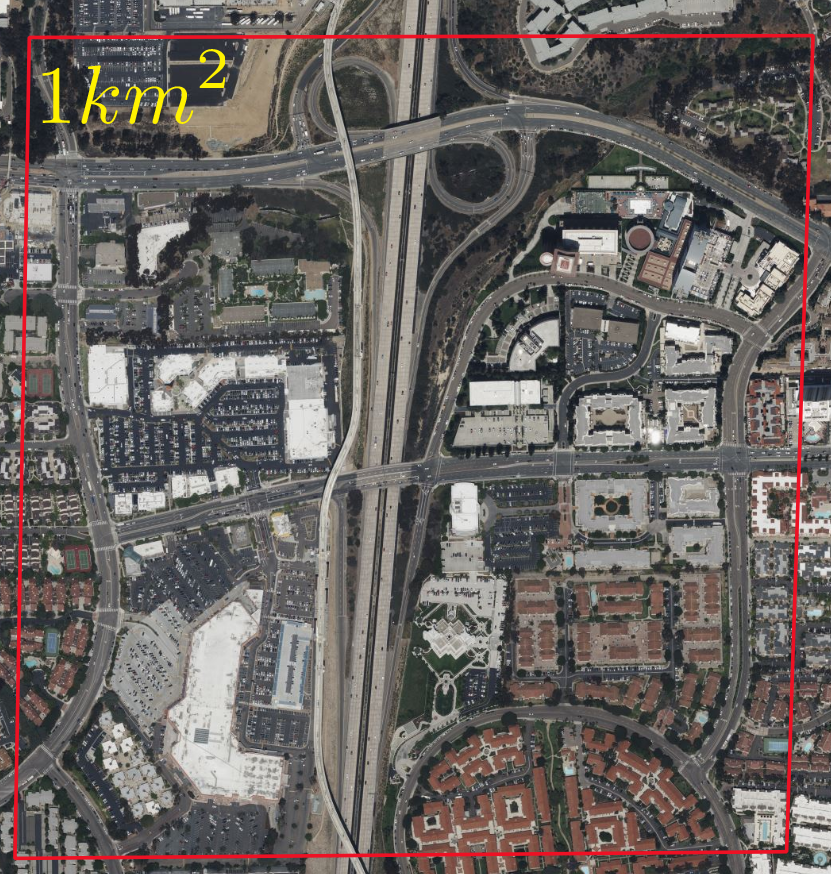}
      \caption{UCSD}
    \end{subfigure}
  \caption{Spatial extents of each AOI (\red{red} box) in the SatDepth dataset overlaid on Bing Maps.}
  \label{fig:sat_coverage_datasheet}
  \vspace{-1em}
\end{figure}

The SatDepth dataset is a subset from the larger CORE3D \cite{Brown_SPIE2018_core3d,core3d_iarpa} and MVS3DM \cite{bosch2016multiple_mvs3d} datasets. To create the SatDepth dataset, we selected a spatial subset of the four AOIs as shown in \refFig{fig:sat_coverage_datasheet}. We chose Jacksonville as our large-area AOI as we had Ground Control Points (GCPs) available for this region, allowing us to assess the accuracy of the SatDepth dataset. For other regions in the SatDepth dataset, we arbitrarily chose a small AOI such that the entire AOI was visible in all the images. Furthermore, for Jacksonville, we discard the tiles with total water coverage or tiles which failed during bundle adjustment as detailed in \cref{sec:dataset_splits}.

\begin{figure}[!h]
  \centering
  \begin{subfigure}[b]{0.2\linewidth}
    \includegraphics[width=\linewidth]{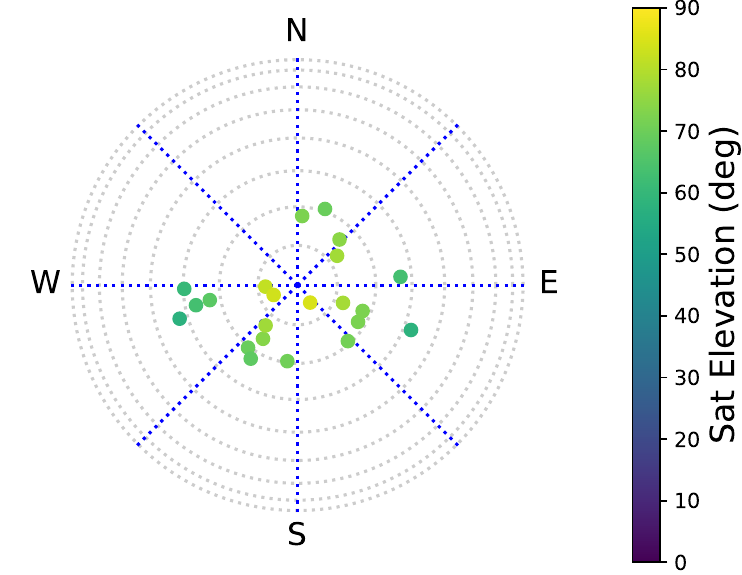}
    \caption{Jacksonville}
  \end{subfigure}
  \begin{subfigure}[b]{0.2\linewidth}
    \includegraphics[width=\linewidth]{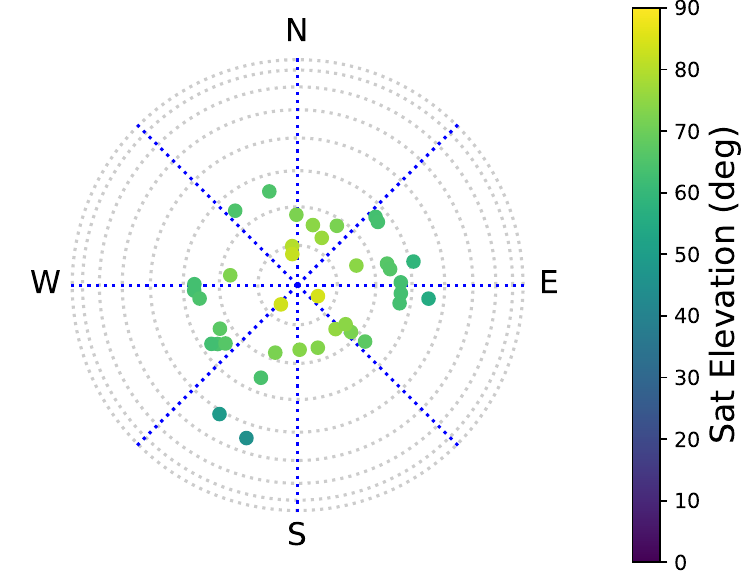}
    \caption{San Fernando}
  \end{subfigure}
  \begin{subfigure}[b]{0.2\linewidth}
    \includegraphics[width=\linewidth]{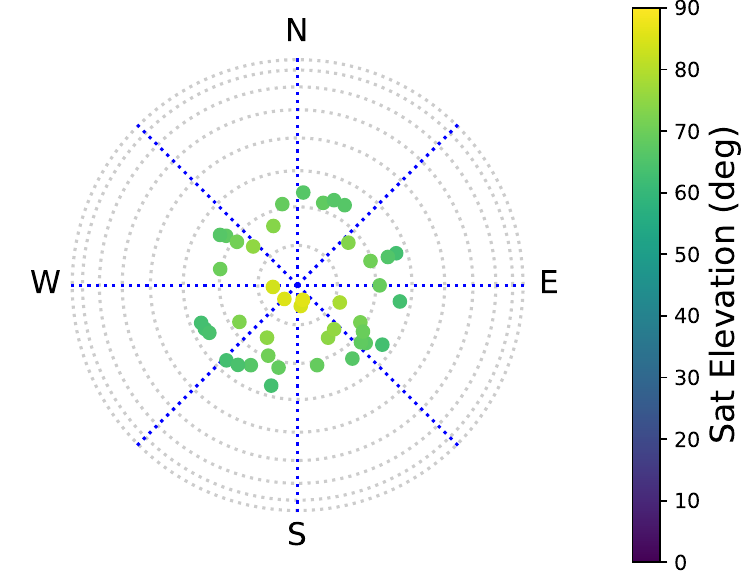}
    \caption{Omaha}
  \end{subfigure}
  \begin{subfigure}[b]{0.2\linewidth}
    \includegraphics[width=\linewidth]{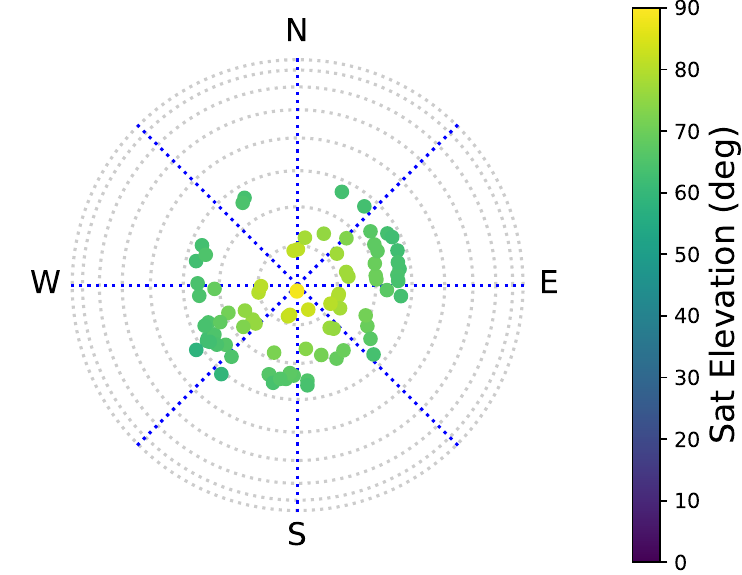}
    \caption{UCSD}
  \end{subfigure}
  \caption{Satellite azimuth and elevation distribution for each AOI in the SatDepth dataset. Images are depicted using colored dots in the polar plot. The satellite azimuth angle can be inferred using the polar plot. The satellite elevation angle is shown using the color bar.}
  \label{fig:sat_radial_plot}
\end{figure}

\begin{figure}[!h]
  \centering
  \begin{subfigure}[b]{0.2\linewidth}
    \includegraphics[width=\linewidth]{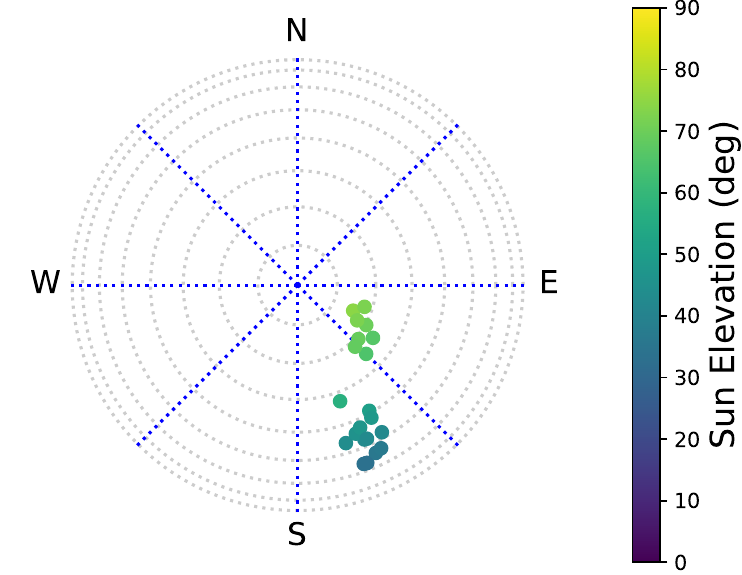}
    \caption{Jacksonville}
  \end{subfigure}
  \begin{subfigure}[b]{0.2\linewidth}
    \includegraphics[width=\linewidth]{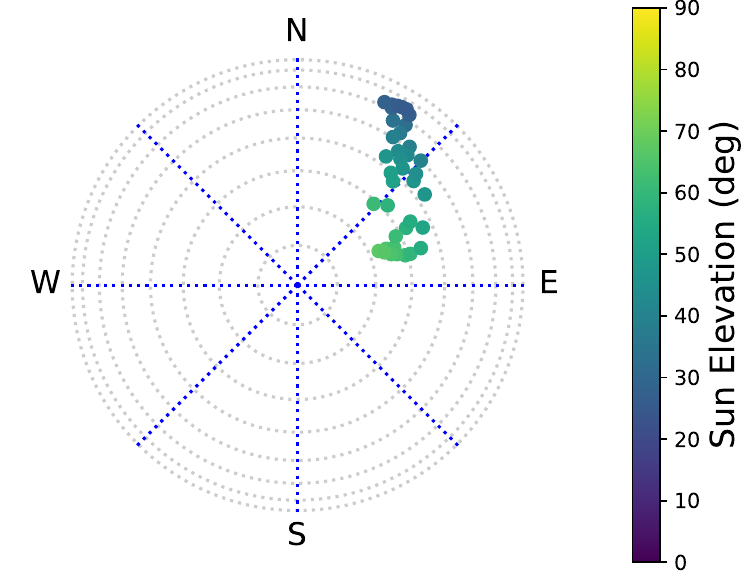}
    \caption{San Fernando}
  \end{subfigure}
  \begin{subfigure}[b]{0.2\linewidth}
    \includegraphics[width=\linewidth]{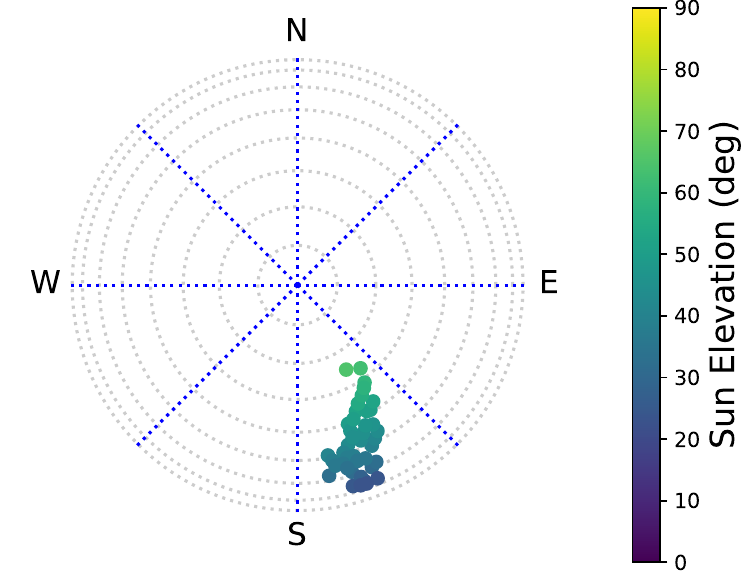}
    \caption{Omaha}
  \end{subfigure}
  \begin{subfigure}[b]{0.2\linewidth}
    \includegraphics[width=\linewidth]{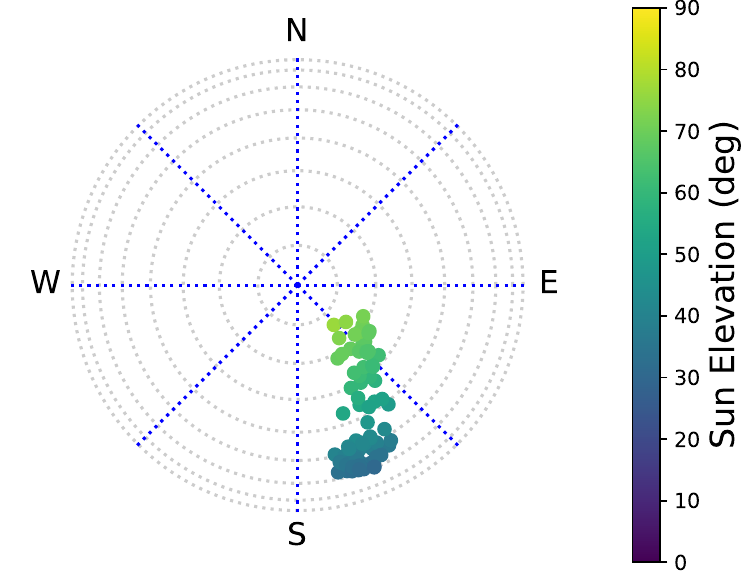}
    \caption{UCSD}
  \end{subfigure}
  \caption{Sun azimuth and elevation distribution for each AOI in the SatDepth dataset. Images are depicted using colored dots in the polar plot. The Sun azimuth angle can be inferred using the polar plot. The Sun elevation angle is shown using the color bar.}
  \label{fig:sun_radial_plot}
\end{figure}

For all AOIs in the SatDepth dataset, we use all the satellite images that are available in the CORE3D and MVS3DM datasets. We do not discard any images from the original datasets. We show the distribution of satellite and sun viewing angles for all the satellite images of the four AOIs in \refFig{fig:sat_radial_plot} and \refFig{fig:sun_radial_plot}, respectively.

\dsquestionex{What data does each instance consist of? “Raw” data (e.g., unprocessed text or images) or features?}{In either case, please provide a description.}

Each instance (\ie tile) in the dataset consists of the following:
\begin{itemize}[itemsep=0.5ex]
  \item \textbf{Satellite Images}: These are PAN band images of the WorldView satellites. The images are in the GeoTIFF format. Each tile has multiple images.
  \item \textbf{RPC Cameras}: These are the Rational Polynomial Coefficients (RPC) cameras for the satellite images. The cameras are provided in the RPB format. Each tile has the same number of cameras as the number of images.
  \item \textbf{Image Metadata}: These are metadata files for the satellite images. The metadata is in the IMD format. Each region has the same number of metadata files as the number of images.
  \item \textbf{DSM}: These are the Digital Surface Models (DSM). The DSM is in the GeoTIFF format. Each tile has a single DSM file.
  \item \textbf{AOI Extents}: These are the geographic extents (rectangular polygon) of the tile in the KML format. Each tile has a single KML file.
  \item \textbf{SatDepth Maps}: These are the $(lat,\,lon,\,h)$ associations for each pixel in the satellite image. The SatDepth Maps store the $(lat,\,lon,\,h)$ values in three separate TIFF files. Each tile has the same number of SatDepth Maps as the number of satellite images.
\end{itemize}

The SatDepth dataset has a file structure as shown in Listing \cref{lst:file_structure}. The first level of folders are for the four AOIs in SatDepth. Then for each AOI, we have a folder called \texttt{IMD} which contains the image metadata (\texttt{*.IMD} files). The folder for each AOI also has subfolders for tiles with the name \texttt{aoi\_rect\_piece\_*}. 

The satellite images and the RPC cameras for a tile are stored in the \texttt{DSM\_Cropped\_Images} subfolder. The SatDepth Maps for the tile are stored in the \texttt{Depth} subfolder. The satellite images (\texttt{*.tif}), the RPC cameras (\texttt{*.RPB}), the image metadata (\texttt{*.IMD}) and the SatDepth Maps (\texttt{*\_lat.tif, *\_lon.tif, *\_depth.tif }) have the same basename for every unique satellite image. The basename is based on the WorldView naming convention. The DSM for the tile is stored with the name \texttt{aoi\_rect\_piece\_*\_DSM-wgs84\_unpadded.tif}. The geographic extents of the tile are stored in the \texttt{aoi\_rect\_piece\_*.kml} file. 

The GCP annotations for the tile are stored in the \texttt{gcp} subfolder. The \texttt{gcp} subfolder is present only in tiles which enclose a GCP (\cref{fig:jax_all_gcps}). It contains the \texttt{annotations} subfolder which contains the \texttt{GCP\_*\_annotations.csv} file for each GCP. The csv file stores the annotated pixel coordinates on the satellite image.

\begin{lstlisting}[caption={File Structure for SatDepth},label={lst:file_structure},basicstyle=\ttfamily,frame=single]
  region
  |-- IMD
  |   `-- *.IMD
  `-- aoi_rect_piece_*
      |-- DSM_Cropped_Images
      |    |-- *.tif
      |    `-- *.RPB
      |-- Depth
      |   |-- *_lat.tif
      |   |-- *_lon.tif
      |   `-- *_depth.tif 
      |-- gcp
      |   |-- annotations
      |       `-- GCP_*_annotations.csv
      |-- aoi_rect_piece_*_DSM-wgs84_unpadded.tif
      `-- aoi_rect_piece_*.kml
\end{lstlisting}

\dsquestionex{Is there a label or target associated with each instance?}{If so, please provide a description.}

The dataset does not have any labels or targets associated with each instance.

\dsquestionex{Is any information missing from individual instances?}{If so, please provide a description, explaining why this information is missing (e.g., because it was unavailable). This does not include intentionally removed information, but might include, e.g., redacted text.}

No information is missing from the instances.

\dsquestionex{Are relationships between individual instances made explicit (e.g., users’ movie ratings, social network links)?}{If so, please describe how these relationships are made explicit.}

\NA

\dsquestionex{Are there recommended data splits (e.g., training, development/validation, testing)?}{If so, please provide a description of these splits, explaining the rationale behind them.}

We recommend using our data splits for future work on satellite image matching. A summary of the image pairs for our data splits is presented in \cref{tab:satdepth_splits}. We provide the rationale behind the data splits in \cref{sec:dataset_splits}. We also provide the data splits in the form of CSV files which can be downloaded from \github.

\begin{table}[!h]
  \caption{SatDepth dataset splits}
  \label{tab:satdepth_splits}
  \centering
  \begin{tabular}{ccccc}
  \toprule
  AOI                   & Split Type & Area (\sqkm) & \# Tiles & \# Image Pairs \\
  \midrule
  \multirow{3}{*}{Jacksonville} & Train      & 99 & 99        & 12977 \\
                                & Validation & 11 & 11        & 1471 \\                      
                                & Test       & 14 & 14        & 3238 \\
  \midrule
  San Fernando  & Test & 1   & 1       &  1081 \\
  \midrule
  Omaha         & Test & 2.2 & 1      &  861 \\
  \midrule
  UCSD          & Test & 1   & 1      &  3081 \\
  \bottomrule
  \end{tabular}
\end{table}

\dsquestionex{Are there any errors, sources of noise, or redundancies in the dataset?}{If so, please provide a description.}

The SatDepth dataset provides a way to extract ground-truth correspondences between satellite images using SatDepth Maps. As discussed in our paper \cite{satdepth_preprint}, the SatDepth Maps are generated using DSM. As the DSM is constructed from a limited number of stereo pairs, it may fail to capture accurate height information in areas experiencing construction or other changes. This error in height information can lead to extraction of false matches for such areas.

\dsquestionex{Is the dataset self-contained, or does it link to or otherwise rely on external resources (e.g., websites, tweets, other datasets)?}{If it links to or relies on external resources, a) are there guarantees that they will exist, and remain constant, over time; b) are there official archival versions of the complete dataset (i.e., including the external resources as they existed at the time the dataset was created); c) are there any restrictions (e.g., licenses, fees) associated with any of the external resources that might apply to a future user? Please provide descriptions of all external resources and any restrictions associated with them, as well as links or other access points, as appropriate.}

The dataset is self-contained.

\dsquestionex{Does the dataset contain data that might be considered confidential (e.g., data that is protected by legal privilege or by doctor-patient confidentiality, data that includes the content of individuals non-public communications)?}{If so, please provide a description.}

There is no confidential data in this dataset.

\dsquestionex{Does the dataset contain data that, if viewed directly, might be offensive, insulting, threatening, or might otherwise cause anxiety?}{If so, please describe why.}

There is no offensive data in this dataset.

\dsquestionex{Does the dataset relate to people?}{If not, you may skip the remaining questions in this section.}

The dataset does not relate to people or disclose any personal information.

\dsquestionex{Does the dataset identify any subpopulations (e.g., by age, gender)?}{If so, please describe how these subpopulations are identified and provide a description of their respective distributions within the dataset.}

\NA

\dsquestionex{Is it possible to identify individuals (i.e., one or more natural persons), either directly or indirectly (i.e., in combination with other data) from the dataset?}{If so, please describe how.}

\NA

\dsquestionex{Does the dataset contain data that might be considered sensitive in any way (e.g., data that reveals racial or ethnic origins, sexual orientations, religious beliefs, political opinions or union memberships, or locations; financial or health data; biometric or genetic data; forms of government identification, such as social security numbers; criminal history)?}{If so, please provide a description.}

\NA

\dsquestion{Any other comments?}

None.

\bigskip
\dssectionheader{Collection Process}
\dsquestionex{How was the data associated with each instance acquired?}{Was the data directly observable (e.g., raw text, movie ratings), reported by subjects (e.g., survey responses), or indirectly inferred/derived from other data (e.g., part-of-speech tags, model-based guesses for age or language)? If data was reported by subjects or indirectly inferred/derived from other data, was the data validated/verified? If so, please describe how.}

To create the SatDepth dataset we utilize two publicly available multi-view, multi-date datasets for Very High Resolution (VHR) satellite images: (1) IARPA CORE3D-Public dataset \cite{Brown_SPIE2018_core3d, core3d_aws_public, core3d_iarpa} is comprised of WorldView (WV-2 and WV-3) images for three cities: Jacksonville, FL; Omaha, NE; and UCSD, CA. It includes WorldView panchromatic and multi-spectral images for each city. And (2) IARPA MVS3DM Challenge dataset \cite{bosch2016multiple_mvs3d} which contains images over San Fernando, Argentina. This dataset spans 100 \sqkm, includes 30 cm airborne LiDAR ground truth data for a 20 \sqkm subset of the larger area. These datasets are hosted through AWS and can be accessed through the project webpages of \href{https://spacenet.ai/core3d/}{CORE3D} and \href{https://spacenet.ai/iarpa-multi-view-stereo-3d-mapping/}{MVS3DM} respectively

\dsquestionex{What mechanisms or procedures were used to collect the data (e.g., hardware apparatus or sensor, manual human curation, software program, software API)?}{How were these mechanisms or procedures validated?}

To collect the raw data, we follow the instructions of each project webpages (\href{https://spacenet.ai/core3d/}{CORE3D} and \href{https://spacenet.ai/iarpa-multi-view-stereo-3d-mapping/}{MVS3DM}) to download the data. These datasets store the WorldView images as tarballs, which were uncompressed to obtain the raw images, cameras, and metadata.

\dsquestion{If the dataset is a sample from a larger set, what was the sampling strategy (e.g., deterministic, probabilistic with specific sampling probabilities)?}

The SatDepth dataset is a subset from the larger CORE3D\cite{Brown_SPIE2018_core3d, core3d_aws_public, core3d_iarpa} and MVS3DM \cite{bosch2016multiple_mvs3d} datasets. To create the SatDepth dataset, we selected a spatial subset of the four AOIs as shown in \refFig{fig:sat_coverage_datasheet}. We chose Jacksonville as our large-area AOI as we had Ground Control Points (GCPs) available for this region, allowing us to assess the accuracy of the SatDepth dataset. For other regions in the SatDepth dataset, we arbitrarily chose a small AOI such that the entire AOI was visible in all the images.

\dsquestion{Who was involved in the data collection process (e.g., students, crowdworkers, contractors) and how were they compensated (e.g., how much were crowdworkers paid)?}

The data was assembled by the authors of this paper \cite{satdepth_preprint}. The actual collection of WorldView satellite images was done by Maxar and the raw images were made publicly available through the IARPA CORE3D and MVS3DM projects.

\dsquestionex{Over what timeframe was the data collected? Does this timeframe match the creation timeframe of the data associated with the instances (e.g., recent crawl of old news articles)?}{If not, please describe the timeframe in which the data associated with the instances was created.}

The dataset contains satellite images collected between 2014-2016 as detailed in \cref{tab:satdepth_summary}.

\dsquestionex{Were any ethical review processes conducted (e.g., by an institutional review board)?}{If so, please provide a description of these review processes, including the outcomes, as well as a link or other access point to any supporting documentation.}

\NA

\dsquestionex{Does the dataset relate to people?}{If not, you may skip the remaining questions in this section.}

The dataset does not relate to people or contain any personal information.

\dsquestion{Did you collect the data from the individuals in question directly, or obtain it via third parties or other sources (e.g., websites)?}

\NA

\dsquestionex{Were the individuals in question notified about the data collection?}{If so, please describe (or show with screenshots or other information) how notice was provided, and provide a link or other access point to, or otherwise reproduce, the exact language of the notification itself.}

\NA

\dsquestionex{Did the individuals in question consent to the collection and use of their data?}{If so, please describe (or show with screenshots or other information) how consent was requested and provided, and provide a link or other access point to, or otherwise reproduce, the exact language to which the individuals consented.}

\NA

\dsquestionex{If consent was obtained, were the consenting individuals provided with a mechanism to revoke their consent in the future or for certain uses?}{If so, please provide a description, as well as a link or other access point to the mechanism (if appropriate).}

\NA

\dsquestionex{Has an analysis of the potential impact of the dataset and its use on data subjects (e.g., a data protection impact analysis) been conducted?}{If so, please provide a description of this analysis, including the outcomes, as well as a link or other access point to any supporting documentation.}

\NA

\dsquestion{Any other comments?}

None.

\bigskip
\dssectionheader{Preprocessing/cleaning/labeling}
\dsquestionex{Was any preprocessing/cleaning/labeling of the data done (e.g., discretization or bucketing, tokenization, part-of-speech tagging, SIFT feature extraction, removal of instances, processing of missing values)?}{If so, please provide a description. If not, you may skip the remainder of the questions in this section.}

The dataset was created using our 3D reconstruction pipeline. We give details of this pipeline in our paper \cite{satdepth_preprint}.

\dsquestionex{Was the “raw” data saved in addition to the preprocessed/cleaned/labeled data (e.g., to support unanticipated future uses)?}{If so, please provide a link or other access point to the “raw” data.}

The raw images are available through the IARPA CORE3D and MVS3DM projects. We don't save intermediate data generated by our pipeline.

\dsquestionex{Is the software used to preprocess/clean/label the instances available?}{If so, please provide a link or other access point.}

The 3D reconstruction pipeline is not available for public use. However, we provide access to the \textbf{\textit{depthifypp}} module which generates the SatDepth Maps from DSMs. This module will be made available through \github.

\dsquestion{Any other comments?}

None.

\bigskip
\dssectionheader{Uses}
\dsquestionex{Has the dataset been used for any tasks already?}{If so, please provide a description.}

We demonstrate the use of the dataset for satellite image matching in our paper \cite{satdepth_preprint}.

\dsquestionex{Is there a repository that links to any or all papers or systems that use the dataset?}{If so, please provide a link or other access point.}

SatDepth is a new dataset. It has not been used for any other papers or systems yet.

\dsquestion{What (other) tasks could the dataset be used for?}

In addition to image matching, the dataset can possibly be used for monocular height estimation. 

\dsquestionex{Is there anything about the composition of the dataset or the way it was collected and preprocessed/cleaned/labeled that might impact future uses?}{For example, is there anything that a future user might need to know to avoid uses that could result in unfair treatment of individuals or groups (e.g., stereotyping, quality of service issues) or other undesirable harms (e.g., financial harms, legal risks) If so, please provide a description. Is there anything a future user could do to mitigate these undesirable harms?}

None to the best of our knowledge.

\dsquestionex{Are there tasks for which the dataset should not be used?}{If so, please provide a description.}

None to the best of our knowledge.

\dsquestion{Any other comments?}

None.

\bigskip
\dssectionheader{Distribution}
\dsquestionex{Will the dataset be distributed to third parties outside of the entity (e.g., company, institution, organization) on behalf of which the dataset was created?}{If so, please provide a description.}

The dataset will be made publicly available.

\dsquestionex{How will the dataset will be distributed (e.g., tarball on website, API, github)}{Does the dataset have a digital object identifier (DOI)?}

Details will be conveyed through our \\github page.



\dsquestion{When will the dataset be distributed?}
\TBD


\dsquestionex{Will the dataset be distributed under a copyright or other intellectual property (IP) license, and/or under applicable terms of use (ToU)?}{If so, please describe this license and/or ToU, and provide a link or other access point to, or otherwise reproduce, any relevant licensing terms or ToU, as well as any fees associated with these restrictions.}

\TBD

\dsquestionex{Have any third parties imposed IP-based or other restrictions on the data associated with the instances?}{If so, please describe these restrictions, and provide a link or other access point to, or otherwise reproduce, any relevant licensing terms, as well as any fees associated with these restrictions.}

Not applicable.

\dsquestionex{Do any export controls or other regulatory restrictions apply to the dataset or to individual instances?}{If so, please describe these restrictions, and provide a link or other access point to, or otherwise reproduce, any supporting documentation.}

Not applicable.

\dsquestion{Any other comments?}

None.

\bigskip
\dssectionheader{Maintenance}
\dsquestion{Who will be supporting/hosting/maintaining the dataset?}

The authors will be supporting/hosting/maintaining the dataset.

\dsquestion{How can the owner/curator/manager of the dataset be contacted (e.g., email address)?}

The authors can be contacted through email and \github ~issues.


\dsquestionex{Is there an erratum?}{If so, please provide a link or other access point.}

Not applicable. 

\dsquestionex{Will the dataset be updated (e.g., to correct labeling errors, add new instances, delete instances)?}{If so, please describe how often, by whom, and how updates will be communicated to users (e.g., mailing list, github)?}

The dataset may be updated to add new instances in the future. However, we do not anticipate frequent updates to the dataset. Any updates will be communicated through the \github ~page.

\dsquestionex{If the dataset relates to people, are there applicable limits on the retention of the data associated with the instances (e.g., were individuals in question told that their data would be retained for a fixed period of time and then deleted)?}{If so, please describe these limits and explain how they will be enforced.}

The dataset does not relate to people. There are no such limits applicable.

\dsquestionex{Will older versions of the dataset continue to be supported/hosted/maintained?}{If so, please describe how. If not, please describe how its obsolescence will be communicated to users.}

We will maintain the older versions of the dataset for the foreseeable future. Any changes made to the dataset will be communicated through the \github ~page.

\dsquestionex{If others want to extend/augment/build on/contribute to the dataset, is there a mechanism for them to do so?}{If so, please provide a description. Will these contributions be validated/verified? If so, please describe how. If not, why not? Is there a process for communicating/distributing these contributions to other users? If so, please provide a description.}

We will accept all such requests made through the \github ~issues. However, 
it will be up to the users who make such requests to
validate/verify their contributions.

\dsquestion{Any other comments?}

None.

{
    \small
    \bibliographystyle{ieeetr}
    \bibliography{reference}
}

\end{document}